\documentclass{article} 
\usepackage{iclr2027_conference,times}
\usepackage{amsmath,amssymb}
\usepackage{graphicx}
\usepackage{wrapfig}
\usepackage{booktabs}
\usepackage{tabularx}

\usepackage{amsmath,amsfonts,bm}

\def\eqref#1{equation~\ref{#1}}

\def\1{\bm{1}}

\DeclareMathAlphabet{\mathsfit}{\encodingdefault}{\sfdefault}{m}{sl}
\SetMathAlphabet{\mathsfit}{bold}{\encodingdefault}{\sfdefault}{bx}{n}

\usepackage{amsmath}   
\usepackage{amssymb}
\usepackage{graphicx}  
\usepackage{booktabs}  
\usepackage{tabularx}  
\usepackage{longtable} 
\usepackage[T1,OT1]{fontenc} 
\usepackage{CJKutf8}          
\usepackage{pifont}           
\usepackage{fontawesome5}     
\usepackage[table]{xcolor}
\usepackage[most]{tcolorbox}
\newcommand{\zh}[1]{\begin{CJK}{UTF8}{gbsn}#1\end{CJK}}
\definecolor{apxDark}{HTML}{3D3D3D}   \definecolor{apxBody}{HTML}{F3F3F3}
\definecolor{apxPeach}{HTML}{FCE9D6}  \definecolor{apxOrange}{HTML}{E9A23B}
\definecolor{apxPink}{HTML}{F4CCCC}   \definecolor{apxGreen}{HTML}{D9EAD3}
\definecolor{apxBlueP}{HTML}{E6EEF8}  \definecolor{apxBlueFrame}{HTML}{8DB2E2}
\definecolor{apxGreenQ}{HTML}{DDE8D2} \definecolor{apxGreenFrame}{HTML}{9CC48A}
\definecolor{apxPassRow}{HTML}{E9F4E9} \definecolor{apxFailRow}{HTML}{FBDADA} \definecolor{apxUnclearRow}{HTML}{FFF3CD}
\definecolor{apxPassFg}{HTML}{2E8B32}  \definecolor{apxFailFg}{HTML}{D32F2F}   \definecolor{apxUnclearFg}{HTML}{B7791F}
\definecolor{apxHlP}{HTML}{1F5BD6}     \definecolor{apxHlQ}{HTML}{C2187A}     \definecolor{apxHlB}{HTML}{E07A00}
\colorlet{apxPanelP}{apxBlueP}\colorlet{apxPanelFrameP}{apxBlueFrame}\colorlet{apxPanelQ}{apxGreenQ}\colorlet{apxPanelFrameQ}{apxGreenFrame}
\definecolor{apxBanner}{HTML}{EDEDED}  \definecolor{apxGrayFrame}{HTML}{9A9A9A}
\newtcolorbox{promptbox}[1]{enhanced,breakable,colback=apxBody,colframe=apxDark,coltitle=white,
  fonttitle=\small,title={#1},arc=1.5mm,boxrule=0.7pt,left=6pt,right=6pt,top=4pt,bottom=4pt,
  before upper={\begin{CJK}{UTF8}{gbsn}\fontencoding{T1}\selectfont\small\sloppy\setlength{\parskip}{3pt}},
  after upper={\end{CJK}}}
\newtcolorbox{jsonbox}[1]{enhanced,breakable,colback=apxBody,colframe=apxDark,coltitle=white,
  fonttitle=\small,title={#1},arc=1.5mm,boxrule=0.7pt,left=6pt,right=6pt,top=4pt,bottom=4pt,
  before upper={\begin{CJK}{UTF8}{gbsn}\fontencoding{T1}\selectfont\footnotesize\ttfamily\sloppy\setlength{\parskip}{2pt}},
  after upper={\end{CJK}}}
\newtcolorbox{protocolbox}[1]{enhanced,breakable,colback=apxBody,colframe=apxDark,coltitle=white,
  fonttitle=\small,title={#1},arc=1.5mm,boxrule=0.7pt,left=6pt,right=6pt,top=4pt,bottom=4pt,
  before upper={\begin{CJK}{UTF8}{gbsn}\small},after upper={\end{CJK}}}
\newcommand{\apxicon}[2]{\tikz[baseline=-0.6ex]\node[circle,fill=#1,draw=white,line width=1pt,minimum size=8.5mm,inner sep=0pt,text=white]{\large #2};}
\newcommand{\casefail}{\apxicon{apxFailFg!85}{\faTimes}}
\newcommand{\casepass}{\apxicon{apxPassFg!90}{\faCheck}}
\newcommand{\caseinfo}{\apxicon{apxHlP!85}{\faInfo}}
\newcommand{\casewarn}{\apxicon{apxUnclearFg!90}{\faExclamation}}
\newtcolorbox{casebox}[2]{enhanced,colback=white,colframe=apxGrayFrame,boxrule=0.8pt,sharp corners,
  left=5pt,right=5pt,top=13mm,bottom=5pt,
  overlay={\node[anchor=north,fill=apxBanner,minimum height=8mm,text width=0.72\linewidth,align=center,
      font=\sffamily\bfseries\normalsize] at ([yshift=-2mm]frame.north) {#2};
    \node at ([xshift=8mm,yshift=-6mm]frame.north west) {#1};}}
\newtcolorbox{caseband}{colback=apxPeach,colframe=apxPeach,boxrule=0pt,sharp corners,left=5pt,right=5pt,top=3pt,bottom=3pt,
  before upper={\begin{CJK}{UTF8}{gbsn}\fontencoding{T1}\selectfont\footnotesize\sffamily},after upper={\end{CJK}}}
\newtcolorbox{casepanel}[2][apxGreen]{colback=#1,colframe=#1,boxrule=0pt,sharp corners,left=5pt,right=5pt,top=3pt,bottom=3pt,
  fonttitle=\sffamily\bfseries\footnotesize,coltitle=black,colbacktitle=#1,title={#2},
  before upper={\begin{CJK}{UTF8}{gbsn}\fontencoding{T1}\selectfont\footnotesize\sffamily\raggedright},after upper={\end{CJK}}}
\newcommand{\casecols}[2]{\par\noindent\begin{minipage}[t]{0.49\linewidth}#1\end{minipage}\hfill\begin{minipage}[t]{0.49\linewidth}#2\end{minipage}\par}
\newcommand{\ok}{\textcolor{apxPassFg}{\faCheckSquare}}
\newcommand{\no}{\textcolor{apxFailFg}{\faWindowClose}}
\newcommand{\unk}{\textcolor{apxUnclearFg}{\faQuestionCircle}}
\newcommand{\rowpass}{\rowcolor{apxPassRow}}
\newcommand{\rowfail}{\rowcolor{apxFailRow}}
\newcommand{\rowunclear}{\rowcolor{apxUnclearRow}}
\newcommand{\PASS}{\textcolor{apxPassFg}{\ding{51}\,\textsf{\textbf{PASS}}}}
\newcommand{\FAIL}{\textcolor{apxFailFg}{\ding{55}\,\textsf{\textbf{FAIL}}}}

\newcommand{\badge}[2]{\textcolor{#1}{\textsf{\textbf{#2}}}}
\newtcolorbox{taskpanel}[1]{enhanced,colback=apxPeach,colframe=apxOrange,boxrule=0.8pt,arc=3mm,
  left=5pt,right=5pt,top=3pt,bottom=4pt,
  before upper={\begin{CJK}{UTF8}{gbsn}\centering{\sffamily\bfseries #1}\par\raggedright\fontencoding{T1}\selectfont\footnotesize\sffamily},after upper={\end{CJK}}}
\newtcolorbox{userpanel}[2][P]{enhanced,arc=3mm,boxrule=0.8pt,left=5pt,right=5pt,top=3pt,bottom=4pt,
  colback=apxPanel#1,colframe=apxPanelFrame#1,
  before upper={\begin{CJK}{UTF8}{gbsn}\centering{\sffamily\bfseries #2}\par\raggedright\fontencoding{T1}\selectfont\footnotesize\sffamily},after upper={\end{CJK}}}
\newtcolorbox{dashedgroup}{enhanced,colback=white,colframe=white,boxrule=0pt,arc=4mm,left=3pt,right=3pt,top=3pt,bottom=3pt,
  borderline={0.6pt}{0pt}{apxGreenFrame,dashed}}
\newcommand{\iconrow}[3][apxHlP]{\par\noindent\begin{minipage}[c]{9mm}\centering\textcolor{#1}{\LARGE #2}\end{minipage}\hspace{1.5mm}\begin{minipage}[c]{\dimexpr\linewidth-11mm\relax}\raggedright #3\end{minipage}\par\smallskip}
\newcommand{\hlP}[1]{\textcolor{apxHlP}{#1}}   
\newcommand{\hlQ}[1]{\textcolor{apxHlQ}{#1}}   

\usepackage{microtype}
\usepackage{xurl}
\usepackage{placeins}
\usepackage{hyperref}
\hypersetup{hidelinks,pdftitle={What Happens During Autonomous Deep Research After the User Steps Away?},pdfauthor={Yimin Liu, Yijia Zhang, Yanmin Li, Tangwen Luo, Yuze Li, Ziling Yao, Zhi Yang}}
\usepackage{url}

\title{What Happens During Autonomous\\Deep Research After the User Steps Away?}

\newcommand{\DRauthor}[3]{%
  \href{https://openreview.net/profile?id=#3}{#1}%
  \raisebox{4.6bp}{\normalfont\fontsize{7.3bp}{8bp}\selectfont #2}}
\newcommand{\DRaffiliation}[3]{%
  \raisebox{3.7bp}{\fontsize{6.6bp}{8bp}\selectfont\color{black!65}#1}%
  \,\href{#3}{#2}}
\author{%
  \DRauthor{Yimin Liu}{1}{\string~Yimin_Liu11}\hspace{14bp}%
  \DRauthor{Yijia Zhang}{2}{\string~Yijia_Zhang23}\hspace{14bp}%
  \DRauthor{Yanmin Li}{3}{\string~Yanmin_Li3}\hspace{14bp}%
  \DRauthor{Tangwen Luo}{4}{\string~Tangwen_Luo1}\\
  \DRauthor{Yuze Li}{5}{\string~Yuze_Li7}\hspace{14bp}%
  \DRauthor{Ziling Yao}{6}{\string~Ziling_Yao1}\hspace{14bp}%
  \DRauthor{Zhi Yang}{6}{\string~Zhi_Yang11}%
}
\newcommand{\DRaffiliations}{%
  \DRaffiliation{1}{University of Southern California}{https://usc.edu}\\
  \DRaffiliation{2}{University of Michigan, Ann Arbor}{https://umich.edu}\\
  \DRaffiliation{3}{Institute of Automation, Chinese Academy of Sciences}{https://ia.ac.cn}\\
  \DRaffiliation{4}{Zhejiang University}{https://zju.edu.cn}\\
  \DRaffiliation{5}{Beijing University of Posts and Telecommunications}{https://bupt.edu.cn}\\
  \DRaffiliation{6}{Shanghai University of Finance and Economics}{https://sufe.edu}%
}

\iclrfinalcopy
\fancypagestyle{dralignedfirst}{%
  \fancyhead{}%
  \renewcommand{\headrulewidth}{0pt}%
}
\newcommand{\DRfirstpageheader}{%
  \AddToShipoutPictureFG*{%
    \AtPageUpperLeft{%
      \put(\LenToUnit{108bp},\LenToUnit{-15bp}){%
        \raisebox{-\height}{\includegraphics[width=160bp]{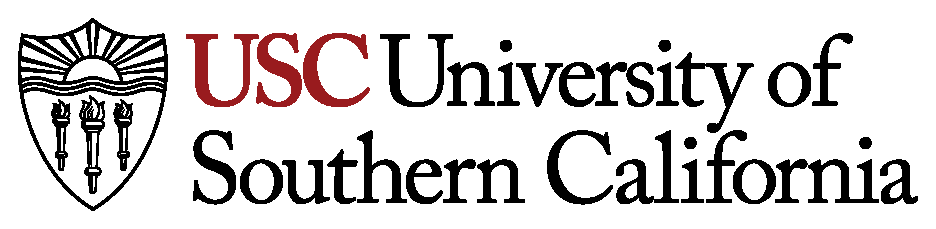}}}%
      \put(\LenToUnit{108bp},\LenToUnit{-65bp}){%
        \color{black!82}\rule{396bp}{0.4bp}}%
    }%
  }%
}
\makeatletter
\newlength{\DRtitleheight}
\def\@maketitle{%
  \thispagestyle{dralignedfirst}%
  \DRfirstpageheader
  \vbox to \DRtitleheight{\hsize\textwidth
    \kern 12bp
    {\centering\LARGE\scshape \@title\par}%
    \kern 17bp
    {\centering\fontsize{12.2bp}{18bp}\selectfont\bfseries \@author\par}%
    \kern 6bp
    {\centering\normalfont\fontsize{9.7bp}{12.6bp}\selectfont
      \color{black!85}\DRaffiliations\par}%
    \vfil
  }%
}
\makeatother
\begin{document}

\maketitle

\begin{abstract}
In autonomous deep research, a user provides a task and relevant background, then leaves the agent to conduct an extended investigation without further human intervention. We study how this initial user information is reflected in intermediate actions and how these actions relate to final recommendations. We introduce DRaligned, a counterfactual behavioral evaluation framework built on PDR-Bench. By varying one task-relevant user factor while keeping the remaining context fixed, we compare acquisition requests, working drafts, and final reports. Source-grounded extraction, blinded local judgments, and deterministic aggregation yield coarse directional measurements while leaving ambiguous cases unresolved. Our experiments show that strong user-specific delivery can emerge from a largely shared research process: agents investigate similar broad questions but allocate requests differently, and final recommendations distinguish user conditions more clearly than explicit requests do. Reports can also integrate user factors that were not jointly visible during acquisition. In readable draft-to-report comparisons, recommendations often retain their coarse user-specific direction despite substantial rewriting. Final directional differences recur across tested agent models, execution harnesses, and evaluator models, even as execution paths vary. These findings describe how initial user information shapes autonomous research and clarify the relationship between the process an agent follows and the recommendations it delivers.
\end{abstract}

\section{Introduction}
Autonomous deep-research agents investigate a task and prepare a report from an initial prompt containing user background and constraints. Personalization benchmarks assess whether outputs fit a user profile \citep{salemi2024lamp,samuel2025personagym,zhao2025personalens,tao2025personafeedback}; PDR-Bench extends this evaluation to research reports \citep{liang2026pdr}. We ask a complementary question: \textbf{once the user steps away, what happens during autonomous research, and how do intermediate actions relate to final recommendations?}

We study \textbf{non-interactive long-horizon execution}: the initial task and user information remain fixed, while the agent acquires external evidence without further human intervention. This differs from personalization through conversation \citep{li2025personaconvbench}, clarification \citep{luo2025clarifymt}, or requests for human assistance \citep{trinh2026hilbench}. The setting is relevant to software engineering, deep research, and artifact generation \citep{jimenez2024swebench,liang2026pdr,zheng2025pptagent}; our question concerns how previously supplied information appears in subsequent behavior, not how agents obtain additional guidance.

We introduce \textbf{DRaligned}, a counterfactual behavioral evaluation framework. From PDR-Bench tasks and structured user profiles, we construct matched conditions $P$ and $Q$ differing in one task-relevant factor. We observe \textbf{Acquisition} ($A$) requests, \textbf{Workspace} ($W$) artifacts, and \textbf{Finalization} ($F$). Source-grounded extraction, assignment-blinded local judgments, mirrored endpoint presentation, and deterministic aggregation identify whether behavior expresses one endpoint, both, neither, or remains unresolved. The resulting \textbf{Matrix} and temporal \textbf{Flow} describe coarse directional behavior. Evaluators annotate observed content; controlled comparisons between conditions establish the user-related contrast. The framework therefore studies behavioral user control without assigning an unrestricted personalization score or inferring hidden reasoning. Acquisition, Workspace, and Finalization are observation channels rather than mandatory stages: agents may return to retrieval after writing, or finish without a public draft.

Our analysis shows that \textbf{strong user-specific delivery can emerge from a largely shared research process}. Agents often investigate similar broad questions while allocating requests differently. Final recommendations distinguish user conditions more clearly than explicit requests do and can combine factors not jointly visible during acquisition. A public recommendation often retains its coarse direction despite substantial rewriting. These are factor-specific observations, not evidence that global user control switches on and off. Comparisons across models, reasoning settings, and harnesses reveal different execution paths leading to similar directional outcomes, without establishing equivalence between those settings.

Our contributions are:
\begin{itemize}
\item \textbf{A counterfactual study of autonomous research:} isolating how task-relevant user information relates to intermediate actions and final recommendations after human input stops.
\item \textbf{A selective behavioral measurement framework:} combining source-linked extraction with directional Matrix/Flow readings and explicit uncertainty.
\item \textbf{Empirical findings on research and delivery:} shared questions, changing request allocation, integration of user factors, and directional persistence through rewriting across tested settings.
\end{itemize}

\section{Related Work}
\label{sec:related_work}
\subsection{Personalization and User Alignment Evaluation}
LaMP evaluates generation from user histories \citep{salemi2024lamp}; PersonaGym studies persona adherence \citep{samuel2025personagym}; PersonaFeedback separates response tailoring from persona inference \citep{tao2025personafeedback}. PrefEval, PersonaMem, and AlpsBench evaluate preference following, evolving profiles, and personalized memory \citep{zhao2025prefeval,jiang2025personamem,xiao2026alpsbench}. PDR-Bench evaluates personalization, content quality, and factual reliability in research reports \citep{liang2026pdr}. We use its task--user data to vary one task-relevant factor and compare observable requests and recommendations. This complements report-level evaluation rather than replacing it; local semantic annotation remains model-based. Crucially, user influence is assessed through the controlled comparison, not directly assigned by the evaluator from a report alone.

\subsection{Deep-Research Evaluation and Behavioral Analysis}
DeepResearch Bench and ResearcherBench assess report quality, research responses, and grounding \citep{du2026deepresearchbench,xu2025researcherbench}. Mind2Web~2 evaluates agentic search with task-specific judges \citep{gou2025mind2web2}; DeepResearchGym provides reproducible retrieval \citep{coelho2025deepresearchgym}. LiveDRBench represents intermediate claims and analyzes search structure \citep{java2026livedrbench}. AgentBoard measures progress \citep{ma2024agentboard}, AgentIF evaluates instruction constraints \citep{qi2025agentif}, and Counterfactual Trace Auditing compares executions with and without agent skills \citep{zhou2026cta}. We instead intervene on user information and read factor-specific directions, without requiring step-by-step alignment between independently generated trajectories. Our contribution is thus not intermediate evaluation alone, but its use in studying initial user information during autonomous research.

\subsection{Interactive Adaptation and Human-in-the-Loop Agents}
PersonaLens and UserBench examine personalization through dialogue and incremental preference disclosure \citep{zhao2025personalens,qian2025userbench}. IntentRL learns clarification before deep research \citep{luo2026intentrl}; $\tau$-bench and HiL-Bench study tool--user interaction and help-seeking during execution \citep{yao2024taubench,trinh2026hilbench}. PAHF combines clarification, corrective feedback, and user memory \citep{liang2026pahf}. We study the subsequent execution after the initial task-and-user prompt is fixed. Tools can supply new evidence, but no further human decisions or user-specific information are introduced. We do not test whether eliminating interaction improves performance.

\section{Problem Formulation}
\label{sec:problem}

\subsection{Long-Horizon Deep-Research Execution}
\label{sec:dr_formalization}

A user provides a research task $T$ and initial user information $U$, after
which an agent completes the task without further human intervention or user
input. Let $\pi_\theta$, $\mathcal E$, and $\mathcal H$ denote the LLM policy,
external environment, and execution harness, respectively. We represent the
system state as $S_t=(T,U,H_t,I_t,W_t,D_t,\eta_{\mathcal H})$, where $H_t$
records observable events, $I_t$ contains acquired information, $W_t$ contains
intermediate work products, $D_t$ is the delivery state, and
$\eta_{\mathcal H}$ specifies the harness configuration. An event $e_{t+1}$
updates the state through $S_{t+1}=\mathcal F(S_t,e_{t+1})$, producing a rollout
\begin{equation}
\tau=(S_0,e_1,S_1,\ldots,e_K,S_K)
\sim p_{\pi_\theta,\mathcal H,\mathcal E}(\,\cdot\mid T,U).
\label{eq:rollout}
\end{equation}
The task and user information remain fixed, while external observations and
work products may change. At submission, the delivered artifact is
$Y=\operatorname{render}(D_K)$. Keeping event history separate from the
current artifacts allows us to observe both revisions and their outcomes.

\paragraph{Execution harness.}
A harness specifies how research functions are implemented and connected.
We write $\mathcal H=(\mathcal R,\mathcal B,\Lambda,\Sigma,\Gamma)$, where
$\mathcal R=(R_A,R_W,R_F)$ realizes acquisition, synthesis, and delivery;
$\mathcal B=(B_{AW},B_{WF})$ specifies their boundaries; $\Lambda$ controls
state access; $\Sigma$ specifies scheduling and coordination; and $\Gamma$
provides logging and provenance. Our canonical harness $\mathcal H_{00}$
couples both boundaries in one persistent executor, without mandatory
handoffs or intermediate writing. Alternative harnesses vary these execution
relations without changing the task or user intervention
(Appendix~\ref{app:harness}).

\subsection{Functional Observation Channels}
\label{sec:channels}

We distinguish three functions: \textbf{Acquisition} ($A$), which seeks
external information; \textbf{Workspace} ($W$), which records intermediate
analyses, drafts, and revisions; and \textbf{Finalization} ($F$), which
commits a user-facing deliverable. These functions concern the information,
workspace, and delivery components of $S_t$, respectively. For classified
behavioral events, $\chi(e_t)\in\{A,W,F\}$ identifies the function, and
$P_c(\tau)=(e_t:\chi(e_t)=c)$ denotes the corresponding ordered subsequence.
Acquisition and workspace activity may interleave; the channels are not fixed
stages. This functional view accommodates different tools and execution
architectures without equating their internal implementations.

\subsection{Counterfactual User Control}
\label{sec:counterfactual_control}

For a task-relevant factor $f$, we construct user conditions
$U_f^P=C\oplus z_f^P$ and $U_f^Q=C\oplus z_f^Q$, where $C$ is shared
background and $\oplus$ denotes prompt assembly. Only the selected factor
changes, and both conditions must remain valid for the task. Under matched
model, harness, and environment configurations, the conditions induce rollout
distributions $\tau_f^u$, $u\in\{P,Q\}$. Writing
$\mathcal B_{f,c}^{u}=\mathcal L(P_c(\tau_f^u))$ for the channel-specific
behavior distribution, we study
\begin{equation}
\Delta_{f,c}=\mathfrak D_f\!\left(
\mathcal B_{f,c}^{P},\mathcal B_{f,c}^{Q}\right),
\qquad c\in\{A,W,F\},
\label{eq:directional_contrast}
\end{equation}
where $\mathfrak D_f$ is the factor-relative directional comparison
operationalized below. Behavioral user control concerns a systematic response
to this intervention, not merely the presence of user information in a prompt
or output. A non-directional observation for $f$ does not imply independence
from all other user information.

\section{Evaluation Construction and Directional Measurement}
\label{sec:measurement}

\begin{figure}[t]
    \centering
    \includegraphics[width=\linewidth]{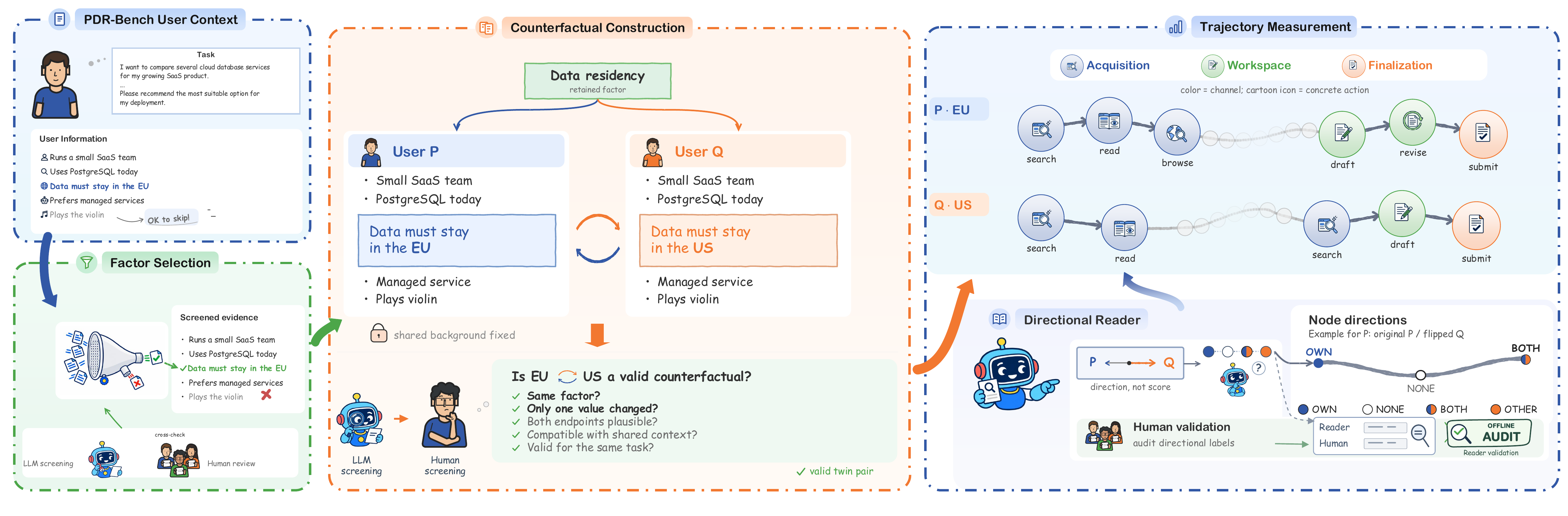}
    \caption{\textbf{Experimental construction and trajectory measurement.}
    We select a task-relevant user factor from PDR-Bench context, construct matched
    $P/Q$ conditions that differ only on that factor, and observe autonomous
    behavior across Acquisition, Workspace, and Finalization. Directional labels and
    targeted human checks are applied.}
    \label{fig:overview}
\end{figure}

We derive experimental materials from PDR-Bench \citep{liang2026pdr},
which provides 50 research tasks across 10 domains and profiles of 25 users.
Our inputs use its Chinese tasks and structured profiles, not its separate
personalized-context files (Appendix~\ref{app:appA-source}). Not every contextual detail is relevant to a given
task. We therefore first construct controlled interventions on task-relevant
user factors (Figure~\ref{fig:overview}), then measure their expression
through local annotation and deterministic aggregation
(Figure~\ref{fig:measurement}).

\subsection{From User Context to Load-Bearing Counterfactuals}
\label{sec:factor_construction}

\paragraph{Evidence grounding and task relevance.}
For each selected task--user pair, we form an evidence pool
$\mathcal E_U=\{e_1,\ldots,e_m\}$, retaining source locations and qualifiers.
A factor is rejected if the evidence can plausibly be true while the
proposed factor is false. Given the task's open delivery decisions
$\mathcal D_T=\{(D_i,\mathcal R_i)\}$, with valid realizations $\mathcal R_i$,
a factor is \emph{load-bearing} if changing it can alter the suitability
ordering of at least one decision. This selects meaningful interventions
without requiring the evaluated agent to respond. The resulting $P/Q$
conditions change only the selected factor and retain all other user
evidence as shared background. Construction is fixed before evaluation
rollouts.

\paragraph{Human screening.}
The author-reported screening protocol involved four Ph.D./master's students
independently reviewing over 200 items each, with retention requiring
unanimous agreement and agreement with the automated assessment. Checks covered
source fidelity, a shared underlying factor, single-factor change, a clear
and plausible contrast, and absence of answer leakage. The retained archive covers an earlier, smaller construction review and
does not independently establish this expanded coverage. These screening
criteria are not an accuracy estimate; Appendix~\ref{app:human_validation}
separates the reported protocol from recovered annotations.

\subsection{Grounded Directional Measurement}
\label{sec:directional_measurement}

For an observation $x$, the factor-specific readout is
$r_f(x)\in\{P,Q,B,N,\bot\}$: the first four labels indicate the $P$ endpoint
only, the $Q$ endpoint only, both, or neither; $\bot$ denotes abstention.
Evaluators receive the task and anonymized endpoints, but not assignment
metadata, paired trajectories, or model/harness identifiers. These are
offline readings: an Acquisition packet can contain other requests,
including later requests, whereas a document read uses that artifact alone
(Table~\ref{tab:appC-visibility} in the appendix). User information may still be inferable
from the observed text.

\paragraph{\textbf{Acquisition requests.}}
For $A$, the observation is an executed search query or read request.
Evaluators read the request rather than its return, measuring the direction
expressed in the information sought, not the evidence received.

\begin{wrapfigure}{r}{0.48\textwidth}
    \centering
    \vspace{-0.5\baselineskip}
    \includegraphics[width=\linewidth]{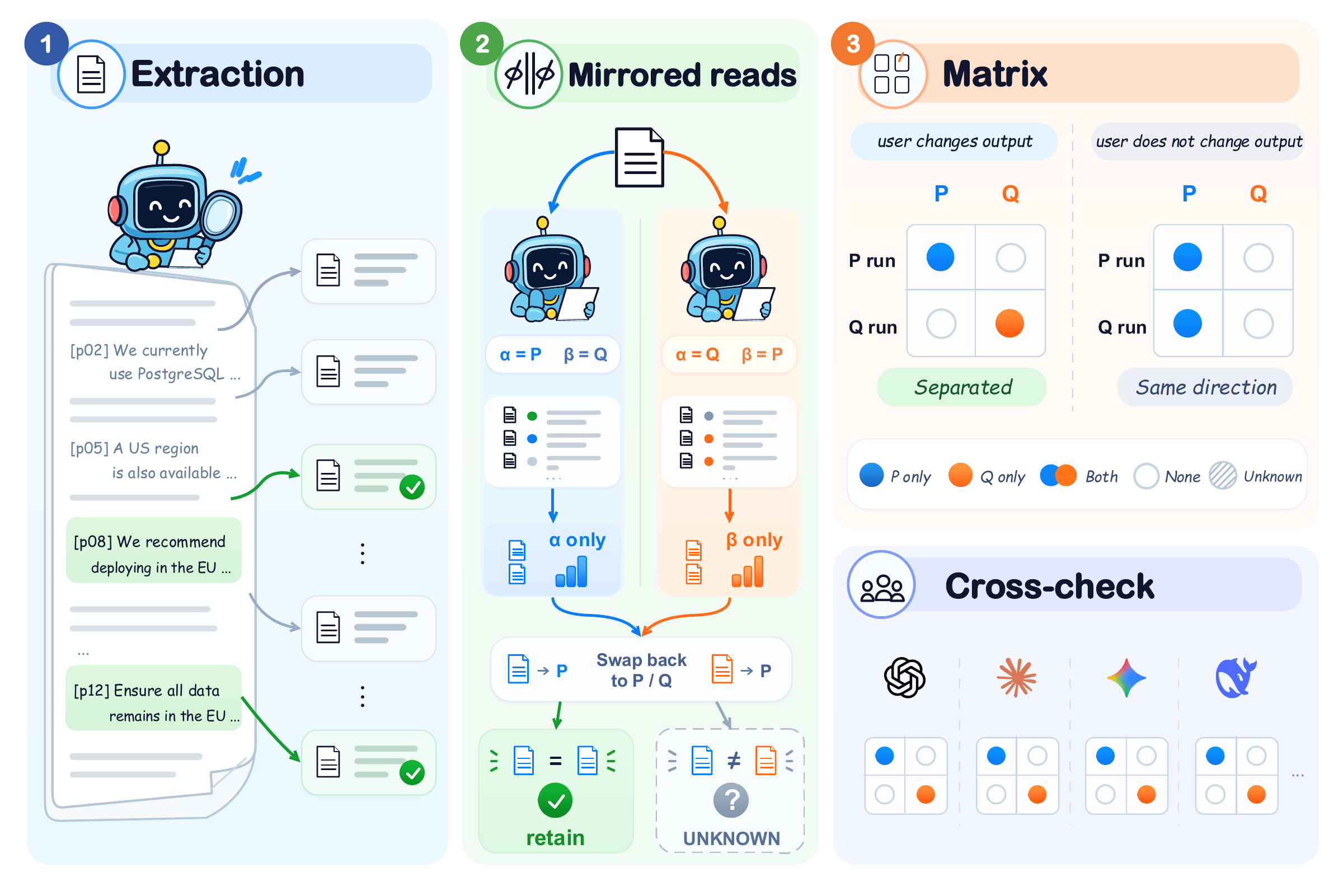}
    \caption{\textbf{Directional measurement.}
    Full-artifact extraction, mirrored reads,
    and counterfactual Matrix construction.}
    \label{fig:measurement}
    \vspace{-0.4\baselineskip}
\end{wrapfigure}

\paragraph{\textbf{Workspace and final artifacts.}}
For $W/F$, each read receives the full artifact with mechanically indexed
paragraphs and extracts source-linked items $\mathcal J_f(a)$
(Figure~\ref{fig:measurement}, left). Items record their endpoint, semantic
type, and recommendation role, separating the agent's own advice from
user restatements and background, and main advice from alternatives,
comparisons, mentions, and rejections. Items are checked against the
schema and source identifiers before aggregation. Only the agent's own
main advice enters $\mathcal J_f^{\mathrm{main}}(a)$.
Code then computes
\begin{equation}
b_v(a)=\mathbf{1}\!\left[
\exists j\in\mathcal J_f^{\mathrm{main}}(a):
j\text{ expresses endpoint }v
\right],
\qquad v\in\{P,Q\}.
\label{eq:artifact_presence}
\end{equation}
The pairs $(1,0),(0,1),(1,1),(0,0)$ encode $P,Q,B,N$, respectively,
subject to the acceptance rule below. This is presence aggregation,
not a vote over item counts. Source references make the annotations
auditable, but do not guarantee exhaustive extraction.

\paragraph{\textbf{Mirrored evaluation.}}
Each evaluator performs two separate reads with the anonymous endpoint
order exchanged (Figure~\ref{fig:measurement}, center).
Labels are restored before comparison: $\alpha$-only under $\alpha=P$
and $\beta$-only under $\beta=P$ both map to $P$.
For artifacts, each pass independently extracts its own items and
aggregates their restored labels. An artifact readout is accepted only
when both passes agree on direction and have an item-based main-advice
basis. Absence of extracted main advice, including restatement-only
outputs, therefore yields $\bot$, not $N$. Disagreement, unresolved
judgments, and invalid reads also remain $\bot$.
Other evaluator-model families repeat the procedure separately to check coverage and directional conclusions, not to vote on a pooled label
(Figure~\ref{fig:measurement}, lower right). Full prompts, schemas,
extraction limits, and acceptance rules appear in
Appendix~\ref{app:measurement_spec}.

\subsection{Matrix, Flow, and Validation}
\label{sec:matrix_flow}

\paragraph{Matrix and Flow.}
The Matrix $M_f(u,b)=Z_f(b;u)$ records directional states at behavioral
loci $b$ under condition $u$. Figure~\ref{fig:measurement} shows the
absolute $P/Q$ coordinates; relative to assignment, these become
\textsc{Own}, \textsc{Other}, \textsc{Both}, or \textsc{None}, with
abstention remaining \textsc{Unknown}. Individual labels describe
behavior; user effects are assessed through the $P/Q$ comparison,
not inferred by the evaluator. Comparisons do not require step-by-step
alignment of tool calls across runs. Within a run, ordered, comparable
states form the Flow $Z_{f,1}\to\cdots\to Z_{f,L}$.
Transitions from \textsc{Own} to \textsc{Both}, \textsc{None}, or
\textsc{Other} denote broadening, dropout, or reversal; re-entry is the
observed return of a previously expressed direction. These are
factor-specific transitions, not changes in global alignment.
Unknown observations never fill gaps or become neutral states.

\paragraph{Directional summaries and scope.}
For identified artifact pairs, define $s(P)=1$, $s(Q)=-1$, and
$s(B)=s(N)=0$. The signed contrast is
\begin{equation}
d_i=\frac{s(r_i^P)-s(r_i^Q)}{2},
\qquad d_i\in\{-1,-\tfrac12,0,\tfrac12,1\},
\label{eq:pair_direction}
\end{equation}
with positive values indicating expected directional separation and
$s(\bot)$ undefined. Acquisition instead contrasts within-run mean
request readouts. These summaries do not replace the Matrix:
\textsc{Both} and \textsc{None} remain distinct. We report readable
coverage and fixed-panel missingness bounds with the specified channel
aggregation (Appendix~\ref{app:measurement_spec}). The channels do not
share a calibrated scale of alignment strength, and unchanged coarse
states do not exclude finer semantic changes. Earlier measurement designs
are documented in Appendix~\ref{app:alternative_measurements}.

\paragraph{Human evaluation.}
The reported protocol comprised two $13\times24$-item rounds with four
Ph.D./master's students and a doctoral adjudicator. The supplied numerical
accounting scales proportions from a preserved 13-item audit to this expanded
protocol; it is not a new item-level validation. In the preserved audit,
both principal reviewers matched the four accepted directions ($4/4$ each);
pre-gate agreement among abstained items was $77.8\%$, and principal-reviewer
agreement was $61.5\%$. The reported absence of opposite-direction assignments
concerns those principal-reviewer disagreements, not every adjudication.
Section~\ref{sec:exp-reliability} retains the expanded accounting, while
Appendix~\ref{app:human_validation} presents the recoverable original records.

\section{Experiments}
\label{sec:experiments}

\subsection{Experimental Setup}
\label{sec:exp-setup}

We evaluate five agent-model families on an author-reconciled roster of
24 PDR-Bench tasks~\citep{liang2026pdr}; coverage is in
Table~\ref{tab:setup}. The main study reports 272 runs and a 110-pair
directional panel. Pairs differ in one user factor; agents receive no further
human guidance. Three evaluator families independently read saved behavior.
Re-evaluation does not add runs. Appendices~\ref{app:harness}
and~\ref{app:additional_results} preserve earlier analyzed subsets separately
from the expanded roster.

\begin{table}[t]
\centering
\caption{\textbf{Experimental setup and coverage.} Author-reconciled study totals;
completed/started runs count agent executions, not repeated evaluations.
Each study and analysis subset retains its own denominator.}
\label{tab:setup}
\small
\setlength{\tabcolsep}{3pt}
\renewcommand{\arraystretch}{1.08}
\begin{tabularx}{\linewidth}{@{}>{\raggedright\arraybackslash}p{1.16in}r>{\raggedright\arraybackslash}X>{\raggedright\arraybackslash}p{0.87in}rr@{}}
\toprule
\multicolumn{6}{@{}l}{\textit{Agent runs}}\\
\textbf{Study} & \textbf{Tasks} & \textbf{Models / setting} & \textbf{Interface} & \textbf{Pairs} & \textbf{Runs}\\
\midrule
Initial randomized & 10 & DeepSeek / off & Report tool & 60 & 180/180\\
Exploratory & 8 & DeepSeek / off & 4 setups & 64 & 127/128\\
Main & 24 & Five families / lowest & Report tool & 110 & 272/272\\
Reasoning follow-up & 24 & Five families / high & Report tool & 88 & 175/176\\
Factor follow-up & 12 & Five families / high & Report tool & 128 & 236/256\\
Harness comparison & 24 & Five families / lowest & H-R, H-W, H-D & 360 & 701/720\\
\bottomrule
\end{tabularx}
\par\smallskip
\begin{tabularx}{\linewidth}{@{}>{\raggedright\arraybackslash}p{1.00in}>{\raggedright\arraybackslash}p{1.62in}>{\raggedright\arraybackslash}X>{\raggedleft\arraybackslash}p{0.58in}@{}}
\toprule
\multicolumn{4}{@{}l}{\textit{Evaluation}}\\
\textbf{Role} & \textbf{Model / annotators} & \textbf{Scope} & \textbf{Reads / item}\\
\midrule
Primary & deepseek-flash / off & Main, follow-ups, harness & 2\\
Auxiliary & gpt-5.4; claude-sonnet-5 & Cross-family checks & 2 each\\
Exploratory & deepseek-v4.1-flash & Exploratory study & 2\\
Human protocol & 4 reviewers + 1 adjudicator & $13\times24=312$ targeted items per round\textsuperscript{\dag} & 1 each\\
\bottomrule
\end{tabularx}
\par\vspace{3pt}
{\footnotesize\raggedright
Five recorded model families: gpt-5.4, claude-sonnet-5, gemini-3.8-flash,
grok-4.6, deepseek-v4-pro. ``Off'', ``lowest'', and ``high'' denote requested
reasoning settings, not matched compute. The six studies report
1,691 completed executions from 1,732 starts; they are not pooled.
\textsuperscript{\dag}Expanded human counts use scaled early-audit proportions
(Table~\ref{tab:human-evaluation}); recovered records are reported separately.
\par}
\end{table}

For each channel $c$ and evaluator, directional contrasts are averaged
within tasks and then across tasks:
\begin{equation}
 d_i^c=\frac{y_{i,P}^c-y_{i,Q}^c}{2},\qquad
 \widehat D^c=\frac{1}{|\mathcal T_c|}\sum_{t\in\mathcal T_c}
 \frac{1}{|\mathcal I_{t,c}|}\sum_{i\in\mathcal I_{t,c}}d_i^c .
 \label{eq:exp-direction}
\end{equation}
Here $y^c$ is the channel-specific score, $\mathcal I_{t,c}$ the readable
pairs in task $t$, and $\mathcal T_c$ the contributing tasks. Unknowns are
not zero; BOTH and NONE remain distinct. Confidence intervals and fixed-plan
missingness ranges answer different questions (Appendix~\ref{app:statistics}).

\subsection{Effects of User Information}
\label{sec:exp-user-information}

\begin{figure}[t]
\centering
\includegraphics[width=\linewidth]{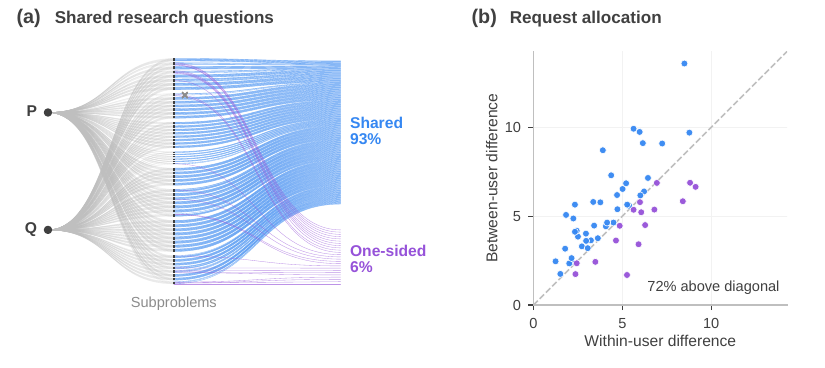}
\caption{
\textbf{Research questions and request allocation.}
(a) Shared and one-sided subproblem visits under paired user conditions.
(b) Between-user versus within-user differences in request allocation.
The dashed line marks equality.
}
\label{fig:research-allocation}
\end{figure}

\noindent\textbf{Shared research questions can lead to different recommendations.}
An initial randomized study establishes an effect of user information on
delivery ($T=0.34$, $p=6.3\times10^{-4}$).
In the exploratory analysis, both user conditions visit the same broad
subproblem in 363/390 comparisons.
Cross-user differences in request allocation exceed differences between
repeated runs under the same condition in 41/57 comparable subproblems
(Fig.~\ref{fig:research-allocation}).
Thus, user information can change how research is distributed without
necessarily replacing its broad agenda.

In the 110-pair primary panel, task-equal contrasts are
$\widehat D^A=0.15$ (95\% CI $[0.08,0.24]$) and
$\widehat D^F=0.83$ ($[0.72,0.92]$), with 79.1\% and 62.7\% readable,
respectively. The same 55 common-readable pairs give 0.17 and 0.83
(50.0\% of the plan). Among $d^A=0$ pairs, 59.6\% have readable Final
comparisons, and 27/28 of these (96.4\%) separate in the assigned direction.
These primary-panel outcomes are not rescaled to the expanded roster;
channel scores do not share an alignment-strength scale. Model-level values
are retained in Appendix~\ref{app:additional_results}.

\begin{figure}[t]
\centering
\includegraphics[width=0.70\linewidth]{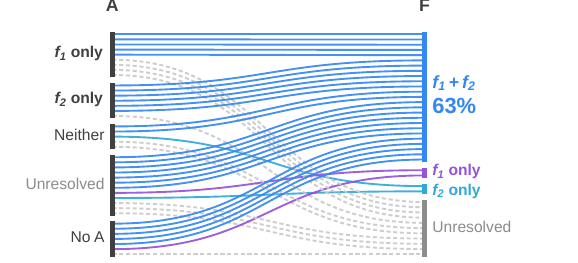}
\caption{
\textbf{User factors across requests and final reports.}
Each line represents one run.
A summarizes expression across all requests;
F shows expression in the final recommendations.
$f_1{+}f_2$ denotes both assigned factors.
Dashed lines indicate unresolved combinations; percentages use all runs.
}
\label{fig:user-factors}
\end{figure}

\medskip
\noindent\textbf{Final reports can combine factors that are not jointly expressed in requests.}
In the primary two-factor analysis, 0/266 requests express both assigned
factors, compared with 25/29 readable final reports
(Fig.~\ref{fig:user-factors}).
Additional profiles show the same qualitative pattern
(2/275 requests; 26/32 readable final reports;
Fig.~\ref{fig:user-factors-extension}).
Requests and reports are different units, so this is not a growth rate.
The factor follow-up's expanded coverage is listed in Table~\ref{tab:setup}.

\medskip
\noindent\textbf{Searching for an alternative does not mean recommending it.}
Documented cases include requests associated with the opposite endpoint
followed by main recommendations for the assigned endpoint
(Table~\ref{tab:supp-search-cases}).
These are existence examples, not evidence that opposite-side exploration
is common. Transition denominators count eligible opportunities, not runs.

\subsection{Analysis of Drafts and Final Reports}
\label{sec:exp-drafts}

\noindent\textbf{Reports are usually saved near the end of research.}
Among main-study runs with both retrieval and a saved draft, 84/86 save their
first draft after the last retrieval action. This is the eligible timing subset, not all runs; other timing results and
notes are in Table~\ref{tab:appF-process} in the appendix.

\begin{figure}[t]
\centering
\includegraphics[width=\linewidth]{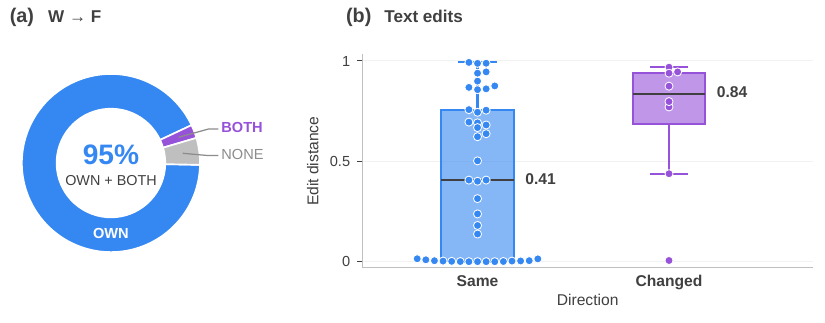}
\caption{
\textbf{Changes from draft to final report.}
(a) Final states of readable, text-different comparisons starting from
\textsc{Own}; retention includes \textsc{Both}.
(b) Edit distance for unchanged and changed directional states.
Dots show run--factor comparisons; labels mark medians.
}
\label{fig:draft-final}
\end{figure}

\medskip
\noindent\textbf{Recommendations often retain their direction after rewriting.}
For the main-study draft-to-final analysis, we compare the last saved draft
with the Final report after excluding identical texts and unresolved labels.
Among 41 readable, text-different comparisons beginning from
\textsc{Own}, 39 retain the assigned side and 38 remain
\textsc{Own} (Fig.~\ref{fig:draft-final}).
Among the 38 \textsc{Own}$\rightarrow$\textsc{Own} comparisons,
18/38 have normalized content-token edit distance of at least $0.5$.
Across readable text-different comparisons, median edit distance is 0.41 when
the directional state is unchanged and 0.84 when it changes.
Thus, large edits can preserve direction, although direction-changing
revisions tend to involve larger edits; they are not necessarily cosmetic.

\subsection{Effects of Execution Settings}
\label{sec:exp-settings}

\noindent\textbf{Different paths can reach the same recommendation direction.}
Same-task examples differ in decisions, requests, and public drafting while
reaching the same coarse Final direction (Table~\ref{tab:supp-paths}); this
does not establish equal quality or efficiency. The reasoning follow-up
reports 175/176 completed runs over 88 planned pairs. Its available matched
analyses show longer execution without a consistent directional improvement;
GPT also changes API protocol, limiting attribution to reasoning alone.

The harness comparison reports 360 planned pairs and 701/720 completed runs:
120 pairs and 240 planned runs per interface. Directional comparisons in
the archived analyzed subset remain positive across interfaces, but neither
equivalence nor delegated execution is established. Delegation was available
but unused there, and two file-based interfaces submit the saved report
itself, making draft--Final identity structural
(Appendix~\ref{app:additional_results}). These subset findings are not
extrapolated to the expanded execution counts.

\subsection{Reliability of Directional Evaluation}
\label{sec:exp-reliability}

\noindent\textbf{Agreement depends on the comparison level.}
Evaluator agreement is computed on common-readable documents or pairs, not
agent-run totals. The archived comparisons distinguish report-label,
exact-value, and pair-sign agreement; agreement on signs does not establish
semantic ground truth (Appendix~\ref{app:additional_results}).

\medskip
\noindent\textbf{Human reference and expanded accounting.}
The author-reported protocol uses four Ph.D./master's students and a doctoral
adjudicator in two $13\times24$-item rounds: 312 items per round and
624 item-round observations. As specified in the supplied source, the
expanded counts apply preserved 13-item proportions, giving 96 accepted
and 216 abstained items, $192/312=61.5\%$ reviewer agreement,
$120/312=38.5\%$ disagreement, and $168/216=77.8\%$ pre-gate agreement
among abstentions (Table~\ref{tab:human-evaluation}). The absence of
opposite assignments describes the preserved principal-reviewer pattern.
These scaled counts do not add independent annotation evidence; the actual
recovered audit and its limits are in Appendix~\ref{app:human_validation}.

\begin{table}[t]
\centering
\caption{\textbf{Updated human-evaluation accounting.}
Counts apply the preserved 13-item proportions to the
$13\times24=312$-item evaluation set.}
\label{tab:human-evaluation}
\small
\setlength{\tabcolsep}{4pt}
\begin{tabular*}{\linewidth}{@{\extracolsep{\fill}}l r r@{}}
\toprule
\textsf{\textbf{Quantity}} &
\textsf{\textbf{Count}} &
\textsf{\textbf{Rate}}\\
\midrule
Items per round                     & 312 & 100.0\%\\
Instrument accepted                 & 96  & 30.8\%\\
Instrument abstained                & 216 & 69.2\%\\
Principal-reviewer exact agreement  & 192 & 61.5\%\\
Principal-reviewer disagreement     & 120 & 38.5\%\\
Pre-gate agreement among abstentions& 168/216 & 77.8\%\\
\bottomrule
\end{tabular*}
\par\vspace{4pt}
{\footnotesize\raggedright
Two rounds contain 624 item-round observations in total.
Counts retain the source's proportional scaling and are not independently
recovered item-level judgments or a corpus-wide accuracy estimate.
\par}
\end{table}

\section{Limitations}
\label{sec:limitations}
\paragraph{Measurement resolution.}
We primarily identify whether a selected user factor is expressed, not how
far behavior deviates from the user's needs. Coarse endpoint changes are
visible, but altered priorities, omitted qualifications, factual errors, and
subtle trade-offs may preserve the label. Absence from a request also need
not mean forgetting. Future work will measure recommendation strength,
constraint satisfaction, and consequential deviations at finer resolution,
with explicit uncertainty and independent validation.

\paragraph{Observation and validation.}
Public artifacts do not expose internal attention or reasoning. Extraction
can omit content, and offline request packets can include later requests.
Abstention limits coverage; evaluator agreement is not accuracy. Recovered
human records cover early targeted audits; expanded human totals use
proportional accounting rather than additional independently recovered labels.

\paragraph{Scope.}
The selected Chinese tasks, constructed factors, model routes, and shared
retrieval backend limit generalization. Many low-effort runs are short;
follow-ups reuse tasks, and the archived subsets do not fully cover expanded
execution totals. API changes, unused delegation, and same-file submission
restrict individual comparisons. The findings do not identify causal
stabilization, harness equivalence, or benefits of removing human interaction
(Appendices~\ref{app:alternative_measurements} and~\ref{app:limitations}).

\section{Conclusion}
\label{sec:conclusion}
We studied autonomous deep research after users provide their task and
background. DRaligned combines controlled user-information changes with
source-grounded directional readings of requests, drafts, and final reports.
A largely shared research agenda can accompany user-specific recommendations;
reports can integrate factors not jointly visible in requests; and
recommendation direction can persist through substantial rewriting.
Different observable paths can reach the same coarse outcome in the tested
settings. These findings distinguish user influence from report quality alone
and from the assumption that every request must express user priorities.
The framework supports finer, auditable investigation of autonomous behavior
and its relation to delivery.

\clearpage
\section*{AI Use Statement}
AI assistants were used in developing research questions and experimental
plans, implementing and checking analysis code, reviewing outputs,
translating source material, and drafting or editing text, figures, and
LaTeX. Evaluator-model use within the experiments is specified separately
in Section~\ref{sec:measurement}. The authors are responsible for the
submitted claims and citations. AI-assisted recomputation is not described
as independent human validation; unresolved source limitations are disclosed.

\section*{Reproducibility Statement}
The supplementary material provides the prompts, tool schemas, aggregation
rules, worked examples, and records underlying the reported analyses
(Appendices~\ref{app:harness}--\ref{app:reproducibility}). Tables distinguish
execution totals, analyzed subsets, and recovered human records. Original
source files and translated excerpts are linked through manifests; unavailable
records and historical corrections are documented rather than reconstructed.

\bibliography{references_fixed}
\bibliographystyle{iclr2027_conference}
\clearpage
\appendix

\section{Tasks and User Conditions}
\label{app:factor_construction}
\paragraph{Archive scope.}
The records below document the construction rules, prompts, worked examples, and archived qualitative analyses.
Archived task, user-profile, model-roster, and execution-scale counts are intentionally omitted in this version because those roster totals were reconciled separately in the main paper.
The main-text experimental setup is the authoritative source for current study scale.

\noindent\textit{Language note.} Chinese source material throughout this appendix (prompts, user messages, model outputs, reviewer materials and report excerpts) is shown in English translation prepared for this appendix; the untranslated originals are supplied with their hashes (Appendix~\ref{app:reproducibility}).

This appendix documents the source data, how one user fact became a pair of user conditions, which conditions were run, and real construction decisions.
Construction and evaluation prompts are in Appendix~\ref{app:measurement_spec}; the human review of constructed factors is in Appendix~\ref{app:human_validation}.
Unless stated otherwise, construction and checking were carried out by language models under written rules.
In the archived material, recorded human screening of individual candidates is available only for the initial randomized study. The expanded protocol described in Section~\ref{sec:measurement} has no corresponding item-level archive in this package; other archived study factors were checked by models.

\subsection{Source Data}
\label{app:appA-source}

All studies use the Chinese task and persona files of PDR-Bench~\citep{liang2026pdr} from the archived repository revision.
The query mapping is used only to identify which source users are paired with a task; conversation histories, English source files, and the PDR evaluation criteria are not supplied to constructors or agents.
PDR-Bench's source participants belong to the source dataset and are distinct from reviewers or annotators in this work.
Tasks exposed in earlier development studies are excluded from the main-study roster before construction.
In the main study, the agent receives one user message containing the task text, a shared user background, and the endpoint sentence(s), separated by line breaks (Appendix~\ref{app:harness}).
Recorded attempts use this frozen task-and-user message.
In the exploratory study, the message also began with a research reference date; any task-specific date clarification was applied identically across the paired conditions.

\begin{figure}[tb]
\centering
\begin{taskpanel}{Deep Research Task (PDR-Bench task 16, Travel)}
I plan to take a two-week backpacking trip within the next three months, with the goal of exploring in depth the diverse cultures and natural wonders of Southeast Asia, such as tropical rainforests, historical sites and local communities. Please tailor an itinerary for me, including recommended destinations, modes of transportation, accommodation options, local specialty food, as well as safety advice, cost estimates and tips on local customs.
\end{taskpanel}
\vspace{1mm}
\begin{tcbraster}[raster columns=2,raster equal height,raster column skip=3mm]
\begin{userpanel}[P]{User condition P}
\iconrow[apxOrange]{\faUserGraduate}{I am an undergraduate computer science student, and I live in Suzhou, Jiangsu.}
\iconrow[apxOrange]{\faWallet}{My income comes mainly from my parents' support and scholarships[.\allowbreak{}.\allowbreak{}.\allowbreak{}]}
\iconrow[apxOrange]{\faBed}{When I travel I usually stay in budget hotels or youth hostels; I value cost-effectiveness, and I also care about whether the place is clean.}
\iconrow[apxOrange]{\faLandmark}{When traveling I'm used to independent travel and planning my own itinerary, and I'm quite interested in historical and cultural sites, museums and science and technology exhibitions[.\allowbreak{}.\allowbreak{}.\allowbreak{}]}
\iconrow[apxHlP]{\faMale}{\hlP{Also, I am male.}}
\end{userpanel}
\begin{userpanel}[Q]{User condition Q}
\iconrow[apxOrange]{\faUserGraduate}{I am an undergraduate computer science student, and I live in Suzhou, Jiangsu.}
\iconrow[apxOrange]{\faWallet}{My income comes mainly from my parents' support and scholarships[.\allowbreak{}.\allowbreak{}.\allowbreak{}]}
\iconrow[apxOrange]{\faBed}{When I travel I usually stay in budget hotels or youth hostels; I value cost-effectiveness, and I also care about whether the place is clean.}
\iconrow[apxOrange]{\faLandmark}{When traveling I'm used to independent travel and planning my own itinerary, and I'm quite interested in historical and cultural sites, museums and science and technology exhibitions[.\allowbreak{}.\allowbreak{}.\allowbreak{}]}
\iconrow[apxHlQ]{\faFemale}{\hlQ{Also, I am female.}}
\end{userpanel}
\end{tcbraster}
\vspace{1mm}
\begin{dashedgroup}
\begin{CJK}{UTF8}{gbsn}\footnotesize\sffamily
\textbf{Left out of both inputs} (\texttt{shared\_background\_exclusions}, verbatim)
\iconrow[apxGrayFrame]{\faBan}{No name is written, since a name may carry gender associations.}
\iconrow[apxGrayFrame]{\faBan}{Sports hobbies such as playing basketball every week or learning yoga off campus are not written: in common perception these carry gender associations and would make one endpoint seem more natural.}
\iconrow[apxGrayFrame]{\faBan}{``Recently started planning to travel alone'' is not written, since that is exactly the second factor itself.}
\end{CJK}
\end{dashedgroup}
\caption{\textbf{User conditions for one main-study task.} Top: the task text as sent (English translation). Panels: four sentences of the shared background as sent ([...] marks elided text; full messages in the released sources) and the highlighted endpoint sentence (\emph{translation:} P ``Also, I am male.'', Q ``Also, I am female.''). The two user messages differ only in this last line. Black rows are identical in both conditions and include facts considered but not selected as a factor, such as the home city. Bottom: persona information the constructor left out of both inputs; the third entry is the source of this task's second factor, which was built but not selected (Table~\ref{tab:appA-decisions}).}
\label{fig:appA-case}
\end{figure}

\subsection{From User Information to Tested Factors}
\label{app:appA-procedure}

A factor is one piece of user information with two endpoint sentences: $P$ states the user's recorded situation and $Q$ is a minimal counterfactual that changes only that piece, so that $U^P=C\oplus z^P$ and $U^Q=C\oplus z^Q$ share the background $C$ (Section~\ref{sec:counterfactual_control}).
The studies used different construction routes under the same counterfactual-construction principles; archived route-size totals are omitted here.

\refstepcounter{table}\label{tab:appA-routes}

\paragraph{Main-study construction.}
The PDR users paired with each task were ordered by a fixed hash, and the constructor used the first eligible record unless it offered no valid factor.
Each constructor saw only the task text, these persona records and the written guide, never other tasks' factors, trajectories or results.
Scanning the persona fields in stored order, it took the first task-relevant statement meeting seven rules (box below), wrote $P$ in the first person and $Q$ as the minimal counterfactual, and recorded a verbatim source statement of at least six characters with its field path.
The shared background (150--400 characters, first person) came from the same record and had to omit the factor and anything coupled with it; deliberate omissions are listed in \texttt{shared\_background\_exclusions}.
Each factor also carries evaluation rubrics for requests and report items (Appendix~\ref{app:measurement_spec}).
A mechanical validator checked the fields, that the source statement occurs verbatim in the persona record, that each profile equals the background followed by the endpoint text, and that factor identifiers are neutral.

\begin{promptbox}{Rules for the background and the factor in the construction guide (English translation)}
-\allowbreak{} First person, 150–400 characters, written as what this user would say to a research assistant.\\
-\allowbreak{} The content is drawn from the **task-relevant** real information in this user's persona (identity, circumstances, resources, existing arrangements, etc.); it may be rewritten in colloquial language, but facts not present in the persona must not be invented. A neutral transitional sentence is allowed only when necessary for coherence (for example, ``Finally, let me add one more thing...'').\\
-\allowbreak{} **It contains neither the focal factor itself nor any information coupled with the focal factor that would make one endpoint appear more reasonable or more urgent**. Content that is deliberately excluded is written into `shared\_\allowbreak{}background\_\allowbreak{}exclusions`, with each entry explaining why it was excluded.\\
{}[.\allowbreak{}.\allowbreak{}.\allowbreak{}]\\
**How to choose**: following the order of the persona fields, take the first piece of task-relevant information that satisfies all seven conditions below at the same time. Do not go and pick the one where ``the effect is easiest to see''.
\par
1.\allowbreak{} This information has real, consequential relevance to the task (it would change how a good plan should be written);\\
2.\allowbreak{} Both endpoints are reasonable and could both occur in reality;\\
3.\allowbreak{} The original task text does not force either endpoint;\\
4.\allowbreak{} After switching to the other endpoint, this user remains coherent as a whole;\\
5.\allowbreak{} It is a minimal counterfactual: change only this one item and leave everything else verbatim unchanged;\\
6.\allowbreak{} Neither endpoint directly contains the answer;\\
7.\allowbreak{} It can coexist with the shared background.
\par
**Two endpoints**: `P\_\allowbreak{}endpoint\_\allowbreak{}text` is the persona's original state, written in the first person; `Q\_\allowbreak{}endpoint\_\allowbreak{}text` is the minimal counterfactual.\\
The two endpoints are close in length and consistent in tone, and differ only in this one piece of information.
\end{promptbox}

Independent AI reviewers then saw the task text, the background, and the two endpoints (as anonymous labels in hash order, without rubrics) and checked the frozen fatal-problem categories:
\texttt{contradiction}, \texttt{endpoint\_\allowbreak illegal\_\allowbreak or\_\allowbreak implausible}, \texttt{task\_\allowbreak forces\_\allowbreak one\_\allowbreak endpoint}, \texttt{endpoint\_\allowbreak states\_\allowbreak the\_\allowbreak answer} and \texttt{more\_\allowbreak than\_\allowbreak one\_\allowbreak thing\_\allowbreak changes}.
Under a rule fixed beforehand, a factor was excluded only if both reviewers flagged the same category.
The archived review record contains a split flag illustrated in Figure~\ref{fig:appA-case-split}; factors were frozen before agent execution.
The reviewers did not judge whether a factor is load-bearing (rule~1 was the constructor's judgment), and their review is not a human validation.

\paragraph{What reaches the agent.}
Persona information falls into three groups.
(i)~The tested factor: its endpoint sentence is the only text that differs between conditions.
(ii)~Shared background: identical in both conditions, including facts that were considered but not selected as the factor, such as the home city in t16 (Figure~\ref{fig:appA-case}).
(iii)~Omitted information: absent from both inputs, either because it is coupled with the factor (\texttt{shared\_background\_exclusions}) or because the background has no room for it.
Not selecting a statement as the factor therefore does not remove it from the input; only group~(iii) is absent.

\subsection{Main-Study Factors}
\label{app:appA-factors}
The exact archived task and factor roster is omitted from this count-free appendix.
Construction rules and representative accepted, rejected, and disputed factor examples are retained below; the current experimental roster is reported only in the main-text setup.
\refstepcounter{table}\label{tab:appA-factors}

\subsection{Two-Factor Tasks}
\label{app:appA-two-factor}

The two-factor design combines the endpoints of factor~1 and factor~2 with one shared background, yielding profiles PP, PQ, QP and QQ.
For example, one task contrasts living with parents versus living alone and, separately, two dog breeds.
The primary comparison changes factor~1 while holding factor~2 at its recorded endpoint; complementary profiles hold factor~2 at its counterfactual endpoint.
Table~\ref{tab:appA-two-factor} states which factor changes in each comparison.
Figure~\ref{fig:user-factors} uses the primary comparison, and Figure~\ref{fig:user-factors-extension} shows the complementary profiles.

\begin{table}[tb]
\centering
\caption{\textbf{Comparisons in a two-factor task} (all four are defined in the derivation code).}
\label{tab:appA-two-factor}
\footnotesize
\begin{tabular}{@{}llll@{}}
\toprule
\textsf{\textbf{Comparison}} & \textsf{\textbf{Factor changed}} & \textsf{\textbf{Factor held fixed}} & \textsf{\textbf{Use}}\\
\midrule
PP vs.\ QP & factor 1 & factor 2 at $P$ & primary pair (main study and follow-ups)\\
PQ vs.\ QQ & factor 1 & factor 2 at $Q$ & additional profiles (main study)\\
PP vs.\ PQ & factor 2 & factor 1 at $P$ & main study; not in the main-text estimates\\
QP vs.\ QQ & factor 2 & factor 1 at $Q$ & main study; not in the main-text estimates\\
\bottomrule
\end{tabular}
\end{table}

\subsection{Examples of Construction Decisions}
\label{app:appA-examples}

\begin{figure}[tb]
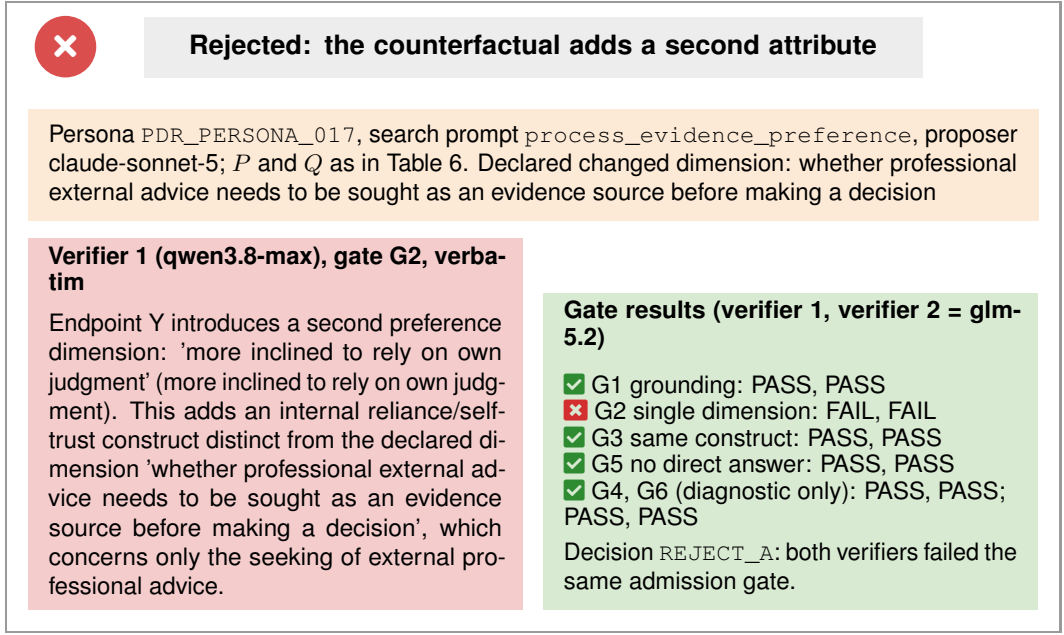

\begin{casebox}{\casefail}{Rejected: the counterfactual adds a second attribute}
\begin{caseband}Persona \texttt{PDR\_PERSONA\_017}, search prompt \texttt{process\_evidence\_preference}, proposer claude-sonnet-5; $P$ and $Q$ as in Table~\ref{tab:appA-decisions}. Declared changed dimension: whether professional external advice needs to be sought as an evidence source before making a decision\end{caseband}
\casecols{\begin{casepanel}[apxPink]{Verifier 1 (qwen3.8-max), gate G2, verbatim}{\fontencoding{T1}\selectfont Endpoint Y introduces a second preference dimension:\allowbreak{} 'more inclined to rely on own judgment' (more inclined to rely on own judgment).\allowbreak{} This adds an internal reliance/\allowbreak{}self-\allowbreak{}trust construct distinct from the declared dimension 'whether professional external advice needs to be sought as an evidence source before making a decision',\allowbreak{} which concerns only the seeking of external professional advice.\allowbreak{}}\end{casepanel}}{\begin{casepanel}[apxGreen]{Gate results (verifier 1, verifier 2 = glm-5.2)}
\ok\ G1 grounding: PASS, PASS\\
\no\ G2 single dimension: FAIL, FAIL\\
\ok\ G3 same construct: PASS, PASS\\
\ok\ G5 no direct answer: PASS, PASS\\
\ok\ G4, G6 (diagnostic only): PASS, PASS; PASS, PASS\\[2pt]
Decision \texttt{REJECT\_A}: both verifiers failed the same admission gate.\end{casepanel}}
\end{casebox}
\caption{\textbf{Rejected candidate (construction pipeline of the initial randomized study).} Both verifiers failed the single-dimension gate because the counterfactual adds a second preference (\emph{translation:} ``prefers to rely on own judgment). The verifier text is verbatim; the stored record of verifier 2 makes the same point.}
\label{fig:appA-case-rejected}
\end{figure}

\begin{figure}[tb]
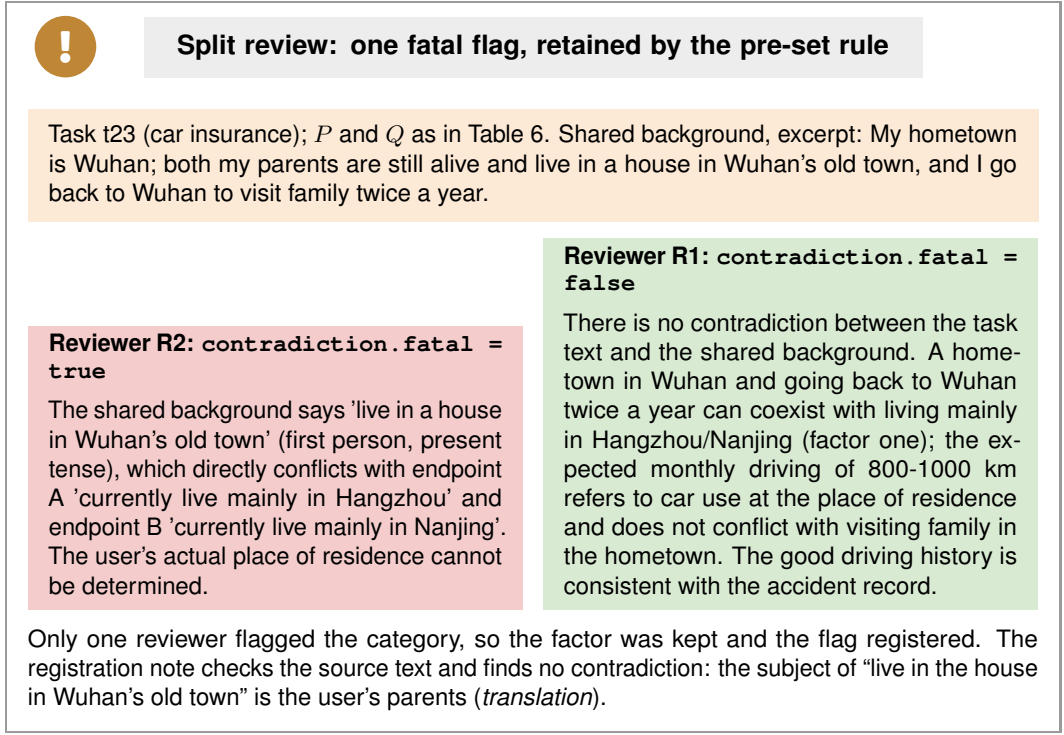

\begin{casebox}{\casewarn}{Split review: one fatal flag, retained by the pre-set rule}
\begin{caseband}Task t23 (car insurance); $P$ and $Q$ as in Table~\ref{tab:appA-decisions}. Shared background, excerpt: My hometown is Wuhan; both my parents are still alive and live in a house in Wuhan's old town, and I go back to Wuhan to visit family twice a year.\end{caseband}
\casecols{\begin{casepanel}[apxPink]{Reviewer R2: \texttt{contradiction.fatal = true}}{\fontencoding{T1}\selectfont The shared background says 'live in a house in Wuhan's old town' (first person, present tense), which directly conflicts with endpoint A 'currently live mainly in Hangzhou' and endpoint B 'currently live mainly in Nanjing'. The user's actual place of residence cannot be determined.}\end{casepanel}}{\begin{casepanel}[apxGreen]{Reviewer R1: \texttt{contradiction.fatal = false}}{\fontencoding{T1}\selectfont There is no contradiction between the task text and the shared background. A hometown in Wuhan and going back to Wuhan twice a year can coexist with living mainly in Hangzhou/\allowbreak{}Nanjing (factor one); the expected monthly driving of 800-\allowbreak{}1000 km refers to car use at the place of residence and does not conflict with visiting family in the hometown. The good driving history is consistent with the accident record.}\end{casepanel}}
\par\smallskip{\footnotesize\sffamily Only one reviewer flagged the category, so the factor was kept and the flag registered. The registration note checks the source text and finds no contradiction: the subject of ``live in the house in Wuhan's old town'' is the user's parents (\emph{translation}).}
\end{casebox}
\caption{\textbf{Split review in the main study (task t23).} Reviewer answers are verbatim (English translation); R1 and R2 are the two AI reviewers of this task.}
\label{fig:appA-case-split}
\end{figure}

Table~\ref{tab:appA-decisions} shows one real decision of each recorded outcome type with its original endpoint texts; two are expanded in Figures~\ref{fig:appA-case-rejected} and~\ref{fig:appA-case-split}.
They were chosen to illustrate the outcome types and are not a sample of the construction record.
Shading follows the recorded outcome (green accepted, red rejected, yellow split judgment); the unshaded row was never judged negatively but was not selected.

\begin{table}[tb]
\centering
\caption{\textbf{Real construction decisions.} Endpoint texts verbatim (English translation). Outcomes are the recorded decisions of the rules in force, not correctness judgments by the authors.}
\label{tab:appA-decisions}
\footnotesize
\setlength{\tabcolsep}{3pt}
\begin{tabularx}{\linewidth}{@{}>{\raggedright\arraybackslash}p{0.98in}>{\raggedright\arraybackslash}X>{\raggedright\arraybackslash}p{1.42in}>{\raggedright\arraybackslash}p{0.78in}@{}}
\toprule
\textsf{\textbf{Candidate}} & \textsf{\textbf{$P$ / $Q$ endpoints}} & \textsf{\textbf{Recorded checks}} & \textsf{\textbf{Outcome}}\\
\midrule
\rowpass \textsf{\textbf{Accepted:}} main study, t16, persona gender field & $P$: \zh{Also, I am male.}\newline $Q$: \zh{Also, I am female.} & Validator passed; neither reviewer flagged a fatal problem & \PASS\newline used in the archived study\\
\addlinespace[1pt]
\textsf{\textbf{Built, not selected:}} main study, t16 second factor (source \zh{Recently started planning to travel solo}) & $P$: \zh{This time I plan to go by myself; I want to try traveling solo.}\newline $Q$: \zh{This time I plan to go with two friends, traveling together.} & Validator passed; no fatal flag; not selected by the frozen two-factor assignment rule & \badge{apxHlP}{NOT SELECTED}\newline in no run input\\
\addlinespace[1pt]
\rowfail \textsf{\textbf{Rejected:}} initial randomized study pipeline, persona \texttt{PDR\_\allowbreak PERSONA\_\allowbreak 017} & $P$: \zh{Consults professionals for advice before making investment decisions}\newline $Q$: \zh{Does not consult professionals for advice before making investment decisions, preferring to rely on own judgment} & Both LLM verifiers failed gate G2 (single dimension) & \FAIL\newline not sent to human review\\
\addlinespace[1pt]
\rowunclear \textsf{\textbf{Split review:}} main study, t23 & $P$: \zh{I currently live in Hangzhou, in a rented studio apartment in the city center, fairly close to my company.}\newline $Q$: \zh{I currently live in Nanjing, in a rented studio apartment in the city center, fairly close to my company.} & One reviewer flagged a contradiction, the other did not; exclusion required both & \badge{apxUnclearFg}{RETAINED}\newline flag registered\\
\addlinespace[1pt]
\rowunclear \textsf{\textbf{Disputed:}} factor follow-up, t46\_x1 & $P$: \zh{On spending: we are quite willing to invest in our child's education, and generally don't care much about price when signing up for classes or buying learning resources.}\newline $Q$: \zh{On spending: we hope to save as much as possible on our child's education, and generally pick the cheaper options first when signing up for classes or buying learning resources.} & Designed as a soft preference; one validator judged the endpoints mutually exclusive, the other agreed with the design & \badge{apxUnclearFg}{DISPUTED}\newline run, not in the primary factor contrast\\
\bottomrule
\end{tabularx}
\end{table}

\section{Execution and Harness Details}
\label{app:harness}

This appendix records how the agent runs were executed.
It keeps three things apart: the harness formalism and its canonical basis as defined in the project's frozen research foundation (Appendix~\ref{app:harness-basis}); the implementations that were actually run for the studies in Table~\ref{tab:setup} (Appendix~\ref{app:harness-impl}); and the conformance checks that were recorded for them (Appendix~\ref{app:harness-checks}).
The verbatim system prompts and tool schemas are given in Appendix~\ref{app:measurement_spec}; here we list tool names and argument fields only.
Code field names are kept as recorded.
Chinese source material is shown in English translation prepared for this appendix; the untranslated originals are supplied with their hashes (Appendix~\ref{app:reproducibility}).

\subsection{Harness formalism and canonical basis}
\label{app:harness-basis}

The frozen research foundation defines a harness as $\mathcal H=(\mathcal R,\mathcal B,\Lambda,\Sigma,\Gamma)$.
$\mathcal R=(R_A,R_W,R_F)$ are the mechanisms that realize acquisition, workspace and finalization.
$\mathcal B=(B_{AW},B_{WF})$ are the two boundary operators; each is either \emph{coupled} or an \emph{explicit typed handoff}.
$\Lambda$ is an access matrix, $\Lambda(\text{actor},\text{state})\in\{\text{none},\text{read},\text{append},\text{write},\text{overwrite},\text{spawn}\}$, stating what each agent, tool or sub-agent may do to the user information $U$, acquired information $I_t$, workspace $W_t$ and delivery state $D_t$.
$\Sigma$ is the execution topology (single agent; sub-agent spawn, delegation and return; coordinators; routing and scheduling).
$\Gamma$ is the instrumentation contract: event logging, provenance, artifact versioning, checkpoint/fork/replay, deterministic cache and environment rules, and state restoration.
Sub-agents, web tools, memory, planners and verifiers are mechanisms inside $\mathcal H$, not additional channels.

Crossing the two boundary states gives the canonical basis $\{\mathcal H_{00},\mathcal H_{10},\mathcal H_{01},\mathcal H_{11}\}$; the first index is $B_{AW}$ and the second $B_{WF}$ (0 = coupled, 1 = explicit typed handoff).
The foundation calls this $2\times2$ a representational coordinate system, not the main experimental matrix: $\mathcal H_{00}$ is implemented in full, and the other cells need only minimal adapters with conformance tests.
Table~\ref{tab:appB-basis} gives each cell's frozen contract and run status; the $\mathcal H_{00}$ contract used by the main study and the follow-ups is quoted below.

\begin{table}[t]
\centering
\caption{\textbf{Canonical harness basis: frozen contracts and analysis status.} Roles and tools are shown as registered in each cell's contract. Explicit-handoff cells were implemented in earlier development work but are not analyzed in this paper.}
\label{tab:appB-basis}
\scriptsize
\setlength{\tabcolsep}{3pt}
\renewcommand{\arraystretch}{1.12}
\begin{tabularx}{\linewidth}{@{}l l l >{\raggedright\arraybackslash}X >{\raggedright\arraybackslash}p{1.25in}@{}}
\toprule
\textsf{\textbf{Cell}} & \textsf{\textbf{$B_{AW}$}} & \textsf{\textbf{$B_{WF}$}} & \textsf{\textbf{Roles (tools)}} & \textsf{\textbf{Run status}}\\
\midrule
H00 & coupled & coupled & one persistent executor (\texttt{search}, \texttt{read}, \texttt{write\_report}, \texttt{submit\_report}) & \badge{apxPassFg}{RUN}: main study, both follow-ups, H-R; also the exploratory study's report-tool setups\\
H10 & handoff & coupled & W-controller (\texttt{write\_report}, \texttt{submit\_report}, \texttt{delegate\_acquisition}); a fresh A-realizer per delegation (\texttt{search}, \texttt{read}) & \badge{apxUnclearFg}{NOT ANALYZED}: earlier development study\\
H01 & coupled & handoff & upstream executor (\texttt{search}, \texttt{read}, \texttt{write\_report}, \texttt{handoff\_to\_finalizer}); separate Finalizer (\texttt{submit\_report}, \texttt{request\_upstream}) & \badge{apxUnclearFg}{NOT ANALYZED}: earlier development study\\
H11 & handoff & handoff & W-controller, A-realizer and Finalizer as above; the controller has neither retrieval nor submission & \badge{apxUnclearFg}{NOT ANALYZED}: earlier development study\\
\bottomrule
\end{tabularx}
\end{table}

\begin{jsonbox}{Frozen contract of the canonical harness H00 (English translation; selected keys, other keys omitted [...])}
\{\\
\hspace*{0.5em}\textquotedbl{}harness\_\allowbreak{}cell\textquotedbl{}:\allowbreak{} \textquotedbl{}H00\textquotedbl{},\allowbreak{}\\
\hspace*{0.5em}\textquotedbl{}A\_\allowbreak{}W\_\allowbreak{}boundary\textquotedbl{}:\allowbreak{} \textquotedbl{}COUPLED\textquotedbl{},\allowbreak{}\\
\hspace*{0.5em}\textquotedbl{}W\_\allowbreak{}F\_\allowbreak{}boundary\textquotedbl{}:\allowbreak{} \textquotedbl{}COUPLED\textquotedbl{},\allowbreak{}\\
\hspace*{0.5em}\textquotedbl{}six\_\allowbreak{}mappings\textquotedbl{}:\allowbreak{} \{\\
\hspace*{1.0em}\textquotedbl{}R\textquotedbl{}:\allowbreak{} \textquotedbl{}The three channels A /\allowbreak{} W /\allowbreak{} F are all realized by the same persistent executor in the same context;\allowbreak{} evidence gathering,\allowbreak{} drafts and submission are different tools on the same message chain.\allowbreak{}\textquotedbl{},\allowbreak{}\\
\hspace*{1.0em}\textquotedbl{}B\_\allowbreak{}AW\textquotedbl{}:\allowbreak{} \textquotedbl{}Coupled.\allowbreak{} Raw tool returns enter the same context directly;\allowbreak{} there is no handoff object and no cross-\allowbreak{}context boundary.\allowbreak{}\textquotedbl{},\allowbreak{}\\
\hspace*{1.0em}\textquotedbl{}B\_\allowbreak{}WF\textquotedbl{}:\allowbreak{} \textquotedbl{}Coupled.\allowbreak{} The drafts of write\_\allowbreak{}report and the final of submit\_\allowbreak{}report are produced by the same executor.\allowbreak{}\textquotedbl{},\allowbreak{}\\
\hspace*{1.0em}\textquotedbl{}Lambda\textquotedbl{}:\allowbreak{} \textquotedbl{}The single role can read T/\allowbreak{}U,\allowbreak{} its own full history,\allowbreak{} all raw tool returns and all draft versions;\allowbreak{} it can write drafts and the delivery.\allowbreak{}\textquotedbl{},\allowbreak{}\\
\hspace*{1.0em}\textquotedbl{}Sigma\textquotedbl{}:\allowbreak{} \textquotedbl{}Single process,\allowbreak{} serial,\allowbreak{} single loop;\allowbreak{} no sub-\allowbreak{}context.\allowbreak{}\textquotedbl{},\allowbreak{}\\
\hspace*{1.0em}\textquotedbl{}Gamma\textquotedbl{}:\allowbreak{} \textquotedbl{}Reuses the nd\_\allowbreak{}harness event stream (assistant /\allowbreak{} tool\_\allowbreak{}call /\allowbreak{} provider\_\allowbreak{}error) and the artifacts version chain;\allowbreak{} records model\_\allowbreak{}exposure\_\allowbreak{}status item by item.\allowbreak{}\textquotedbl{}\\
\hspace*{0.5em}\},\allowbreak{}\\
\hspace*{0.5em}[...]\\
\}
\end{jsonbox}

\noindent (Summary.) One persistent executor realizes all three channels in one context; raw tool returns enter that context without a handoff object; the same executor writes the \texttt{write\_report} drafts and the \texttt{submit\_report} final; it can read $T/U$, its history, all raw returns and all drafts, and write drafts and the delivery; a single serial loop; events and the artifact version chain are logged.

\paragraph{What was formally represented and what was run.}
Every run analyzed in the paper executed with both boundaries coupled: one persistent executor retrieved material, wrote any drafts and submitted within one context.
H-W and H-D are not further basis cells; they keep both boundaries coupled but change the tool and state interface (persistent files, a shell, file submission).
H-D also offers delegation to context-isolated workers, which would create explicit handoffs, but no recorded run used delegation.
H10, H01 and H11 were implemented in an earlier development study that is not among the studies in Table~\ref{tab:setup}; we report no results from it and make no claim about explicit handoffs.
The exploratory study's four setups are not the four basis cells either: they cross two system-prompt wordings with the presence or absence of a draft tool inside one coupled loop.

\subsection{Implementations actually run}
\label{app:harness-impl}

Table~\ref{tab:appB-impl} summarizes each executed implementation: its tools and argument fields, the state the agent can see, and how drafts and the final report are recorded.

\begin{table}[t]
\centering
\caption{\textbf{Executed implementations.} Tools are listed with their argument fields. ``Final'' is the delivered report that the evaluators read.}
\label{tab:appB-impl}
\scriptsize
\setlength{\tabcolsep}{3pt}
\renewcommand{\arraystretch}{1.12}
\begin{tabularx}{\linewidth}{@{}>{\raggedright\arraybackslash}p{0.95in} >{\raggedright\arraybackslash}p{1.45in} >{\raggedright\arraybackslash}X >{\raggedright\arraybackslash}X@{}}
\toprule
\textsf{\textbf{Implementation (studies)}} & \textsf{\textbf{Tools (arguments)}} & \textsf{\textbf{State visible to the agent}} & \textsf{\textbf{Drafts and final}}\\
\midrule
Initial loop (initial randomized) & \texttt{search(query)}, \texttt{read(url)}, \texttt{write\_report(content)}, \texttt{submit()} with no argument & One context: system prompt, a scripted clarification exchange carrying the user condition, the task request, own history, raw tool returns & Each \texttt{write\_report} overwrites one draft; \texttt{submit()} delivers the current draft, so the final equals the last saved draft by construction\\
\addlinespace[2pt]
H00 report tool (exploratory S00/S10, main, both follow-ups; H-R) & \texttt{search(query)}, \texttt{read(url)}, \texttt{write\_report(content)}, \texttt{submit\_report(content)} & One context: system prompt, one user message, own history, raw tool returns (search: 6 results with 400-character snippets; read: first 6{,}000 characters of normalized page text) & Each \texttt{write\_report} is a new saved version; the final is the \texttt{submit\_report} content or a text-only reply, a separate object from the drafts\\
\addlinespace[2pt]
Direct submission (exploratory S01/S11) & \texttt{search(query)}, \texttt{read(url)}, \texttt{submit\_report(content)} & As H00, without a draft tool & No drafts; final as in H00\\
\addlinespace[2pt]
H-W file workspace (harness comparison) & \texttt{search(query)}, \texttt{read(url)}, \texttt{shell(command)}, \texttt{view\_file(path, offset, length)}, \texttt{write\_file(path, content)}, \texttt{edit\_file(path, old\_text, new\_text)}, \texttt{submit\_file(path)} & One root context plus a persistent sandboxed \texttt{/workspace}; the shell has no network, so external material arrives only through \texttt{search} and \texttt{read} & Files are snapshotted after every tool call; \texttt{/workspace/\allowbreak deliverable.md} forms the report-draft chain; the final is the exact file text read at \texttt{submit\_file}, or a text-only reply\\
\addlinespace[2pt]
H-D delegation (harness comparison) & H-W tools plus \texttt{delegate\_batch(tasks)}, where each task carries \{\texttt{assignment}, \texttt{input\_files}\}; workers: \texttt{search}, \texttt{read}, \texttt{shell}, \texttt{view\_file}, \texttt{write\_file}, \texttt{edit\_file} & Root as in H-W. A worker sees only its assignment text, a read-only snapshot of the selected files, and a private scratch directory; not the root history, sibling workers or other files & As in H-W; only the root can submit. Worker replies return verbatim in a fixed template. Offered, but no recorded run used delegation\\
\bottomrule
\end{tabularx}
\end{table}

\paragraph{Message organization and user information.}
In the initial randomized study, the system prompt is followed by a fixed scripted clarification exchange (assistant question, user answer) that states the shared background and, except in the task-only arm, the randomized user fact, and then by the task request.
Its system prompt asks the agent to keep the draft current rather than write only at the end, and after a reply without a tool call the runner appends a fixed user-role message ``(continue)'' (English translation of the recorded message) with no user information.
In all later implementations the user condition appears once, in the only user message (task text, shared background, endpoint text; Section~\ref{sec:counterfactual_control}); the system prompt states that no further user replies will arrive, and a reply without a tool call ends the episode.
Every saved record of the main study, the follow-ups and the harness comparison has exactly one user message and no injected one (checked for this appendix); the rest are assistant turns and tool returns, which are external material, not user feedback.
One follow-up checking condition, outside the primary samples, appended once to a tool return the fixed sentence ``Before the next substantive decision, re-read the frozen user information and verify that the decision remains consistent with it. Do not otherwise change the task.''
The three harness-comparison implementations send the same user message; H-W and H-D append one paragraph to the base system prompt, stating that files in \texttt{/workspace} persist, that external material comes only through \texttt{search} and \texttt{read}, and that \texttt{submit\_file} submits a file and ends the task (the original text and an English translation are released with the source files).
An H-D worker receives its own short system prompt and, as its only user message, the assignment written by the root followed by a fixed list of its input files.

\paragraph{Draft identity and versions.}
In the H00 family, each \texttt{write\_report} call is saved as a version (\texttt{artifact\_version\_id}, \texttt{write\_event\_id}, \texttt{decision\_index}, full \texttt{text}), and the submitted report is stored separately in \texttt{terminal.delivered\_text}.
In H-W and H-D, every workspace file is hashed after each tool call; non-empty versions of \texttt{/workspace/deliverable.md} form the report-draft chain (\texttt{W\_REPORT}), other text files form separate note chains (\texttt{W\_NOTES}), and handoff, export, input, log and cache files are never read as drafts.
Versions are ordered by event number, not file time, and identical texts are marked \texttt{IDENTICAL\_TEXT}.
The initial loop keeps one overwritten draft, recoverable from the saved messages, plus snapshots after 5, 15 and 40 effective actions and at the end that the agent never sees.

\begin{figure}[t]
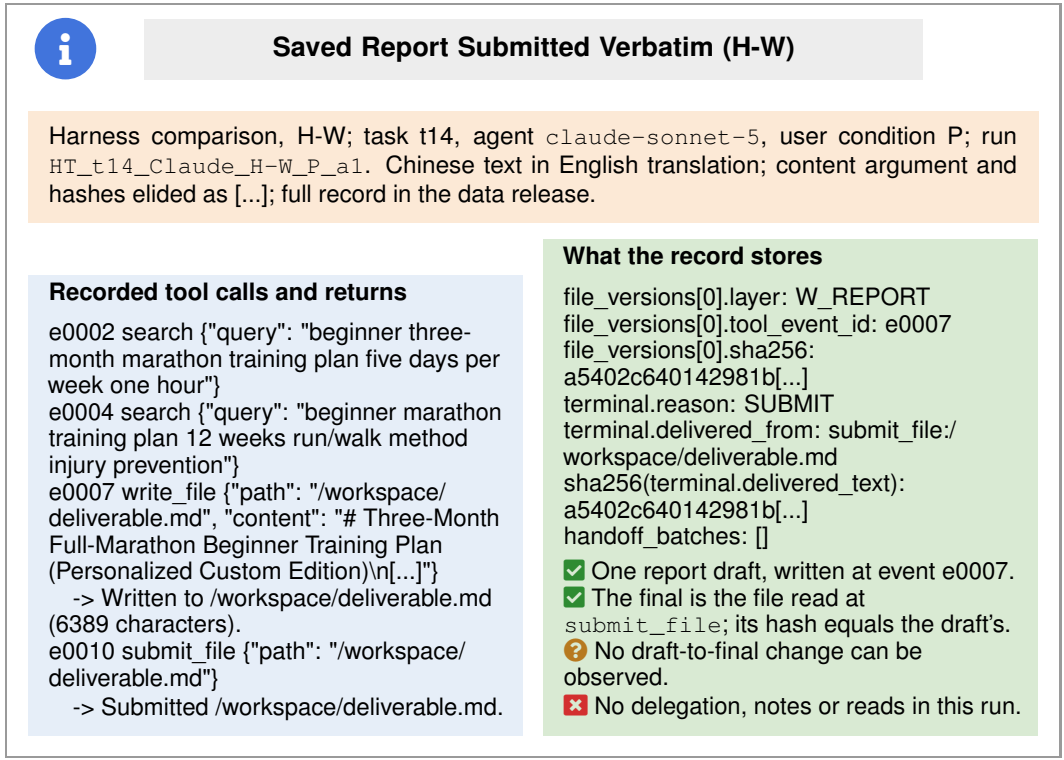

\begin{casebox}{\caseinfo}{Saved Report Submitted Verbatim (H-W)}
\begin{caseband}Harness comparison, H-W; task t14, agent \texttt{claude-sonnet-5}, user condition P; run \texttt{HT\_t14\_Claude\_H-W\_P\_a1}. Chinese text in English translation; content argument and hashes elided as [...]; full record in the data release.\end{caseband}
\casecols{\begin{casepanel}[apxBlueP]{Recorded tool calls and returns}{\fontencoding{T1}\selectfont
e0002 search \{\textquotedbl{}query\textquotedbl{}:\allowbreak{} \textquotedbl{}beginner three-\allowbreak{}month marathon training plan five days per week one hour\textquotedbl{}\}\\
e0004 search \{\textquotedbl{}query\textquotedbl{}:\allowbreak{} \textquotedbl{}beginner marathon training plan 12 weeks run/\allowbreak{}walk method injury prevention\textquotedbl{}\}\\
e0007 write\_\allowbreak{}file \{\textquotedbl{}path\textquotedbl{}:\allowbreak{} \textquotedbl{}/\allowbreak{}workspace/\allowbreak{}deliverable.\allowbreak{}md\textquotedbl{},\allowbreak{} \textquotedbl{}content\textquotedbl{}:\allowbreak{} \textquotedbl{}\# Three-\allowbreak{}Month Full-\allowbreak{}Marathon Beginner Training Plan (Personalized Custom Edition)\textbackslash{}n[.\allowbreak{}.\allowbreak{}.\allowbreak{}]\textquotedbl{}\}\\
\hspace*{1.0em}-\allowbreak{}\textgreater{} Written to /\allowbreak{}workspace/\allowbreak{}deliverable.\allowbreak{}md (6389 characters).\allowbreak{}\\
e0010 submit\_\allowbreak{}file \{\textquotedbl{}path\textquotedbl{}:\allowbreak{} \textquotedbl{}/\allowbreak{}workspace/\allowbreak{}deliverable.\allowbreak{}md\textquotedbl{}\}\\
\hspace*{1.0em}-\allowbreak{}\textgreater{} Submitted /\allowbreak{}workspace/\allowbreak{}deliverable.\allowbreak{}md.\allowbreak{}}\end{casepanel}}{\begin{casepanel}[apxGreen]{What the record stores}{\fontencoding{T1}\selectfont
file\_\allowbreak{}versions[0].\allowbreak{}layer:\allowbreak{} W\_\allowbreak{}REPORT\\
file\_\allowbreak{}versions[0].\allowbreak{}tool\_\allowbreak{}event\_\allowbreak{}id:\allowbreak{} e0007\\
file\_\allowbreak{}versions[0].\allowbreak{}sha256:\allowbreak{} a5402c640142981b[.\allowbreak{}.\allowbreak{}.\allowbreak{}]\\
terminal.\allowbreak{}reason:\allowbreak{} SUBMIT\\
terminal.\allowbreak{}delivered\_\allowbreak{}from:\allowbreak{} submit\_\allowbreak{}file:\allowbreak{}/\allowbreak{}workspace/\allowbreak{}deliverable.\allowbreak{}md\\
sha256(terminal.\allowbreak{}delivered\_\allowbreak{}text):\allowbreak{} a5402c640142981b[.\allowbreak{}.\allowbreak{}.\allowbreak{}]\\
handoff\_\allowbreak{}batches:\allowbreak{} []}\par\smallskip
\ok\ One report draft, written at event e0007.\\
\ok\ The final is the file read at \texttt{submit\_file}; its hash equals the draft's.\\
\unk\ No draft-to-final change can be observed.\\
\no\ No delegation, notes or reads in this run.
\end{casepanel}}
\end{casebox}
\caption{\textbf{A recorded H-W run in which the saved report is submitted verbatim.} Excerpts of the run record (event identifiers, tool arguments, tool returns and record fields), verbatim except that Chinese text is shown in English translation; the equality of the two hashes was computed for this appendix. Selected as the first H-W run file in lexical order; illustrative only.}
\label{fig:appB-hw-case}
\end{figure}

\paragraph{Saved report equals final by construction.}
In H-W and H-D, \texttt{submit\_file} delivers the saved report file itself, so the last saved report can be structurally identical to the final report (Fig.~\ref{fig:appB-hw-case}); these harnesses therefore do not provide an independent draft-to-final persistence test.
The same holds in the initial loop, whose \texttt{submit()} delivers the current draft.
A saved draft and the final can differ only when the final carries its own content (\texttt{submit\_report} in the H00 family, including H-R) or is a text-only reply.

\paragraph{Termination, retries and censoring.}
Natural endpoints are \texttt{SUBMIT} (the submitted content or file is the final) and \texttt{ASSISTANT\_FINAL} (the visible text of a reply without tool calls is the final).
\texttt{BUDGET\_END} and \texttt{PROVIDER\_LENGTH\_CUT} end a run without a new final: no extra model call is made and a draft is never promoted to a final.
An infrastructure failure after behavior has started (\texttt{INFRA\_FAILED}) is right-censored and stays in the planned denominator.
From the main study on, a run slot follows a fixed limited-attempt policy, and a new attempt is allowed only if the failed one produced no behavior; client transport retries resend the same request, and there is no semantic retry or best-of-$n$.
The harness comparison did not resume an episode after a failed slot attempt.
The initial randomized study allowed one automatic replacement per slot for a failed model call; truncation at its cap counted as an outcome and delivered the current draft.

\paragraph{Sandbox isolation (H-W, H-D).}
Each episode has its own file system (about 256\,MiB writable).
Shell commands run under bubblewrap with no network, a separate process namespace, no capabilities, read-only system directories, no host repository or credentials, a cleared environment and a 60-second limit; file tools refuse paths outside the workspace and symbolic links.
Because the shell has no network, every acquisition event is an executed \texttt{search} or \texttt{read} in all three harnesses.

\subsection{Model requests, reasoning settings and budgets}
\label{app:harness-models}

Table~\ref{tab:appB-models} lists the requested model identifiers and reasoning controls; sampling settings, dates and budgets follow.

\begin{table}[t]
\centering
\caption{\textbf{Agent model requests.} Requested model identifiers and reasoning controls as sent. Some routes use a local OpenAI-compatible gateway; a matching model echo does not verify the underlying weights.}
\label{tab:appB-models}
\scriptsize
\setlength{\tabcolsep}{3pt}
\renewcommand{\arraystretch}{1.12}
\begin{tabularx}{\linewidth}{@{}l l >{\raggedright\arraybackslash}p{1.05in} >{\raggedright\arraybackslash}X >{\raggedright\arraybackslash}X@{}}
\toprule
\textsf{\textbf{Family}} & \textsf{\textbf{Requested ID}} & \textsf{\textbf{Protocol}} & \textsf{\textbf{Lowest setting (main, harness comparison)}} & \textsf{\textbf{High setting (both follow-ups)}}\\
\midrule
GPT      & \texttt{gpt-5.4}          & chat completions; responses in the follow-ups & \texttt{reasoning\_effort=none} (reasoning tokens not visible) & \texttt{reasoning\_effort=high}; dated echo \texttt{gpt-5.4-2026-03-05} accepted\\
Claude   & \texttt{claude-sonnet-5}  & chat completions & no reasoning field (no extended thinking by default) & \texttt{reasoning\_effort=high}\\
Gemini   & \texttt{gemini-3.8-flash} & chat completions & \texttt{reasoning\_effort=low} (lowest accepted) & \texttt{reasoning\_effort=high}\\
Grok     & \texttt{grok-4.6}         & responses & \texttt{reasoning.effort=none}; still reports reasoning tokens & not evaluated at the higher setting\\
DeepSeek & \texttt{deepseek-v4-pro}  & chat completions (provider API) & \texttt{thinking: disabled}; zero reasoning tokens required per call & \texttt{thinking: enabled}, \texttt{reasoning\_effort=high}\\
\bottomrule
\end{tabularx}
\par\vspace{4pt}
{\footnotesize\raggedright The initial randomized and exploratory studies used \texttt{deepseek-v4-pro} through a third-party hosted endpoint with thinking disabled; later studies used DeepSeek's own API. A small ablation subset sent no reasoning field (provider default) on the relevant routes.\par}
\end{table}

\paragraph{Sampling, dates and budgets.}
All studies used temperature 1.0 (not sent to Grok, whose route does not accept it), at most 16{,}000 output tokens per call and no seed; from the exploratory study on, requests set \texttt{parallel\_tool\_calls=false}, and tool calls returned together were executed in order.
Runs took place on 2026-09-02 (initial randomized study), 2026-09-10 (exploratory), 2026-09-23 local time (main study, follow-ups) and 2026-09-23 UTC (harness comparison).
From the main study on, one episode ledger holds 96 model decisions, 192 tool actions, 96 acquisition requests and 128{,}000 output tokens; in H-W and H-D the root and any workers share it with tokens reserved before each call, while H00 and H-R let the last call exceed the total by up to one call's cap.
The exploratory study capped decisions (96) and tool actions (192); the initial loop capped effective tool actions at 120 (at most 360 model calls).
No budget was shown to the agent; no implementation imposes a minimum number of steps, a phase order, a writing step or a per-episode wall-clock limit; from the main study on, retrieval requests time out after 90 seconds with at most three retries.
``Thinking disabled'' has two qualifications: the initial randomized study did not save request bodies, so its setting is the client default rather than a per-call record; and in the exploratory study three requests in two analyzed runs were sent without the thinking-off field (\texttt{thinking\_field\_sent} = false) and report 21, 14 and 110 reasoning tokens (checked for this appendix).
From the main study on, the DeepSeek client halts the route instead of dropping the field.

\subsection{Recorded conformance checks}
\label{app:harness-checks}

Table~\ref{tab:appB-checks} lists the checks recorded for the harness comparison and the two execution defects recorded in earlier studies.
The tests were written by independent AI verifier agents (not humans) against a frozen acceptance list (A01--A64) and run on the frozen code before the first main call.
Passing a check does not show that a capability was used: delegation passed offline and development tests with all tested families but was never exercised in the main runs.

\begin{table}[t]
\centering
\caption{\textbf{Recorded conformance checks.} Row shading gives the recorded outcome of each check at the time it ran; fixes are stated in the evidence column. Requirement codes (A01--A64) refer to the frozen acceptance list.}
\label{tab:appB-checks}
\scriptsize
\setlength{\tabcolsep}{3pt}
\renewcommand{\arraystretch}{1.15}
\begin{tabularx}{\linewidth}{@{}>{\raggedright\arraybackslash}p{1.35in} X l@{}}
\toprule
\textsf{\textbf{Check}} & \textsf{\textbf{Recorded evidence}} & \textsf{\textbf{Outcome}}\\
\midrule
\multicolumn{3}{@{}l}{\textit{Harness comparison: before the first main call}}\\
\rowpass \textsf{\textbf{Same input}} & Identical user message in all three harnesses, P and Q differing only in the frozen endpoint (A01; no recorded mismatch at the end); H-R shows the agent the same messages as the frozen H00 loop on a simulated response sequence (A09) & \PASS\\
\rowpass \textsf{\textbf{Endpoints:}} no forced final & A text-only reply is a natural final; budget end makes no extra call and never promotes a draft (A13, A15) & \PASS\\
\rowfail \textsf{\textbf{Sandbox:}} process environment & The sandbox's first process inherited the orchestrator environment, including a provider credential, and a command line naming host paths and the run identifier, which contains the user-condition label (D19). Fixed before any main call; no development shell call had read it & \FAIL\\
\rowpass \textsf{\textbf{Isolation and budget}} & After the fixes: no path traversal or host links, secrets, other-arm text or shell network (A17--A20, A28); a worker sees only its assignment and selected files, cannot submit or delegate, and returns verbatim (A21--A25); one shared ledger that the shell cannot bypass (A29, A30, A33). In all, 219 tests for 64 requirements passed and 52 injected mutations were detected & \PASS\\
\addlinespace[2pt]
\multicolumn{3}{@{}l}{\textit{Harness comparison: during and after the main runs}}\\
\rowfail \textsf{\textbf{Route halt}} & A provider error unrelated to reasoning halted the whole DeepSeek route; an independent verifier judged this inconsistent with the frozen routing rule. Minimal fix; the affected run was recorded as censored after behavior and 6 unstarted slots were resumed; regression 220 of 221 tests passed, 1 skipped by design (D26, D27) & \FAIL\\
\rowpass \textsf{\textbf{Wave health checks}} & the frozen wave-level checks (model echo, reasoning witness, budget, file provenance, no delegation in H-W, record hashes) before and after the fix & \PASS\\
\rowunclear \textsf{\textbf{Delegation used}} & Offered in H-D and exercised in development runs by all tested families; not used in the analyzed harness runs, so delegated execution was not tested & \badge{apxUnclearFg}{NOT EXERCISED}\\
\rowpass \textsf{\textbf{Mechanical audit}} & All 240 slots audited; one flag, a request interrupted by a retrieval error before its resource was recorded (URL kept in the tool arguments); not a violation & \PASS\\
\rowpass \textsf{\textbf{Recomputation}} & Independent recomputation from raw evaluator replies, without importing the production modules: 7214 of 7214 items equal & \PASS\\
\addlinespace[2pt]
\multicolumn{3}{@{}l}{\textit{Earlier studies}}\\
\rowfail \textsf{\textbf{Main study:}} tool calls & The gateway split some Claude replies into several choices and the client kept only the first, dropping tool calls. Fixed by merging choices; the 4 affected attempts were invalidated and replaced & \FAIL\\
\rowfail \textsf{\textbf{Initial study:}} runner & The retrieval tool crashed the runner on a non-ASCII URL in an archived run; replaced as an infrastructure failure, harness left unchanged during the run & \FAIL\\
\bottomrule
\end{tabularx}
\end{table}

Two limits follow from these records.
First, the harness comparison varies the tool and state interface while all executed runs keep both boundaries coupled; it does not test cross-context delegation or explicit finalization.
Second, the initial loop differs from H00 in its system prompt (which encourages continuous drafting), its argument-free submission and its step cap, so the two ``report tool'' labels in Table~\ref{tab:setup} denote related but not identical implementations.

\providecommand{\appCIsrc}[1]{\par\vspace{-2pt}{\scriptsize\raggedright\textsf{Source:} #1\par}\vspace{4pt}}

\section{Prompts and Directional Evaluation}
\label{app:measurement_spec}

This appendix reproduces the text that models actually received and specifies how evaluator outputs become the directional labels of Section~\ref{sec:directional_measurement}.
This first part covers the generation side: the prompts used to construct user conditions (Appendix~\ref{sec:appC-construction-prompts}) and the inputs given to the research agent (Appendix~\ref{sec:appC-agent-prompts}).
The second part covers the evaluation side: the prompt for acquisition requests, the extraction prompt and schema for drafts and final reports, the mirrored reads, and the deterministic rules for aggregation, acceptance, and abstention.

\paragraph{Conventions.}
Each box is a mechanical copy of a source file in the version used by the runs: LaTeX special characters are escaped, line breaks follow the source, and the wording is unchanged.
Titles state the original language; boxes contain no translations.
[\ldots] marks an omission, \texttt{<REPO>} replaces a local path, and bold sans-serif labels inside a box are our annotations.
The line under each box gives the file, the locator, and the first 16 hexadecimal digits of the file's SHA-256; complete copies are in the supplementary material (\nolinkurl{appendix_sources/C1/}).

\subsection{Prompts for Constructing User Conditions}
\label{sec:appC-construction-prompts}

User conditions were fixed before the first agent run of each study, but the procedure differed across studies (Table~\ref{tab:appC-construction}).
In the main study, whose cards the reasoning follow-up and the harness comparison reuse, every construction judgment was made by a model.
One constructor session per task (claude-opus-5-5, a workflow subagent with file and shell access) followed a written protocol and wrote a factor card: the shared background $C$, the $P$ endpoint (the profile's own state, with a verbatim quote of the profile field), the $Q$ endpoint (a minimal counterfactual), and the rubrics later used by the evaluators.
The protocol fixes the choice in advance: use the first profile in a fixed hash order and take the first profile field, in field order, that meets seven conditions, rather than the factor expected to show the largest effect.
After a script accepted the card, two reviewer sessions (claude-haiku-4-5-20251001) read an anonymized packet, with the endpoints shown as \zh{A} and \zh{B} in hash order and the rubrics withheld, and checked five fatal problems.
Before its task message, each constructor and reviewer session also received a one-line operator message relayed by the workflow harness; it concerned agent reasoning settings and gave no construction instruction (copy in the supplementary material).
No human reviewed the main-study cards; the human screening in Section~\ref{sec:factor_construction} was applied to the candidate pool of the initial randomized study (Appendices~\ref{app:factor_construction} and~\ref{app:human_validation}).

\refstepcounter{table}\label{tab:appC-construction}

\begin{promptbox}{Prompt for factor-card construction, main study (English translation)}
\footnotesize\setlength{\parskip}{2pt}
\raggedright
You are the factor-card constructor for the third-round experiment, responsible only for task t8. Working directory: \textless{}REPO\textgreater{}/\allowbreak{}exp\_\allowbreak{}v2/\allowbreak{}ROUND3\_\allowbreak{}FINAL\_\allowbreak{}V1
\par
First read the procedure in full: \textless{}REPO\textgreater{}/\allowbreak{}exp\_\allowbreak{}v2/\allowbreak{}ROUND3\_\allowbreak{}FINAL\_\allowbreak{}V1/\allowbreak{}R3\_\allowbreak{}CARD\_\allowbreak{}CONSTRUCTION\_\allowbreak{}GUIDE.\allowbreak{}md , and follow it strictly.\\
Input: \textless{}REPO\textgreater{}/\allowbreak{}exp\_\allowbreak{}v2/\allowbreak{}ROUND3\_\allowbreak{}FINAL\_\allowbreak{}V1/\allowbreak{}CONSTRUCTION\_\allowbreak{}INPUTS/\allowbreak{}t8.\allowbreak{}json (the task text, whether it is a core task, and the raw PDR user personas sorted in a fixed order).\\
Output: \textless{}REPO\textgreater{}/\allowbreak{}exp\_\allowbreak{}v2/\allowbreak{}ROUND3\_\allowbreak{}FINAL\_\allowbreak{}V1/\allowbreak{}CARDS\_\allowbreak{}DRAFT/\allowbreak{}t8.\allowbreak{}json
\par
Hard requirements:\\
-\allowbreak{} Work only from the task text and the personas. Do not open any trajectory, reading or result file from old experiments, and do not look at the cards of other tasks under CARDS\_\allowbreak{}DRAFT.\\
-\allowbreak{} After writing, run: cd \textless{}REPO\textgreater{}/\allowbreak{}exp\_\allowbreak{}v2/\allowbreak{}ROUND3\_\allowbreak{}FINAL\_\allowbreak{}V1 \&\& python3 scripts/\allowbreak{}r3\_\allowbreak{}card\_\allowbreak{}validate.\allowbreak{}py CARDS\_\allowbreak{}DRAFT/\allowbreak{}t8.\allowbreak{}json ; if it outputs FAIL, revise according to the hints until it gives PASS. Change only the problems that were pointed out.\\
-\allowbreak{} After PASS, run: cd \textless{}REPO\textgreater{}/\allowbreak{}exp\_\allowbreak{}v2/\allowbreak{}ROUND3\_\allowbreak{}FINAL\_\allowbreak{}V1 \&\& python3 scripts/\allowbreak{}r3\_\allowbreak{}review\_\allowbreak{}packet.\allowbreak{}py 8 , to generate the anonymized review packet.\\
-\allowbreak{} If the task is judged ineligible under Section 8 of the procedure, write a file containing only \{\textquotedbl{}taskid\textquotedbl{},\allowbreak{}\textquotedbl{}status\textquotedbl{}:\allowbreak{}\textquotedbl{}INELIGIBLE\textquotedbl{},\allowbreak{}\textquotedbl{}reason\textquotedbl{}\}, and do not run the validator or the review-packet script.\\
-\allowbreak{} Write all card content in Chinese.
\par
Finally, return according to the schema: status, which user was used, how many were skipped, the second-factor status (fill in NOT\_\allowbreak{}CORE for non-core tasks), the verbatim text of the validator's last output, the review-packet path, and a one-sentence explanation.
\end{promptbox}
\appCIsrc{workflow \texttt{r3-factor-cards}, session \texttt{construct:t8}, second user turn (transcript \texttt{agent-accc0eecf70ec83d9.jsonl}, sha256 \texttt{fb7d7eae7962c38d}); template in workflow record \texttt{wf\_4c15923f-4c8.json} (\texttt{008883c634271f80}). The other tasks received the same text with their own task number.}

\begin{promptbox}{Prompt for factor-card construction: protocol read by the constructor, \S4 (English translation)}
\footnotesize\setlength{\parskip}{2pt}
\#\# 4. Focal factor `focal\_\allowbreak{}factor` (ID fixed as `t\textless{}task number\textgreater{}\_\allowbreak{}f1`)
\par
**How to choose**: following the order of the persona fields, take the first piece of task-relevant information that satisfies all seven conditions below. Do not go picking the one where \textquotedbl{}the effect is easiest to see\textquotedbl{}.
\par
1.\allowbreak{} The information has real, consequential relevance in the task (it would change how a good plan should be written);\\
2.\allowbreak{} Both endpoints are reasonable and could both occur in reality;\\
3.\allowbreak{} The task text does not force either endpoint;\\
4.\allowbreak{} After switching to the other endpoint, this user is still coherent as a whole;\\
5.\allowbreak{} It is a minimal counterfactual: change only this one item and leave everything else verbatim unchanged;\\
6.\allowbreak{} Neither endpoint directly contains the answer;\\
7.\allowbreak{} It can coexist with the shared background.
\par
**Two endpoints**: `P\_\allowbreak{}endpoint\_\allowbreak{}text` is the persona's original state, written in the first person; `Q\_\allowbreak{}endpoint\_\allowbreak{}text` is the minimal counterfactual.\\
The two endpoints are close in length, consistent in tone, and differ only in this one piece of information.\\
`source\_\allowbreak{}user\_\allowbreak{}statement` must be taken **verbatim** from the persona record (the record is JSON and its field values are strings; directly copy a contiguous span of at least 6 characters from one of them);\\
`source\_\allowbreak{}persona\_\allowbreak{}field` gives its location in the persona, for example `Financial information.\allowbreak{}Financial situation.\allowbreak{}Asset status`.
\par
[\ldots]
\end{promptbox}
\appCIsrc{\nolinkurl{exp_v2/ROUND3_FINAL_V1/R3_CARD_CONSTRUCTION_GUIDE.md}, lines 31--46 of 136 (sha256 \texttt{b365edab646f7e74}, as registered in the main-study design freeze). Omitted: lines 1--30 (deliverable; use the first profile unless it has no usable task-relevant information; shared background of 150--400 characters that excludes the factor and anything coupled to it), lines 47--60 (fields \texttt{kind}, \texttt{scope}, \texttt{relation\_between\_endpoints}, \texttt{eligibility\_rule}), lines 61--136 (evaluator rubrics, second factor, profile assembly, ineligibility, template).}

\begin{promptbox}{Prompt for input review, main study (English translation)}
\footnotesize\setlength{\parskip}{2pt}
\raggedright
You are an input reviewer (ID R1), reviewing the user-condition design of a research task. Read only this one file: \textless{}REPO\textgreater{}/\allowbreak{}exp\_\allowbreak{}v2/\allowbreak{}ROUND3\_\allowbreak{}FINAL\_\allowbreak{}V1/\allowbreak{}REVIEW\_\allowbreak{}PACKETS/\allowbreak{}t8.\allowbreak{}md\\
Do not open any file outside that directory (in particular, do not look at CARDS\_\allowbreak{}DRAFT, REVIEW\_\allowbreak{}PACKETS\_\allowbreak{}PRIVATE\_\allowbreak{}KEYS, or any experimental results). Answer independently, without consulting anyone else's opinion.
\par
The packet contains: the original text of one task; the shared background of one user; and the two anonymous endpoints (A/\allowbreak{}B) of one swapped piece of user information. The core task also has a second, independently swapped piece of information.\\
The experiment will have a research assistant complete the task independently under ``shared background + A'' or ``shared background + B'', and then compare the behavior on the two sides.
\par
I. Fatal-problem check (judge fatal=\allowbreak{}true only when the problem genuinely holds, and in detail point out which two places in the original text are involved):\\
\hspace*{1.0em}1 contradiction: the original task text, the shared background, and the endpoints contradict one another;\\
\hspace*{1.0em}2 endpoint\_\allowbreak{}illegal\_\allowbreak{}or\_\allowbreak{}implausible: an endpoint is unreasonable in reality, or is not self-consistent when placed together with the shared background;\\
\hspace*{1.0em}3 task\_\allowbreak{}forces\_\allowbreak{}one\_\allowbreak{}endpoint: the original task text already forces one of the endpoints, and the other endpoint conflicts with the task requirements;\\
\hspace*{1.0em}4 endpoint\_\allowbreak{}states\_\allowbreak{}the\_\allowbreak{}answer: the endpoint text directly states what recommendation or conclusion should be given;\\
\hspace*{1.0em}5 more\_\allowbreak{}than\_\allowbreak{}one\_\allowbreak{}thing\_\allowbreak{}changes: besides this one piece of information, endpoints A and B also simultaneously change other user information.\\
\hspace*{1.0em}``This information may have little effect on the result'' and ``it is hard to measure'' are not fatal problems; do not judge fatal on those grounds.
\par
[\ldots]
\par
Return according to the schema. Set taskid to 8.
\end{promptbox}
\appCIsrc{workflow \texttt{r3-factor-cards}, session \texttt{review:t8:R1}, second user turn (\texttt{agent-a05ebcbd7810f943b.jsonl}, sha256 \texttt{8db40c5c3d1bb6a4}). Omitted: lines 15--20, a second question on where the factor should act (\texttt{EVIDENCE\_DIFFERENT}, \texttt{SELECTION\_DIFFERENT}, \texttt{BOTH}, \texttt{UNRESOLVED}), recorded as an input-only prior. Packet: \nolinkurl{exp_v2/ROUND3_FINAL_V1/REVIEW_PACKETS/t8.md} (\texttt{b83eb7c51a0cb780}). Reviewer R2 received the same text with ``R2''.}

\begin{promptbox}{Prompt for counterfactual twin construction, initial randomized study (original English; system message)}
\footnotesize\setlength{\parskip}{2pt}
You construct ONE counterfactual twin pair for an experimental user-\allowbreak{}evidence\\
treatment.\allowbreak{}
\par
The original evidence is grounded in a human-\allowbreak{}authored source.\allowbreak{}
\par
Your job is NOT to improve the source wording and NOT to optimize for a judge.\allowbreak{}
\par
Construct:\allowbreak{}\\
-\allowbreak{} one endpoint that faithfully preserves the source-\allowbreak{}grounded user evidence;\allowbreak{}\\
-\allowbreak{} one strong-\allowbreak{}but-\allowbreak{}plausible counterfactual endpoint on the SAME underlying\\
\hspace*{1.0em}user dimension.\allowbreak{}
\par
The two endpoints must differ on only ONE dimension.\allowbreak{}
\par
You MUST NOT introduce:\allowbreak{}\\
-\allowbreak{} a new goal;\allowbreak{}\\
-\allowbreak{} a new budget unless budget is the target dimension;\allowbreak{}\\
-\allowbreak{} a new deadline unless time horizon is the target dimension;\allowbreak{}\\
-\allowbreak{} new authority;\allowbreak{}\\
-\allowbreak{} new expertise;\allowbreak{}\\
-\allowbreak{} a concrete downstream option;\allowbreak{}\\
-\allowbreak{} a direct answer;\allowbreak{}\\
-\allowbreak{} a second independent preference.\allowbreak{}
\par
The counterfactual should be strong enough to create a meaningful contrast,\allowbreak{}\\
but it must remain a coherent human position.\allowbreak{}
\par
Return JSON only.\allowbreak{}
\par
All natural-\allowbreak{}language output values must use the same language as SOURCE\_\allowbreak{}TEXT.\allowbreak{}\\
Do not translate the source evidence or the proposed treatment endpoints.\allowbreak{}\\
JSON keys stay in English;\allowbreak{} only the values follow the source language.\allowbreak{}
\end{promptbox}
\appCIsrc{\nolinkurl{exp_v1/src/experiments/exp_a/prompts.py}, \texttt{A2\_SYSTEM} (file sha256 \texttt{7913d150324a1481}); the prompt hash recomputed from this file equals the pre-call hash \texttt{8623bb9bb5137e77} registered in \texttt{A00\_FREEZE\_MANIFEST\_PRECALL.json}. Not shown: the user template (\texttt{A2\_USER\_TEMPLATE}: source evidence, declared dimension, evidence type, and abstract decision interface in; \texttt{source\_grounded\_endpoint}, \texttt{counterfactual\_endpoint}, \texttt{shared\_core}, \texttt{changed\_dimension}, \texttt{held\_fixed}, \texttt{counterfactual\_rationale} out) and the proposal (A1) and verifier (A4) prompts, which match their registered hashes; all are in the supplementary material.}

\subsection{Agent Prompts, Inputs, and Tools}
\label{sec:appC-agent-prompts}

Every main-study run used the report-tool harness $\mathcal H_{00}$ (Appendix~\ref{app:harness}): one system message (P0), one user message rendered from a fixed template, and four tool schemas.
The template places the task text, the shared background $C$, and the endpoint text $z_f^u$ in that order; in the two-factor extension both endpoint texts fill the last slot.
Afterwards the agent receives only tool results and fixed harness receipts, such as ``Draft saved.'' after \texttt{write\_report}; budgets are not shown.
In the saved records of all main-study runs, the message history holds exactly one system and one user message, and every first request carries the same four tool schemas in the same order.
Other studies and harnesses changed the system message as listed in Table~\ref{tab:appC-agent-settings}; their full texts are in the supplementary material.

\begin{promptbox}{Prompt for the research agent: system message and user-message template (English translation)}
\footnotesize\setlength{\parskip}{2pt}
\textsf{\textbf{System message P0}} {\scriptsize(main study, follow-ups, H-R; exploratory setups S00, S01)}\par
You are a research assistant. Please complete the user's research task, use the available tools as needed, and submit a research report. You will not receive any new user replies during this run.
\par
\textsf{\textbf{User message template}} {\scriptsize(all studies except the initial randomized study)}\par
Research task:\\
\{task\_\allowbreak{}text\}
\par
User background:\\
\{common\_\allowbreak{}background\}\\
\{one\_\allowbreak{}user\_\allowbreak{}factor\_\allowbreak{}endpoint\}
\end{promptbox}
\appCIsrc{\nolinkurl{exp_v2/NATURAL_USER_CONTROL_ROBUST_DISCOVERY_V1/02_ACTOR_PROMPTS_AND_TOOLS.json}, keys \texttt{prompts.P0} and \texttt{user\_render\_template} (file sha256 \texttt{dfc38b4a004892e5}, checked by the runner at load; text hash of P0 \texttt{e947a39d6c767381}, as recorded in every main-study run). Variants in the supplementary material: \texttt{prompts.P1} in the same file; root and worker messages in \nolinkurl{exp_v2/HARNESS_TRANSPORT_V1/HARNESS_CONTRACTS/H-D.json} (\texttt{42ec3c9160e1b3a6}).}

\begin{jsonbox}{Tool schemas sent in the first request of every main-study run (H00; JSON)}
\raggedright
[\\
\{\textquotedbl{}type\textquotedbl{}:\allowbreak{} \textquotedbl{}function\textquotedbl{},\allowbreak{} \textquotedbl{}function\textquotedbl{}:\allowbreak{} \{\textquotedbl{}name\textquotedbl{}:\allowbreak{} \textquotedbl{}search\textquotedbl{},\allowbreak{} \textquotedbl{}description\textquotedbl{}:\allowbreak{} \textquotedbl{}Search external sources and return result titles, URLs, and snippets.\textquotedbl{},\allowbreak{} \textquotedbl{}parameters\textquotedbl{}:\allowbreak{} \{\textquotedbl{}type\textquotedbl{}:\allowbreak{} \textquotedbl{}object\textquotedbl{},\allowbreak{} \textquotedbl{}properties\textquotedbl{}:\allowbreak{} \{\textquotedbl{}query\textquotedbl{}:\allowbreak{} \{\textquotedbl{}type\textquotedbl{}:\allowbreak{} \textquotedbl{}string\textquotedbl{}\}\},\allowbreak{} \textquotedbl{}required\textquotedbl{}:\allowbreak{} [\textquotedbl{}query\textquotedbl{}],\allowbreak{} \textquotedbl{}additionalProperties\textquotedbl{}:\allowbreak{} false\}\}\},\allowbreak{}\\
\{\textquotedbl{}type\textquotedbl{}:\allowbreak{} \textquotedbl{}function\textquotedbl{},\allowbreak{} \textquotedbl{}function\textquotedbl{}:\allowbreak{} \{\textquotedbl{}name\textquotedbl{}:\allowbreak{} \textquotedbl{}read\textquotedbl{},\allowbreak{} \textquotedbl{}description\textquotedbl{}:\allowbreak{} \textquotedbl{}Read the body text of the material at the specified URL.\textquotedbl{},\allowbreak{} \textquotedbl{}parameters\textquotedbl{}:\allowbreak{} \{\textquotedbl{}type\textquotedbl{}:\allowbreak{} \textquotedbl{}object\textquotedbl{},\allowbreak{} \textquotedbl{}properties\textquotedbl{}:\allowbreak{} \{\textquotedbl{}url\textquotedbl{}:\allowbreak{} \{\textquotedbl{}type\textquotedbl{}:\allowbreak{} \textquotedbl{}string\textquotedbl{}\}\},\allowbreak{} \textquotedbl{}required\textquotedbl{}:\allowbreak{} [\textquotedbl{}url\textquotedbl{}],\allowbreak{} \textquotedbl{}additionalProperties\textquotedbl{}:\allowbreak{} false\}\}\},\allowbreak{}\\
\{\textquotedbl{}type\textquotedbl{}:\allowbreak{} \textquotedbl{}function\textquotedbl{},\allowbreak{} \textquotedbl{}function\textquotedbl{}:\allowbreak{} \{\textquotedbl{}name\textquotedbl{}:\allowbreak{} \textquotedbl{}submit\_\allowbreak{}report\textquotedbl{},\allowbreak{} \textquotedbl{}description\textquotedbl{}:\allowbreak{} \textquotedbl{}Submit the final research report in content and end the task. The report uses free text.\textquotedbl{},\allowbreak{} \textquotedbl{}parameters\textquotedbl{}:\allowbreak{} \{\textquotedbl{}type\textquotedbl{}:\allowbreak{} \textquotedbl{}object\textquotedbl{},\allowbreak{} \textquotedbl{}properties\textquotedbl{}:\allowbreak{} \{\textquotedbl{}content\textquotedbl{}:\allowbreak{} \{\textquotedbl{}type\textquotedbl{}:\allowbreak{} \textquotedbl{}string\textquotedbl{}\}\},\allowbreak{} \textquotedbl{}required\textquotedbl{}:\allowbreak{} [\textquotedbl{}content\textquotedbl{}],\allowbreak{} \textquotedbl{}additionalProperties\textquotedbl{}:\allowbreak{} false\}\}\},\allowbreak{}\\
\{\textquotedbl{}type\textquotedbl{}:\allowbreak{} \textquotedbl{}function\textquotedbl{},\allowbreak{} \textquotedbl{}function\textquotedbl{}:\allowbreak{} \{\textquotedbl{}name\textquotedbl{}:\allowbreak{} \textquotedbl{}write\_\allowbreak{}report\textquotedbl{},\allowbreak{} \textquotedbl{}description\textquotedbl{}:\allowbreak{} \textquotedbl{}Save or replace the current report draft. This operation does not submit the report or end the task. The draft uses free text.\textquotedbl{},\allowbreak{} \textquotedbl{}parameters\textquotedbl{}:\allowbreak{} \{\textquotedbl{}type\textquotedbl{}:\allowbreak{} \textquotedbl{}object\textquotedbl{},\allowbreak{} \textquotedbl{}properties\textquotedbl{}:\allowbreak{} \{\textquotedbl{}content\textquotedbl{}:\allowbreak{} \{\textquotedbl{}type\textquotedbl{}:\allowbreak{} \textquotedbl{}string\textquotedbl{}\}\},\allowbreak{} \textquotedbl{}required\textquotedbl{}:\allowbreak{} [\textquotedbl{}content\textquotedbl{}],\allowbreak{} \textquotedbl{}additionalProperties\textquotedbl{}:\allowbreak{} false\}\}\}\\
]
\end{jsonbox}
\appCIsrc{\nolinkurl{exp_v2/ROUND3_FINAL_V1/RUNS_D/requests/R3_a7f5fa6695ff4e007c9b9217_a1.jsonl}, line 1, key \texttt{body.tools} (\texttt{fcd9962de8c373a3}), one tool per line. The request also sets \texttt{"parallel\_tool\_calls": false}.}

Figure~\ref{fig:appC-filled-input} shows one real filled input, the first user message of the paired DeepSeek runs for task 8; the two runs sent it in the same second, and the texts differ only in the last sentence, which states the city of residence.
The complete first request of the $P$ run is in the supplementary material; its parameters are those of the DeepSeek row of Table~\ref{tab:appC-agent-settings}(b).

\begin{figure}[!htbp]
\centering
\begin{taskpanel}{Deep Research Task (PDR-Bench task 8)}
\iconrow[apxOrange]{\faSearch}{Research task:\\ I want to be promoted to a management position at my company in the future. I have experience independently leading projects, but I feel I lack experience in team communication and task coordination. I would like to know how to systematically improve my leadership and management abilities, especially in communication and coordination, motivating a team, and decision-making. Please help me plan a concrete learning path, recommend some suitable books, online courses, or training programs, and tell me which practical opportunities, at work or outside work, I should focus on seizing to build these abilities.}
\end{taskpanel}
\vspace{1mm}
\begin{taskpanel}{Shared background $C$ (identical in both conditions)}
\iconrow[apxOrange]{\faUserTie}{User background:\\ I am around thirty years old and work as an algorithm engineer, with a formal computer-science education; in graduate school I worked on artificial intelligence and recommender systems and also published related papers. Most of my working time is spent writing code and tuning models. My personality is fairly calm and rational, and I like logical reasoning and deep thinking; I am quite talkative with people I know, but fairly introverted with strangers. When making decisions I am used to making up my own mind, usually reaching conclusions through data analysis and logical reasoning; when facing a major choice I listen to the opinions of my parents or my advisor. I am more productive in the evenings and usually like to use evening time to learn new things. With money I am fairly rational and value cost-effectiveness, but I am still willing to invest some in educational courses.}
\end{taskpanel}
\vspace{1mm}
\begin{tcbraster}[raster columns=2,raster equal height,raster column skip=3mm]
\begin{userpanel}[P]{User condition P}
\iconrow[apxHlP]{\faCity}{\hlP{I currently live in Hangzhou, where I rent a studio apartment in the city, fairly close to my company.}}
\end{userpanel}
\begin{userpanel}[Q]{User condition Q}
\iconrow[apxHlQ]{\faCity}{\hlQ{I currently live in Shenzhen, where I rent a studio apartment in the city, fairly close to my company.}}
\end{userpanel}
\end{tcbraster}
\caption{\textbf{A real filled input from the main study} (English translation). The first user message of each run is the task box, a blank line, the background box, and that run's colored endpoint sentence on the next line; the endpoint (Hangzhou vs.\ Shenzhen) is the only difference. Icons are decorative. Runs \texttt{R3\_a7f5fa66\ldots} ($P$) and \texttt{R3\_00e356fb\ldots} ($Q$), DeepSeek; request journals, line~1, key \texttt{body.messages[1].content} (sha256 \texttt{fcd9962de8c373a3} and \texttt{0356b16a99735794}).}
\label{fig:appC-filled-input}
\end{figure}

Table~\ref{tab:appC-agent-settings} lists what differed across studies, harnesses, and model families; tool returns, budgets, and termination are in Appendix~\ref{app:harness}.
Two further inputs reached some agents after the first message.
The runner of the initial randomized study appended the fixed user-role message (continue) (``continue'', translation) when the agent replied without calling a tool; it carries no user information.
A separate mitigation arm of the follow-ups, not counted in Table~\ref{tab:setup}, appended the fixed sentence ``Before the next substantive decision, re-read the frozen user information and verify that the decision remains consistent with it. Do not otherwise change the task.'' once to a tool return (\nolinkurl{exp_v2/ROUND4_FINAL_V1/MITIGATION_DESIGN.json}, key \texttt{checkpoint\_text}; see Appendix~\ref{app:additional_results}).

\begin{table}[!htbp]
\centering
\caption{\textbf{Agent inputs across studies, harnesses, and model families.} (a) Tools are listed in the order sent. (b) Request fields that differed across families, read from the first request of every run; all other fields were shared: temperature 1.0 (not sent to Grok), an output cap of 16000 tokens per call, and \texttt{parallel\_tool\_calls} false.}
\label{tab:appC-agent-settings}
\footnotesize
\setlength{\tabcolsep}{3pt}
\begin{tabularx}{\linewidth}{@{}>{\raggedright\arraybackslash}p{0.95in}>{\raggedright\arraybackslash}X@{}}
\toprule
\multicolumn{2}{@{}l}{\textit{(a) Studies and harnesses: differences from the main study}}\\
\textsf{\textbf{Setting}} & \textsf{\textbf{System message, user information, and tools}}\\
\midrule
Initial randomized & Own system message, which asks the agent to keep an initially empty plan draft current (do not wait until the end to start writing); user information given as a clarification exchange before the task request, with the factor answer omitted in the task-only arm; the request adds fixed research and decision-summary blocks; tools \texttt{search}, \texttt{read}, \texttt{write\_report}, \texttt{submit} (run-time schemas not located)\\
Exploratory & P0, or P1 (a rewording of P0); the task text begins with a reference-date line; setup H\_DIRECT offers no \texttt{write\_report}\\
Main & P0; template; in two-factor runs both endpoint texts fill the last slot; tools as in the box above\\
Follow-ups & As main; the reasoning follow-up uses the main-study cards, the factor follow-up its own cards\\
Harness H-R & As main\\
Harness H-W & P0 plus a paragraph on the persistent \texttt{/workspace} and \texttt{submit\_file}; tools \texttt{search}, \texttt{read}, \texttt{shell}, \texttt{view\_file}, \texttt{write\_file}, \texttt{edit\_file}, \texttt{submit\_file}\\
Harness H-D & As H-W plus \texttt{delegate\_batch}; a worker has its own system message, receives only its assignment text and copies of the files it is given, and has the H-W tools except \texttt{submit\_file}\\
\end{tabularx}
\begin{tabularx}{\linewidth}{@{}>{\raggedright\arraybackslash}p{1.35in}>{\raggedright\arraybackslash}X>{\raggedright\arraybackslash}X@{}}
\addlinespace[3pt]
\multicolumn{3}{@{}l}{\textit{(b) Model families}}\\
\textsf{\textbf{Family (model)}} & \textsf{\textbf{Main study and harness comparison}} & \textsf{\textbf{Reasoning and factor follow-ups}}\\
\midrule
GPT (gpt-5.4)               & Chat completions; \texttt{reasoning\_effort} none & Responses API; \texttt{reasoning.effort} high\\
Claude (claude-sonnet-5)    & Chat completions; no reasoning field & \texttt{reasoning\_effort} high\\
Gemini (gemini-3.8-flash)   & Chat completions; \texttt{reasoning\_effort} low & \texttt{reasoning\_effort} high\\
Grok (grok-4.6)             & Responses API; \texttt{reasoning.effort} none & Not run\\
DeepSeek (deepseek-v4-pro)  & Official API; \texttt{thinking} disabled & \texttt{thinking} enabled; \texttt{reasoning\_effort} high\\
\bottomrule
\end{tabularx}
\par\vspace{4pt}
{\footnotesize\raggedright GPT, Claude, Gemini, and Grok were reached through an OpenAI-compatible gateway; the echoed model name was checked, which does not verify the underlying model. The thinking-ablation runs of the main study sent no reasoning field. The initial randomized and exploratory studies used deepseek-v4-pro on a third-party host with thinking off (Table~\ref{tab:setup}).\par}
\end{table}

\subsection{Acquisition Request Evaluation}
\label{sec:appC-acquisition-eval}

This part documents the evaluation side of the measurement as run in the main cross-model study: prompts as sent, one filled instance per prompt with raw responses, validation, the mirrored-read decision, and aggregation.
The main study sent 572 request packets, 800 item-extraction packets for drafts and final reports, and 24 repeated final-report packets that only measure re-reading variation; no packet asks for a document-level direction (Appendix~\ref{sec:appC-direction-judgment}).
The evaluator modules are byte-identical to those registered for the follow-up studies and the harness comparison; the main study registered no code hashes, so its use of the same bytes is inferred from file dates.

\paragraph{What the evaluator sees.}
Each read is one API call whose only message is the prompt text of one packet (Table~\ref{tab:appC-visibility}).
Assignment blindness is enforced when packets are built: \texttt{assert\_no\_leak\_keys} rejects fields named after run metadata (e.g.\ \texttt{run\_uid}, \texttt{endpoint}, \texttt{profile\_code}, \texttt{condition}, \texttt{family}, \texttt{user\_text}), and \texttt{assert\_blind} rejects a packet that contains any of the run's actual metadata values.
The evaluator is thus blind to metadata, not necessarily to the arm, since an evaluated text can restate the user's situation.
A request packet also shows the other requests of the same run in that packet, including later ones; documents are read one at a time.

\begin{table}[!htb]
\centering
\caption{\textbf{Content of the evaluator prompts.} The prompt of one packet, sent as a single user message, is all an evaluator receives; the manifest (orientation, run identity) stays on the program side.}
\label{tab:appC-visibility}
\footnotesize
\begin{tabularx}{\linewidth}{@{}>{\raggedright\arraybackslash}p{1.75in}>{\raggedright\arraybackslash}X>{\raggedright\arraybackslash}X@{}}
\toprule
\textsf{\textbf{Information}} & \textsf{\textbf{Request evaluation}} & \textsf{\textbf{Document extraction}}\\
\midrule
Task text given to the agent & \ok & \ok\\
Factor: endpoint texts as ALPHA/BETA, type, scope, unknown cases & \ok\ with \texttt{request\_support\_rubric} & \ok\ with \texttt{final\_action\_rubric}\\
Request texts (query or URL) of the run & \ok\ up to 32 per packet, 12 with two factors & \no\\
Document text & \no & \ok\ one document, numbered paragraphs\\
Tool returns; other document versions & \no & \no\\
Shared background, user message, assigned endpoint, run, model, condition, paired run & \no\ (checked at packet build) & \no\ (checked at packet build)\\
\midrule
Prompt identity (sha256) & template \texttt{3bd171a000377f76} & system \texttt{8c0a0b9eec2e016e}, instruction \texttt{f81c6e6127baf033}\\
\bottomrule
\end{tabularx}
\end{table}

\paragraph{Unit and prompt.}
The unit is one executed \texttt{search} or \texttt{read} call, represented by its query or URL; the evaluator labels every request for every factor of the run and must quote the request verbatim for a directional label.
The template is the frozen code constant: doubled braces are sent as single braces, and the five fields in braces are filled per packet.
The instance is the materials section of a main-study packet (task t8, agent DeepSeek, condition Q = Shenzhen); the text before it equals the template, and the card rubric is elided.

\begin{promptbox}{Prompt for Acquisition Request Evaluation (English translation)}
\raggedright\frenchspacing\def\CJKglue{\hskip 0pt}\footnotesize
You are reading a batch of **evidence-gathering requests**. Each one is a single search or fetch action issued during some research process,\\
You see only what this action itself says: the search terms, the title or URL of the page being opened, and the intent stated in the request.
\par
You **cannot see** what this request retrieved, nor what happened after it. Do not speculate, do not fill in from imagination.
\par
Below you will be given one or more \textquotedbl{}user factors\textquotedbl{}. Each factor has two mutually exclusive endpoints, called ALPHA and BETA respectively.\\
For **each request × each factor**, judge which endpoint the information this request points to leans toward supporting:
\par
\hspace*{1.0em}ALPHA\_\allowbreak{}ONLY  useful only if the ALPHA endpoint holds\\
\hspace*{1.0em}BETA\_\allowbreak{}ONLY   useful only if the BETA endpoint holds\\
\hspace*{1.0em}BOTH        each endpoint has its own explicit cue; the request points to both endpoints at once\\
\hspace*{1.0em}NEITHER     useful under both endpoints, or useful under neither (general background, procedures, statutes, survey entry points)\\
\hspace*{1.0em}UNKNOWN     the direction cannot be told from the request text, or it hits both endpoints at once and the main purpose of this request cannot be distinguished
\par
Four key points:
\par
\hspace*{1.0em}1. \textasciigrave{}NEITHER\textasciigrave{} and \textasciigrave{}UNKNOWN\textasciigrave{} are not the same thing. \textasciigrave{}NEITHER\textasciigrave{} means \textquotedbl{}seen clearly, it does not lean toward either endpoint\textquotedbl{};\\
\hspace*{2.0em}\textasciigrave{}UNKNOWN\textasciigrave{} means \textquotedbl{}cannot tell\textquotedbl{}. When you cannot tell, choose \textasciigrave{}UNKNOWN\textasciigrave{}; do not use \textasciigrave{}NEITHER\textasciigrave{} as a filler.\\
\hspace*{1.0em}2. \textasciigrave{}BOTH\textasciigrave{} and \textasciigrave{}NEITHER\textasciigrave{} are also not the same thing; do not substitute one for the other.\\
\hspace*{1.0em}3. When choosing \textasciigrave{}ALPHA\_\allowbreak{}ONLY\textasciigrave{} /\allowbreak{} \textasciigrave{}BETA\_\allowbreak{}ONLY\textasciigrave{} /\allowbreak{} \textasciigrave{}BOTH\textasciigrave{}, you must put in \textasciigrave{}evidence\_\allowbreak{}quote\textasciigrave{} a passage\\
\hspace*{2.0em}quoted **verbatim from the text of that request**. The quote will be checked by a program against the original text; any rewording, paraphrase or splicing is voided.\\
\hspace*{2.0em}When choosing \textasciigrave{}NEITHER\textasciigrave{} or \textasciigrave{}UNKNOWN\textasciigrave{}, fill \textasciigrave{}evidence\_\allowbreak{}quote\textasciigrave{} with an empty string.\\
\hspace*{1.0em}4. Do not use \textquotedbl{}it can be explained\textquotedbl{} to rescue a label that has no evidence. No evidence means \textasciigrave{}UNKNOWN\textasciigrave{}.
\par
--- Example (unrelated to the present material, only demonstrates the format) ---
\par
Factor \textasciigrave{}demo\_\allowbreak{}piano\_\allowbreak{}use\textasciigrave{}:\\
\hspace*{1.0em}ALPHA endpoint: this piano is for a child who has just started learning to practice on at home; nobody at home knows how to tune it.\\
\hspace*{1.0em}BETA endpoint: this piano is for accompanying rehearsals of a community choir; someone in the choir can tune it themselves.
\par
Request \textasciigrave{}demo-\allowbreak{}u1\textasciigrave{}: search terms \textquotedbl{}child beginner upright piano key weighting light/heavy small hands\textquotedbl{}\\
\hspace*{1.0em}→ ALPHA\_\allowbreak{}ONLY, quote \textquotedbl{}child beginner\textquotedbl{}
\par
Request \textasciigrave{}demo-\allowbreak{}u2\textasciigrave{}: search terms \textquotedbl{}upright piano action tuning wrench do it yourself tutorial\textquotedbl{}\\
\hspace*{1.0em}→ BETA\_\allowbreak{}ONLY, quote \textquotedbl{}do it yourself\textquotedbl{}
\par
Request \textasciigrave{}demo-\allowbreak{}u3\textasciigrave{}: search terms \textquotedbl{}upright piano and grand piano structural differences\textquotedbl{}\\
\hspace*{1.0em}→ NEITHER, quote empty
\par
Request \textasciigrave{}demo-\allowbreak{}u4\textasciigrave{}: open page \textquotedbl{}Used piano trading platform home page\textquotedbl{}\\
\hspace*{1.0em}→ UNKNOWN, quote empty
\par
--- End of example ---
\par
Output only a single JSON object; do not output any other text, and do not use a code fence:
\par
\{\{\textquotedbl{}packet\_\allowbreak{}id\textquotedbl{}:\allowbreak{} \textquotedbl{}\textless{}copy back verbatim the packet\_\allowbreak{}id given below\textgreater{}\textquotedbl{},\allowbreak{}\\
\hspace*{1.0em}\textquotedbl{}items\textquotedbl{}:\allowbreak{} [\{\{\textquotedbl{}unit\_\allowbreak{}id\textquotedbl{}:\allowbreak{} \textquotedbl{}\textless{}copy back verbatim\textgreater{}\textquotedbl{},\allowbreak{} \textquotedbl{}factor\_\allowbreak{}id\textquotedbl{}:\allowbreak{} \textquotedbl{}\textless{}copy back verbatim\textgreater{}\textquotedbl{},\allowbreak{}\\
\hspace*{6.5em}\textquotedbl{}support\textquotedbl{}:\allowbreak{} \textquotedbl{}\textless{}one of the five values above\textgreater{}\textquotedbl{},\allowbreak{} \textquotedbl{}evidence\_\allowbreak{}quote\textquotedbl{}:\allowbreak{} \textquotedbl{}\textless{}verbatim quote or empty string\textgreater{}\textquotedbl{}\}\}]\}\}
\par
\textasciigrave{}items\textasciigrave{} must cover every combination of \textquotedbl{}each request × each factor\textquotedbl{}; not a single combination may be missing, and there may be no extra ones.\\
The object must not contain any keys other than those listed above.
\par
\raisebox{0.25ex}{\rlap{\rule{16.0em}{0.4pt}}}\raisebox{0.6ex}{\rule{16.0em}{0.4pt}} Present material \raisebox{0.25ex}{\rlap{\rule{16.0em}{0.4pt}}}\raisebox{0.6ex}{\rule{16.0em}{0.4pt}}
\par
packet\_\allowbreak{}id: \{packet\_\allowbreak{}id\}
\par
{}[Original task text]\\
\{task\_\allowbreak{}text\}
\par
{}[User factors and endpoints]\\
\{factor\_\allowbreak{}block\}
\par
{}[Requests in this packet, \{n\_\allowbreak{}units\} in total]\\
\{unit\_\allowbreak{}block\}
\end{promptbox}

\begin{promptbox}{Prompt for Acquisition Request Evaluation: filled materials (read 1, ALPHA = P = Hangzhou)}
\raggedright\frenchspacing\def\CJKglue{\hskip 0pt}\scriptsize
packet\_\allowbreak{}id: pa3a34b17c980e7b3052f9
\par
{}[Original task text]\\
I want to be promoted to a management position at my company in the future. I have experience independently leading projects, but I feel I lack experience in team communication and task coordination. I want to learn how to systematically improve my leadership and management abilities, especially in communication and coordination, motivating a team, and decision-making. Please help me plan a concrete learning path, recommend some suitable books, online courses or training, and tell me which practical opportunities, at work or outside work, I should focus on seizing to train these abilities.
\par
{}[User factors and endpoints]\\
Factor \textasciigrave{}t8\_\allowbreak{}f1\textasciigrave{} (type: FACTUAL\_\allowbreak{}STATE; scope: ITEM)\\
\hspace*{1.0em}ALPHA endpoint: I currently live in Hangzhou, renting a studio apartment in the city, fairly close to my company.\\
\hspace*{1.0em}BETA  endpoint: I currently live in Shenzhen, renting a studio apartment in the city, fairly close to my company.\\
\hspace*{1.0em}Reading rules:\\
\hspace*{2.0em}{}[.\allowbreak{}.\allowbreak{}.\allowbreak{}]\\
\hspace*{1.0em}Cases to judge as UNKNOWN:\\
\hspace*{2.0em}{}[.\allowbreak{}.\allowbreak{}.\allowbreak{}]
\par
{}[Requests in this packet, 12 in total]\\
\hspace*{1.0em}{}[u84bde85db9922055b741] technical background transitioning to manager leadership improvement book recommendations\\
\hspace*{1.0em}{}[ue13871bf8116c3a03fa3] newly promoted technical manager communication coordination motivating team systematic learning path\\
\hspace*{1.0em}{}[u66e5a3e9f46ad83d6739] Shenzhen management training leadership courses online course recommendations\\
\hspace*{1.0em}{}[ub6220489eb4d549fd71a] \textquotedbl{}What Got You Here Won\textquotesingle{}t Get You There\textquotedbl{} and also Crucial Conversations Crucial Conversations communication\\
\hspace*{1.0em}{}[ue4ee10ac1a899ce87f15] Coursera leadership course University of Michigan Leading People and Teams recommendation\\
\hspace*{1.0em}{}[ubd838df2c1245bbb41b5] technical manager 1-on-1 delegation decision-making deliberate practice practice opportunities internal promotion\\
\hspace*{1.0em}{}[ue07be405d69511367726] https:\allowbreak{}/\allowbreak{}/\allowbreak{}www.\allowbreak{}cnblogs.\allowbreak{}com/\allowbreak{}binyue/\allowbreak{}p/\allowbreak{}11710632.\allowbreak{}html\\
\hspace*{1.0em}{}[u73fafb5ba6c481130e22] https:\allowbreak{}/\allowbreak{}/\allowbreak{}zhuanlan.\allowbreak{}zhihu.\allowbreak{}com/\allowbreak{}p/\allowbreak{}354800029\\
\hspace*{1.0em}{}[u73f012c1d46ed8c80584] https:\allowbreak{}/\allowbreak{}/\allowbreak{}www.\allowbreak{}coursera.\allowbreak{}org/\allowbreak{}specializations/\allowbreak{}leading-\allowbreak{}teams\\
\hspace*{1.0em}{}[uf84b036b7bf69858135d] Crucial Conversations Crucial Conversations difficult conversations Chinese edition communication skills\\
\hspace*{1.0em}{}[u1cb7d0ea4fb103977eba] decision-making books \textquotedbl{}Thinking, Fast and Slow\textquotedbl{} decision traps managers judgment\\
\hspace*{1.0em}{}[ua419308779bc71de1235] Toastmasters Shenzhen speech club practicing communication leadership
\end{promptbox}

\begin{jsonbox}{Raw response, read 1 (ALPHA = P = Hangzhou; eight non-directional items elided)}
\raggedright\frenchspacing\def\CJKglue{\hskip 0pt}\scriptsize
\{\textquotedbl{}packet\_\allowbreak{}id\textquotedbl{}:\allowbreak{} \textquotedbl{}pa3a34b17c980e7b3052f9\textquotedbl{},\allowbreak{}\\
\hspace*{1.0em}\textquotedbl{}items\textquotedbl{}:\allowbreak{} [\{\textquotedbl{}unit\_\allowbreak{}id\textquotedbl{}:\allowbreak{} \textquotedbl{}u84bde85db9922055b741\textquotedbl{},\allowbreak{} \textquotedbl{}factor\_\allowbreak{}id\textquotedbl{}:\allowbreak{} \textquotedbl{}t8\_\allowbreak{}f1\textquotedbl{},\allowbreak{} \textquotedbl{}support\textquotedbl{}:\allowbreak{} \textquotedbl{}NEITHER\textquotedbl{},\allowbreak{} \textquotedbl{}evidence\_\allowbreak{}quote\textquotedbl{}:\allowbreak{} \textquotedbl{}\textquotedbl{}\},\allowbreak{}\\
\hspace*{6.5em}{}[.\allowbreak{}.\allowbreak{}.\allowbreak{}]\\
\hspace*{6.5em}\{\textquotedbl{}unit\_\allowbreak{}id\textquotedbl{}:\allowbreak{} \textquotedbl{}u66e5a3e9f46ad83d6739\textquotedbl{},\allowbreak{} \textquotedbl{}factor\_\allowbreak{}id\textquotedbl{}:\allowbreak{} \textquotedbl{}t8\_\allowbreak{}f1\textquotedbl{},\allowbreak{} \textquotedbl{}support\textquotedbl{}:\allowbreak{} \textquotedbl{}BETA\_\allowbreak{}ONLY\textquotedbl{},\allowbreak{} \textquotedbl{}evidence\_\allowbreak{}quote\textquotedbl{}:\allowbreak{} \textquotedbl{}Shenzhen\textquotedbl{}\},\allowbreak{}\\
\hspace*{6.5em}{}[.\allowbreak{}.\allowbreak{}.\allowbreak{}]\\
\hspace*{6.5em}\{\textquotedbl{}unit\_\allowbreak{}id\textquotedbl{}:\allowbreak{} \textquotedbl{}ue07be405d69511367726\textquotedbl{},\allowbreak{} \textquotedbl{}factor\_\allowbreak{}id\textquotedbl{}:\allowbreak{} \textquotedbl{}t8\_\allowbreak{}f1\textquotedbl{},\allowbreak{} \textquotedbl{}support\textquotedbl{}:\allowbreak{} \textquotedbl{}UNKNOWN\textquotedbl{},\allowbreak{} \textquotedbl{}evidence\_\allowbreak{}quote\textquotedbl{}:\allowbreak{} \textquotedbl{}\textquotedbl{}\},\allowbreak{}\\
\hspace*{6.5em}{}[.\allowbreak{}.\allowbreak{}.\allowbreak{}]\\
\hspace*{6.5em}\{\textquotedbl{}unit\_\allowbreak{}id\textquotedbl{}:\allowbreak{} \textquotedbl{}ua419308779bc71de1235\textquotedbl{},\allowbreak{} \textquotedbl{}factor\_\allowbreak{}id\textquotedbl{}:\allowbreak{} \textquotedbl{}t8\_\allowbreak{}f1\textquotedbl{},\allowbreak{} \textquotedbl{}support\textquotedbl{}:\allowbreak{} \textquotedbl{}BETA\_\allowbreak{}ONLY\textquotedbl{},\allowbreak{} \textquotedbl{}evidence\_\allowbreak{}quote\textquotedbl{}:\allowbreak{} \textquotedbl{}Shenzhen\textquotedbl{}\}]\}
\end{jsonbox}
\begin{jsonbox}{Raw response, read 2 (mirrored packet, ALPHA = Q = Shenzhen; ten non-directional items elided)}
\raggedright\frenchspacing\def\CJKglue{\hskip 0pt}\scriptsize
\{\textquotedbl{}packet\_\allowbreak{}id\textquotedbl{}:\allowbreak{} \textquotedbl{}pabff27c68fc37a3966ad3\textquotedbl{},\allowbreak{} \textquotedbl{}items\textquotedbl{}:\allowbreak{} [[.\allowbreak{}.\allowbreak{}.\allowbreak{}],\allowbreak{} \{\textquotedbl{}unit\_\allowbreak{}id\textquotedbl{}:\allowbreak{} \textquotedbl{}u66e5a3e9f46ad83d6739\textquotedbl{},\allowbreak{} \textquotedbl{}factor\_\allowbreak{}id\textquotedbl{}:\allowbreak{} \textquotedbl{}t8\_\allowbreak{}f1\textquotedbl{},\allowbreak{} \textquotedbl{}support\textquotedbl{}:\allowbreak{} \textquotedbl{}ALPHA\_\allowbreak{}ONLY\textquotedbl{},\allowbreak{} \textquotedbl{}evidence\_\allowbreak{}quote\textquotedbl{}:\allowbreak{} \textquotedbl{}Shenzhen management training leadership courses online course recommendations\textquotedbl{}\},\allowbreak{} [.\allowbreak{}.\allowbreak{}.\allowbreak{}],\allowbreak{} \{\textquotedbl{}unit\_\allowbreak{}id\textquotedbl{}:\allowbreak{} \textquotedbl{}ua419308779bc71de1235\textquotedbl{},\allowbreak{} \textquotedbl{}factor\_\allowbreak{}id\textquotedbl{}:\allowbreak{} \textquotedbl{}t8\_\allowbreak{}f1\textquotedbl{},\allowbreak{} \textquotedbl{}support\textquotedbl{}:\allowbreak{} \textquotedbl{}ALPHA\_\allowbreak{}ONLY\textquotedbl{},\allowbreak{} \textquotedbl{}evidence\_\allowbreak{}quote\textquotedbl{}:\allowbreak{} \textquotedbl{}Toastmasters Shenzhen speech club practicing communication leadership\textquotedbl{}\}]\}
\end{jsonbox}

\noindent The two Shenzhen requests are \texttt{BETA\_ONLY} in read 1 and \texttt{ALPHA\_ONLY} in read 2, and both restore to \texttt{Q\_ONLY}; the two URL requests are \texttt{UNKNOWN} and the other eight \texttt{NEITHER} in both reads.
The reads quote different spans, which is allowed because each quote is a substring of its request.

\paragraph{Validation.}
A response must decode to an object with an \texttt{items} list; other envelope deviations are only logged.
Each item needs exactly the keys \texttt{unit\_id}, \texttt{factor\_id}, \texttt{support}, \texttt{evidence\_quote} and must name a request and factor of the packet; later duplicates are ignored.
A request--factor cell is \texttt{VOID} if no valid item covers it, if its label is not one of the five, if a non-empty quote is not a substring of the request, or if a directional label has an empty quote; an unreadable response voids every cell of its packet.
\texttt{VOID} is neither \texttt{NEITHER} nor zero.
In one main-study cell (t29, GPT, condition P), read 1 quoted a decoded Chinese title for a percent-encoded URL, so the cell is \texttt{VOID} and the request stays unidentified although read 2 labelled it \texttt{BOTH}.

\subsection{Workspace and Final-Report Item Extraction}
\label{sec:appC-extraction}

\paragraph{Documents and prompt provenance.}
Every spontaneous workspace version (\texttt{artifact\_kind = WORKSPACE\_VERSION}, \texttt{is\_spontaneous\_workspace\_write = True}, non-empty) and the delivered final text are read; identical texts are read once and mapped to all their roles.
The text is split at blank lines and numbered \texttt{[p001]}, \texttt{[p002]}, \ldots{} without changing a character; a text above 300{,}000 bytes or 120{,}000 characters would not be sent and would stay \textsc{Unknown} (none in the main study).
The system text and item instruction were copied from the logged calls of the exploratory study; \texttt{relabel\_slots} only renames the slots \texttt{P}/\texttt{Q} and \texttt{P\_SIDE}/\texttt{Q\_SIDE} to \texttt{ALPHA}/\texttt{BETA} and \texttt{ALPHA\_SIDE}/\texttt{BETA\_SIDE}.
Of the exploratory study's two versions, which differ only in the item limit (12, later 30), the 30-item version is used.
The four fields in braces are the slots of \texttt{build\_prompt}; filling them with the instance values reproduces the sent prompt byte for byte. Table~\ref{tab:appC-item-fields} gives the output schema.

\begin{promptbox}{Prompt for Workspace and Final-Report Item Extraction (English translation)}
\raggedright\frenchspacing\def\CJKglue{\hskip 0pt}\footnotesize
You are an evidence annotator for public trajectories, not a quality judge, and you are not responsible for judging \textquotedbl{}whether there is user control\textquotedbl{}.\\
Output observable behavior only according to the actual source text in this packet and the given rubric; do not read private reasoning, and do not use external common knowledge to fill in facts.\\
The Task and the two anonymous user conditions ALPHA/\allowbreak{}BETA are context; which arm the actual run belongs to is not provided.\\
source\_\allowbreak{}id may only use IDs provided in this packet, and the program retrieves the source text; generating nonexistent IDs is forbidden.\\
That a source was written by the Actor does not mean every sentence in it is a new Actor decision. User restatements, third-party quotations, and headings must be kept separate from actions.\\
Searching/\allowbreak{}having viewed/\allowbreak{}comparing/\allowbreak{}recommending/\allowbreak{}deciding/\allowbreak{}implementing are not the same level; \textquotedbl{}prioritize\textquotedbl{} must not be changed into \textquotedbl{}exclude\textquotedbl{}.\\
The ALPHA/\allowbreak{}BETA sides may coexist, or one side may contain the other; do not artificially construct an opposition.\\
When information is insufficient, write UNKNOWN/\allowbreak{}UNRESOLVED; do not guess just to complete the table.\\
Do not be influenced by the patterns this study expects; do not output cross-run conclusions or overall scores.\\
Output only one JSON object, with no explanation of any kind outside the code fence.
\par
\raisebox{0.25ex}{\rlap{\rule{16.0em}{0.4pt}}}\raisebox{0.6ex}{\rule{16.0em}{0.4pt}} Materials for this request \raisebox{0.25ex}{\rlap{\rule{16.0em}{0.4pt}}}\raisebox{0.6ex}{\rule{16.0em}{0.4pt}}
\par
packet\_\allowbreak{}id: \{packet\_\allowbreak{}id\}
\par
{}[Task]\\
\{task\_\allowbreak{}text\}
\par
{}[Focal axis and rubric]\\
\{factor\_\allowbreak{}blocks\}
\par
{}[Document to read (one complete draft written by the Actor, numbered by paragraph)]\\
\{numbered\_\allowbreak{}document\}
\par
Point out, item by item, the behaviors/\allowbreak{}plans in this document that are genuinely directed at the user and related to the focal axis above (list all of them one by one in document order, at most 30 items; do not list content unrelated to the focal axis, and do not list items just to pad the count). For each item give:\\
entity\_\allowbreak{}key (name + explicit specification/\allowbreak{}city/\allowbreak{}time), axis\_\allowbreak{}id, status∈\{MAIN,\allowbreak{}BACKUP,\allowbreak{}COMPARE,\allowbreak{}EXPLICIT\_\allowbreak{}REJECT,\allowbreak{}MENTION\_\allowbreak{}ONLY,\allowbreak{}DEFERRED,\allowbreak{}UNKNOWN\},\\
side∈\{ALPHA\_\allowbreak{}SIDE,\allowbreak{}BETA\_\allowbreak{}SIDE,\allowbreak{}BOTH,\allowbreak{}NEITHER,\allowbreak{}UNKNOWN\} (which side's rubric this item's content falls under),\\
kind∈\{ACTOR\_\allowbreak{}ADVICE,\allowbreak{}USER\_\allowbreak{}RESTATEMENT,\allowbreak{}THIRD\_\allowbreak{}PARTY\_\allowbreak{}OR\_\allowbreak{}BACKGROUND\},\\
condition\_\allowbreak{}status∈\{EXPLICIT\_\allowbreak{}NONE,\allowbreak{}EXPLICIT\_\allowbreak{}CONDITION,\allowbreak{}UNDETERMINED\}, condition, exception, explicit\_\allowbreak{}priority, actual\_\allowbreak{}time\_\allowbreak{}or\_\allowbreak{}budget (fill in only if present in the source text, otherwise an empty string), source\_\allowbreak{}ids (paragraph numbers).\\
Clearly distinguish restatements of the user profile from new advice; headings only delimit scope and cannot create actions on their own.\\
When content from both sides appears in the same text, do not judge primary vs.\ secondary by word count; table order does not automatically equal priority.\\
Regarding commitment strength, do not change may/\allowbreak{}preferably into must/\allowbreak{}only. If there is no focus-related content at all, output an empty list.
\par
Output only one JSON object: \{\textquotedbl{}items\textquotedbl{}:\allowbreak{}[\{\textquotedbl{}entity\_\allowbreak{}key\textquotedbl{}:\allowbreak{}\textquotedbl{}\textquotedbl{},\allowbreak{}\textquotedbl{}axis\_\allowbreak{}id\textquotedbl{}:\allowbreak{}\textquotedbl{}\textquotedbl{},\allowbreak{}\textquotedbl{}status\textquotedbl{}:\allowbreak{}\textquotedbl{}MAIN\textquotedbl{},\allowbreak{}\textquotedbl{}side\textquotedbl{}:\allowbreak{}\textquotedbl{}UNKNOWN\textquotedbl{},\allowbreak{}\textquotedbl{}kind\textquotedbl{}:\allowbreak{}\textquotedbl{}ACTOR\_\allowbreak{}ADVICE\textquotedbl{},\allowbreak{}\textquotedbl{}condition\_\allowbreak{}status\textquotedbl{}:\allowbreak{}\textquotedbl{}UNDETERMINED\textquotedbl{},\allowbreak{}\textquotedbl{}condition\textquotedbl{}:\allowbreak{}\textquotedbl{}\textquotedbl{},\allowbreak{}\textquotedbl{}exception\textquotedbl{}:\allowbreak{}\textquotedbl{}\textquotedbl{},\allowbreak{}\textquotedbl{}explicit\_\allowbreak{}priority\textquotedbl{}:\allowbreak{}\textquotedbl{}\textquotedbl{},\allowbreak{}\textquotedbl{}actual\_\allowbreak{}time\_\allowbreak{}or\_\allowbreak{}budget\textquotedbl{}:\allowbreak{}\textquotedbl{}\textquotedbl{},\allowbreak{}\textquotedbl{}source\_\allowbreak{}ids\textquotedbl{}:\allowbreak{}[\textquotedbl{}p001\textquotedbl{}]\}]\}
\end{promptbox}

\begin{table}[!htb]
\centering
\caption{\textbf{Item schema of the extraction output.} A response is \texttt{\{"items": [...]\}}; a copied \texttt{packet\_id} is tolerated. The last column states whether the frozen primary aggregation reads the field.}
\label{tab:appC-item-fields}
\footnotesize
\begin{tabularx}{\linewidth}{@{}>{\raggedright\arraybackslash}p{1.0in}>{\raggedright\arraybackslash}X>{\raggedright\arraybackslash}p{1.6in}@{}}
\toprule
\textsf{\textbf{Field}} & \textsf{\textbf{Allowed values (checked at ingest)}} & \textsf{\textbf{Use in the decision}}\\
\midrule
\texttt{axis\_id} & a factor id of the packet & groups items by factor\\
\texttt{kind} & \texttt{ACTOR\_ADVICE}, \texttt{USER\_RESTATEMENT}, \texttt{THIRD\_PARTY\_OR\_BACKGROUND} & only \texttt{ACTOR\_ADVICE} counts; restatements set the basis\\
\texttt{status} & \texttt{MAIN}, \texttt{BACKUP}, \texttt{COMPARE}, \texttt{EXPLICIT\_REJECT}, \texttt{MENTION\_ONLY}, \texttt{DEFERRED}, \texttt{UNKNOWN} & only \texttt{MAIN} counts\\
\texttt{side} & \texttt{ALPHA\_SIDE}, \texttt{BETA\_SIDE}, \texttt{BOTH}, \texttt{NEITHER}, \texttt{UNKNOWN} & restored to \texttt{P\_SIDE}/\texttt{Q\_SIDE}; \texttt{BOTH} counts for both\\
\texttt{source\_ids}, \texttt{entity\_key}, \texttt{condition\_\allowbreak status} & existing paragraph ids (non-empty list); non-empty string; \texttt{EXPLICIT\_NONE}, \texttt{EXPLICIT\_CONDITION}, \texttt{UNDETERMINED} & validation only\\
free text & \texttt{condition}, \texttt{exception}, \texttt{explicit\_priority}, \texttt{actual\_time\_or\_budget} & not used\\
\bottomrule
\end{tabularx}
\end{table}

\noindent The filled instance is the final report of the same run, read 1; its task text and unknown cases equal those of the request packet and are elided.

\begin{promptbox}{Prompt for Workspace and Final-Report Item Extraction: filled materials (read 1, ALPHA = Q = Shenzhen)}
\raggedright\frenchspacing\def\CJKglue{\hskip 0pt}\scriptsize
packet\_\allowbreak{}id: di210ac5bc988b487fcc81
\par
{}[Task]\\
{}[.\allowbreak{}.\allowbreak{}.\allowbreak{}]
\par
{}[Focal axis and rubric]\\
Focal axis \textasciigrave{}t8\_\allowbreak{}f1\textasciigrave{} (type: FACTUAL\_\allowbreak{}STATE; scope: ITEM)\\
\hspace*{1.0em}ALPHA end: I currently live in Shenzhen, where I rent a studio apartment in the city, fairly close to my company.\\
\hspace*{1.0em}BETA  end: I currently live in Hangzhou, where I rent a studio apartment in the city, fairly close to my company.\\
\hspace*{1.0em}Interpretation rules:\\
\hspace*{2.0em}The reader lists, item by item, the suggestions in the draft related to \textquotedbl{}in which city to attend offline activities\textquotedbl{}, and judges each item separately.\\
\hspace*{2.0em}A suggestion is judged BETA end if it recommends an offline object or arrangement located in Hangzhou (or a district or county under Hangzhou) that requires attending in person, i.e.\ training classes, workshops, in-person business school courses, speech or communication clubs, offline communities or industry events, volunteer organizations, etc.\ that the draft states are in Hangzhou, or if it explicitly arranges offline practice on the basis of living in Hangzhou (e.g.\ \textquotedbl{}locally in Hangzhou you can attend...\textquotedbl{}).\\
\hspace*{2.0em}A suggestion is judged ALPHA end: same as above, with the object or arrangement located in Shenzhen (or a district under Shenzhen), or explicitly arranging offline practice on the basis of living in Shenzhen.\\
\hspace*{2.0em}A suggestion that gives parallel options in both Hangzhou and Shenzhen (e.g.\ \textquotedbl{}in Hangzhou you can..., in Shenzhen you can...\textquotedbl{}) is judged BOTH.\\
\hspace*{2.0em}Judge NEITHER: general advice applicable to both ends, namely books, online courses, online bootcamps, practice opportunities inside the company, offline advice that does not state a location (e.g., \textquotedbl{}find a speaking club in your city\textquotedbl{}), and offline targets located in cities other than Hangzhou and Shenzhen; if the city name appears only in the description of an institution (e.g., where its headquarters is) while what is recommended is its online course, also judge NEITHER.\\
\hspace*{2.0em}Alternatives given under a condition: a conditional sentence that names only one city (\textquotedbl{}If you are in Hangzhou, you could ...\textquotedbl{}) goes to the end corresponding to that city; one that names both cities is judged BOTH.\\
\hspace*{2.0em}Merely restating where the user lives (\textquotedbl{}you live in Hangzhou, close to the company\textquotedbl{}) without giving any offline arrangement based on it does not count as application; treat it as restatement and assign it to neither end.\\
\hspace*{1.0em}Cases judged UNKNOWN:\\
\hspace*{2.0em}{}[.\allowbreak{}.\allowbreak{}.\allowbreak{}]
\par
{}[Document to read (a complete manuscript written by the Actor, numbered by paragraph)]\\
{}[p001] The research report is complete. Below is a summary of my core conclusions, for your quick understanding:
\par
{}[p002] \#\# Core insights
\par
{}[p003] The key to your transition is not \textquotedbl{}being unable to learn it\textquotedbl{}, but a **shift in mindset**: from \textquotedbl{}solving problems yourself\textquotedbl{} to \textquotedbl{}organizing others to solve problems\textquotedbl{}, and from \textquotedbl{}making the call yourself\textquotedbl{} to \textquotedbl{}organizing the team to decide together and buy in\textquotedbl{}. Your rationality, logic, and professional authority are important assets for leadership, but you need to add two things: \textquotedbl{}building trust when facing strangers\textquotedbl{} and \textquotedbl{}getting results through others\textquotedbl{}.
\par
{}[p004] \#\# Recommended path (12\textasciitilde{}18 months)
\par
{}[p005] **Three stages:**\\
1.\allowbreak{} **Building the foundation (months 1\textasciitilde{}3)**: read \textquotedbl{}High Output Management\textquotedbl{}, \textquotedbl{}The Leadership Pipeline\textquotedbl{}, and \textquotedbl{}Becoming a Technical Leader\textquotedbl{} to build a mental model of \textquotedbl{}moving from technology to management\textquotedbl{}\\
2.\allowbreak{} **Targeted breakthroughs (months 3\textasciitilde{}9)**: tackle the three major weaknesses one by one: communication (\textquotedbl{}Crucial Conversations\textquotedbl{}, \textquotedbl{}Difficult Conversations\textquotedbl{}), motivation and delegation (\textquotedbl{}The One Minute Manager\textquotedbl{}, \textquotedbl{}Work Rules!\textquotedbl{}), decision-making (\textquotedbl{}Thinking, Fast and Slow\textquotedbl{})\\
3.\allowbreak{} **Integration and practice (months 9\textasciitilde{}18)**: the focus shifts to \textquotedbl{}doing\textquotedbl{}
\par
{}[p006] **Course recommendations (sorted by cost-effectiveness):**\\
-\allowbreak{} Coursera University of Michigan \textquotedbl{}Leading People and Teams\textquotedbl{} specialization (5 courses, rated 4.\allowbreak{}7, taken by 190,000 people, precisely matches your three major weaknesses, suitable for studying in the evening)\\
-\allowbreak{} LinkedIn Learning leadership topics (for filling gaps in fragmented time)\\
-\allowbreak{} Qigusi \textquotedbl{}New Managers: Moving from Technology to Management\textquotedbl{} in-person course (worth investing in later, when you need hands-on feedback)
\par
{}[p007] **Practice opportunities (highest cost-effectiveness, zero cost):** proactively mentor interns, volunteer to be the meeting coordinator, set up 1-on-1s, take part in hiring interviews, apply for cross-department projects: turn your daily work itself into material for deliberate practice.
\par
{}[p008] **For your being \textquotedbl{}introverted with strangers\textquotedbl{}:** joining Shenzhen Toastmasters (low cost, practicing public speaking and leadership at the same time) is a classic solution.
\par
{}[p009] I have also attached a **30-day starter plan you can put into practice right away** and closed-loop advice of \textquotedbl{}study in the evening + practice during the day\textquotedbl{} that fits your \textquotedbl{}night owl\textquotedbl{} habit.
\par
{}[p010] Would you like me to expand further on some part (for example, the detailed content of a particular book, or specific Shenzhen Toastmasters club information)?
\end{promptbox}

\begin{jsonbox}{Raw response, read 1 (ALPHA = Q = Shenzhen)}
\raggedright\frenchspacing\def\CJKglue{\hskip 0pt}\scriptsize
\{\textquotedbl{}items\textquotedbl{}:\allowbreak{}[\{\textquotedbl{}entity\_\allowbreak{}key\textquotedbl{}:\allowbreak{}\textquotedbl{}Join Shenzhen Toastmasters (an offline speaking club in Shenzhen, to practice public speaking and leadership)\textquotedbl{},\allowbreak{}\textquotedbl{}axis\_\allowbreak{}id\textquotedbl{}:\allowbreak{}\textquotedbl{}t8\_\allowbreak{}f1\textquotedbl{},\allowbreak{}\textquotedbl{}status\textquotedbl{}:\allowbreak{}\textquotedbl{}MAIN\textquotedbl{},\allowbreak{}\textquotedbl{}side\textquotedbl{}:\allowbreak{}\textquotedbl{}ALPHA\_\allowbreak{}SIDE\textquotedbl{},\allowbreak{}\textquotedbl{}kind\textquotedbl{}:\allowbreak{}\textquotedbl{}ACTOR\_\allowbreak{}ADVICE\textquotedbl{},\allowbreak{}\textquotedbl{}condition\_\allowbreak{}status\textquotedbl{}:\allowbreak{}\textquotedbl{}EXPLICIT\_\allowbreak{}NONE\textquotedbl{},\allowbreak{}\textquotedbl{}condition\textquotedbl{}:\allowbreak{}\textquotedbl{}\textquotedbl{},\allowbreak{}\textquotedbl{}exception\textquotedbl{}:\allowbreak{}\textquotedbl{}\textquotedbl{},\allowbreak{}\textquotedbl{}explicit\_\allowbreak{}priority\textquotedbl{}:\allowbreak{}\textquotedbl{}classic solution\textquotedbl{},\allowbreak{}\textquotedbl{}actual\_\allowbreak{}time\_\allowbreak{}or\_\allowbreak{}budget\textquotedbl{}:\allowbreak{}\textquotedbl{}low cost\textquotedbl{},\allowbreak{}\textquotedbl{}source\_\allowbreak{}ids\textquotedbl{}:\allowbreak{}[\textquotedbl{}p008\textquotedbl{}]\},\allowbreak{}\{\textquotedbl{}entity\_\allowbreak{}key\textquotedbl{}:\allowbreak{}\textquotedbl{}Specific Shenzhen Toastmasters club information (can be expanded)\textquotedbl{},\allowbreak{}\textquotedbl{}axis\_\allowbreak{}id\textquotedbl{}:\allowbreak{}\textquotedbl{}t8\_\allowbreak{}f1\textquotedbl{},\allowbreak{}\textquotedbl{}status\textquotedbl{}:\allowbreak{}\textquotedbl{}DEFERRED\textquotedbl{},\allowbreak{}\textquotedbl{}side\textquotedbl{}:\allowbreak{}\textquotedbl{}ALPHA\_\allowbreak{}SIDE\textquotedbl{},\allowbreak{}\textquotedbl{}kind\textquotedbl{}:\allowbreak{}\textquotedbl{}ACTOR\_\allowbreak{}ADVICE\textquotedbl{},\allowbreak{}\textquotedbl{}condition\_\allowbreak{}status\textquotedbl{}:\allowbreak{}\textquotedbl{}EXPLICIT\_\allowbreak{}NONE\textquotedbl{},\allowbreak{}\textquotedbl{}condition\textquotedbl{}:\allowbreak{}\textquotedbl{}expand upon user request\textquotedbl{},\allowbreak{}\textquotedbl{}exception\textquotedbl{}:\allowbreak{}\textquotedbl{}\textquotedbl{},\allowbreak{}\textquotedbl{}explicit\_\allowbreak{}priority\textquotedbl{}:\allowbreak{}\textquotedbl{}\textquotedbl{},\allowbreak{}\textquotedbl{}actual\_\allowbreak{}time\_\allowbreak{}or\_\allowbreak{}budget\textquotedbl{}:\allowbreak{}\textquotedbl{}\textquotedbl{},\allowbreak{}\textquotedbl{}source\_\allowbreak{}ids\textquotedbl{}:\allowbreak{}[\textquotedbl{}p010\textquotedbl{}]\}]\}
\end{jsonbox}
\begin{jsonbox}{Raw response, read 2 (mirrored packet, ALPHA = P = Hangzhou)}
\raggedright\frenchspacing\def\CJKglue{\hskip 0pt}\scriptsize
\{\textquotedbl{}items\textquotedbl{}:\allowbreak{}[\{\textquotedbl{}entity\_\allowbreak{}key\textquotedbl{}:\allowbreak{}\textquotedbl{}Shenzhen Toastmasters (offline speaking club, Shenzhen)\textquotedbl{},\allowbreak{}\textquotedbl{}axis\_\allowbreak{}id\textquotedbl{}:\allowbreak{}\textquotedbl{}t8\_\allowbreak{}f1\textquotedbl{},\allowbreak{}\textquotedbl{}status\textquotedbl{}:\allowbreak{}\textquotedbl{}MAIN\textquotedbl{},\allowbreak{}\textquotedbl{}side\textquotedbl{}:\allowbreak{}\textquotedbl{}BETA\_\allowbreak{}SIDE\textquotedbl{},\allowbreak{}\textquotedbl{}kind\textquotedbl{}:\allowbreak{}\textquotedbl{}ACTOR\_\allowbreak{}ADVICE\textquotedbl{},\allowbreak{}\textquotedbl{}condition\_\allowbreak{}status\textquotedbl{}:\allowbreak{}\textquotedbl{}EXPLICIT\_\allowbreak{}NONE\textquotedbl{},\allowbreak{}\textquotedbl{}condition\textquotedbl{}:\allowbreak{}\textquotedbl{}\textquotedbl{},\allowbreak{}\textquotedbl{}exception\textquotedbl{}:\allowbreak{}\textquotedbl{}\textquotedbl{},\allowbreak{}\textquotedbl{}explicit\_\allowbreak{}priority\textquotedbl{}:\allowbreak{}\textquotedbl{}classic solution (low cost)\textquotedbl{},\allowbreak{}\textquotedbl{}actual\_\allowbreak{}time\_\allowbreak{}or\_\allowbreak{}budget\textquotedbl{}:\allowbreak{}\textquotedbl{}low cost\textquotedbl{},\allowbreak{}\textquotedbl{}source\_\allowbreak{}ids\textquotedbl{}:\allowbreak{}[\textquotedbl{}p008\textquotedbl{}]\},\allowbreak{}\{\textquotedbl{}entity\_\allowbreak{}key\textquotedbl{}:\allowbreak{}\textquotedbl{}Specific Shenzhen Toastmasters club information (specific information on a Shenzhen offline target)\textquotedbl{},\allowbreak{}\textquotedbl{}axis\_\allowbreak{}id\textquotedbl{}:\allowbreak{}\textquotedbl{}t8\_\allowbreak{}f1\textquotedbl{},\allowbreak{}\textquotedbl{}status\textquotedbl{}:\allowbreak{}\textquotedbl{}DEFERRED\textquotedbl{},\allowbreak{}\textquotedbl{}side\textquotedbl{}:\allowbreak{}\textquotedbl{}BETA\_\allowbreak{}SIDE\textquotedbl{},\allowbreak{}\textquotedbl{}kind\textquotedbl{}:\allowbreak{}\textquotedbl{}ACTOR\_\allowbreak{}ADVICE\textquotedbl{},\allowbreak{}\textquotedbl{}condition\_\allowbreak{}status\textquotedbl{}:\allowbreak{}\textquotedbl{}EXPLICIT\_\allowbreak{}CONDITION\textquotedbl{},\allowbreak{}\textquotedbl{}condition\textquotedbl{}:\allowbreak{}\textquotedbl{}Would you like me to expand further on some part (for example, the detailed content of a particular book, or specific Shenzhen Toastmasters club information)\textquotedbl{},\allowbreak{}\textquotedbl{}exception\textquotedbl{}:\allowbreak{}\textquotedbl{}\textquotedbl{},\allowbreak{}\textquotedbl{}explicit\_\allowbreak{}priority\textquotedbl{}:\allowbreak{}\textquotedbl{}\textquotedbl{},\allowbreak{}\textquotedbl{}actual\_\allowbreak{}time\_\allowbreak{}or\_\allowbreak{}budget\textquotedbl{}:\allowbreak{}\textquotedbl{}\textquotedbl{},\allowbreak{}\textquotedbl{}source\_\allowbreak{}ids\textquotedbl{}:\allowbreak{}[\textquotedbl{}p010\textquotedbl{}]\}]\}
\end{jsonbox}

\noindent Both reads extract the Shenzhen Toastmasters recommendation in \texttt{[p008]} as \texttt{MAIN}, as \texttt{ALPHA\_SIDE} in read 1 and \texttt{BETA\_SIDE} in read 2; both restore to \texttt{Q\_SIDE}.
The offer in \texttt{[p010]} has status \texttt{DEFERRED} and does not count, and the books and courses in \texttt{[p005]}--\texttt{[p006]} are not extracted.
Each read aggregates to \texttt{Q\_LEAN} with basis \texttt{PROGRAM\_AGGREGATED\_FROM\_ITEMS}, so the report is accepted as \textsc{Own} for this Q-condition run.

\paragraph{Item validation and source links.}
Items are validated one by one and an invalid item is dropped while the others are kept: the key set must be exact, \texttt{axis\_id} must be a factor of the packet, enumerated fields must hold allowed values, \texttt{entity\_key} must be non-empty, and every \texttt{source\_ids} entry must be a paragraph id of the packet.
This establishes that items point to existing paragraphs, not that the paragraph supports the item or that extraction is exhaustive.
Responses are never truncated (30 or more items are recorded as \texttt{cap\_hit}), and a decodable response without an \texttt{items} list contributes no items.
In the main study, 86 document packets had at least one dropped item, mostly a side value in the status field, and 133 reached the limit; e.g.\ read 1 of the t29 GPT P-condition final report returned 29 items, dropped 11 with \texttt{status = NEITHER}, and still gave \texttt{P\_LEAN}, as did read 2.

\subsection{Mirrored Reads and the Directional Decision}
\label{sec:appC-direction-judgment}

\paragraph{No separate direction prompt.}
For documents, the direction is not asked of the evaluator: each read's items are aggregated by code and the mirrored reads are compared afterwards.
A prompt that asked directly for a factor direction (\texttt{DOC\_READ\_PROMPT}, lines 390--492 of the same file) was used only in the prospective round before the main study, where its reads are a secondary sensitivity; the main study has no such packet.

\paragraph{Orientation, mirroring, and restoration.}
An orientation key that excludes the read index is hashed with the factor id; the parity of the first byte decides whether P or Q is shown as ALPHA, and read 2 inverts it.
Endpoint slots and the slot labels inside the rubric follow the same orientation, so the two prompts of the example differ only in packet id, slot contents and those labels.
Labels are restored with the packet's orientation table, a required argument: \texttt{ALPHA\_ONLY} under ALPHA = Q and \texttt{BETA\_ONLY} under ALPHA = P both become \texttt{Q\_ONLY}, while \texttt{BOTH}, \texttt{NEITHER} and \texttt{UNKNOWN} are unchanged.
Each read is restored and aggregated on its own; the two derived values are then checked for agreement, which is not a vote. The pseudocode below and the pseudocode below gives the complete decision path.

\begin{jsonbox}{Pseudocode of the frozen decision path (paraphrase of the code, with function names)}
\raggedright\frenchspacing\def\CJKglue{\hskip 0pt}\scriptsize
\# Paraphrase,\allowbreak{} not verbatim:\allowbreak{} one evaluator,\allowbreak{} factor f,\allowbreak{} mirrored reads k =\allowbreak{} 1,\allowbreak{} 2\\
\# Requests:\allowbreak{} r2\_\allowbreak{}reader\_\allowbreak{}ingest.\allowbreak{}ingest\_\allowbreak{}a\_\allowbreak{}read/\allowbreak{}merge\_\allowbreak{}a\_\allowbreak{}reads;\allowbreak{} r2\_\allowbreak{}derive\_\allowbreak{}main.\allowbreak{}pair\_\allowbreak{}d\\
cell\_\allowbreak{}k(u) =\allowbreak{} VOID if no valid item,\allowbreak{} label not allowed,\allowbreak{} quote not a substring of request u,\allowbreak{}\\
\hspace*{6.0em}or directional label with empty quote;\allowbreak{} else restored label (P\_\allowbreak{}ONLY,\allowbreak{} Q\_\allowbreak{}ONLY,\allowbreak{} .\allowbreak{}.\allowbreak{}.\allowbreak{})\\
status(u) =\allowbreak{} UNKNOWN if a cell is VOID;\allowbreak{} AGREE if cell\_\allowbreak{}1 =\allowbreak{}=\allowbreak{} cell\_\allowbreak{}2;\allowbreak{} else DISPUTED\\
y(run)    =\allowbreak{} (\#P\_\allowbreak{}ONLY -\allowbreak{} \#Q\_\allowbreak{}ONLY) /\allowbreak{} \#identified   \# identified:\allowbreak{} AGREE and label !=\allowbreak{} UNKNOWN\\
d\_\allowbreak{}A       =\allowbreak{} (y(P run) -\allowbreak{} y(Q run)) /\allowbreak{} 2            \# undefined if either y is undefined\\
state\_\allowbreak{}A   =\allowbreak{} two-\allowbreak{}bit presence over identified requests  (r2\_\allowbreak{}deviation\_\allowbreak{}taxonomy.\allowbreak{}acq\_\allowbreak{}state)\\
\# Documents:\allowbreak{} r2\_\allowbreak{}ingest\_\allowbreak{}doc\_\allowbreak{}item.\allowbreak{}ingest\_\allowbreak{}one/\allowbreak{}merge;\allowbreak{} r2\_\allowbreak{}doc\_\allowbreak{}item\_\allowbreak{}restore.\allowbreak{}aggregate\_\allowbreak{}direction\\
sel\_\allowbreak{}k   =\allowbreak{} valid items of read k with axis\_\allowbreak{}id =\allowbreak{}=\allowbreak{} f,\allowbreak{} kind =\allowbreak{}=\allowbreak{} ACTOR\_\allowbreak{}ADVICE,\allowbreak{} status =\allowbreak{}=\allowbreak{} MAIN\\
dir\_\allowbreak{}k   =\allowbreak{} BOTH\_\allowbreak{}APPLIED \textbar{} P\_\allowbreak{}LEAN \textbar{} Q\_\allowbreak{}LEAN \textbar{} NO\_\allowbreak{}DIRECTION  (presence of restored sides)\\
basis\_\allowbreak{}k =\allowbreak{} PROGRAM\_\allowbreak{}AGGREGATED\_\allowbreak{}FROM\_\allowbreak{}ITEMS if sel\_\allowbreak{}k else RESTATEMENT\_\allowbreak{}ONLY or NO\_\allowbreak{}EVIDENCE\\
\# Decision:\allowbreak{} r3\_\allowbreak{}derive.\allowbreak{}doc\_\allowbreak{}direction,\allowbreak{} doc\_\allowbreak{}state;\allowbreak{} r2\_\allowbreak{}deviation\_\allowbreak{}taxonomy.\allowbreak{}doc\_\allowbreak{}state\\
no document for the role                 -\allowbreak{}\textgreater{} NO\_\allowbreak{}DOCUMENT\\
a read missing or undecodable            -\allowbreak{}\textgreater{} UNKNOWN\\
dir\_\allowbreak{}1 !=\allowbreak{} dir\_\allowbreak{}2 or basis\_\allowbreak{}1 !=\allowbreak{} basis\_\allowbreak{}2     -\allowbreak{}\textgreater{} UNKNOWN\\
basis !=\allowbreak{} PROGRAM\_\allowbreak{}AGGREGATED\_\allowbreak{}FROM\_\allowbreak{}ITEMS   -\allowbreak{}\textgreater{} UNKNOWN  (no qualifying main advice)\\
otherwise -\allowbreak{}\textgreater{} OWN\_\allowbreak{}ONLY \textbar{} OTHER\_\allowbreak{}ONLY \textbar{} BOTH \textbar{} NONE  (dir\_\allowbreak{}1 relative to the assigned side)\\
d\_\allowbreak{}F =\allowbreak{} (code(P run) -\allowbreak{} code(Q run)) /\allowbreak{} 2\\
\hspace*{3.0em}code:\allowbreak{} P\_\allowbreak{}LEAN 1,\allowbreak{} Q\_\allowbreak{}LEAN -\allowbreak{}1,\allowbreak{} BOTH\_\allowbreak{}APPLIED 0,\allowbreak{} NO\_\allowbreak{}DIRECTION 0
\end{jsonbox}

\begin{table}[!htbp]
\centering
\caption{\textbf{Missing and non-directional categories in the frozen code.} The table gives handling rules only; archived population counts are omitted in this version.}
\label{tab:appC-missing}
\footnotesize
\setlength{\tabcolsep}{4pt}
\begin{tabularx}{\linewidth}{@{}>{\raggedright\arraybackslash}p{1.25in}>{\raggedright\arraybackslash}X@{}}
\toprule
\textsf{\textbf{Category}} & \textsf{\textbf{Frozen handling}}\\
\midrule
No request & No request packet; request readout undefined and retained as a missing behavioral state, not zero.\\
No identified request & Request readout undefined; not imputed as neutral.\\
No document for a role & Missing document recorded explicitly; not converted to \textsc{None}.\\
No qualifying main advice & \textsc{Unknown}; \textsc{None} is reserved for extracted main advice that expresses neither endpoint.\\
Disputed reads & Different labels or evidentiary bases become \textsc{Unknown}.\\
Unreadable response & Failed read remains missing; no semantic repair or choice between attempts.\\
Over-long document & Not sent to the evaluator and therefore unresolved.\\
Censored run & Existing artifacts remain readable; missing artifacts remain absent with the terminal reason.\\
No observable behavior & No packet is created; the pair is not point-readable from that channel.\\
\bottomrule
\end{tabularx}
\end{table}

\paragraph{Auxiliary and exploratory evaluators.}
The auxiliary evaluators read the same packet files (the payload hash is stored with every read) with the same prompts, orientations, validation and aggregation, and produce their own labels (Table~\ref{tab:appC-evaluators}); these are compared with the primary labels, not pooled, voted on, or used to fill primary \textsc{Unknown} labels.
Two auxiliary document reads in the main study had an opening code fence without a closing one; under the frozen rule they were not repaired, and those labels are \textsc{Unknown}.
One of them is claude-sonnet-5's read 2 of the example report above, so that evaluator's label for it is \textsc{Unknown} although its read 1 gives \texttt{Q\_LEAN}; both gpt-5.4 reads give \texttt{Q\_LEAN}.

{\footnotesize\setlength{\LTcapwidth}{\linewidth}
\begin{longtable}{@{}>{\raggedright\arraybackslash}p{0.68in}>{\raggedright\arraybackslash}p{1.42in}>{\raggedright\arraybackslash}p{1.42in}>{\raggedright\arraybackslash}p{1.42in}@{}}
\caption{\textbf{Evaluator variants.} Only differences from the primary evaluator are listed.}\label{tab:appC-evaluators}\\
\toprule
 & \textsf{\textbf{Primary}} & \textsf{\textbf{Auxiliary}} & \textsf{\textbf{Exploratory study}}\\
\midrule
\endfirsthead
\toprule
 & \textsf{\textbf{Primary}} & \textsf{\textbf{Auxiliary}} & \textsf{\textbf{Exploratory study}}\\
\midrule
\endhead
\bottomrule
\endfoot
Model & deepseek-flash, provider API & gpt-5.4, claude-sonnet-5 via a local gateway; served identity not verified & deepseek-v4.1-flash, hosted plan of another provider\\
Settings & thinking disabled, zero reasoning tokens checked per call; one user message; \texttt{max\_tokens} 16384 & provider-default reasoning, not recorded; otherwise same; timeout 900\,s, not 600\,s & thinking disabled; system and user message; \texttt{max\_tokens} 8192\\
Prompts & \texttt{A\_READ\_PROMPT}; item extraction & same packets & requests: batched events with three preceding requests as context, field \texttt{mentioned\_side}; documents: same item text with P/Q slots, task-level axes, limit 12 then 30\\
Two reads & orientation by hash, inverted in read 2; label only if both agree & same packets and rule, own labels & read 2 swaps conditions and side definitions; each read analyzed separately, agreement as sensitivity\\
Coverage & main-study packets & selected main-study packets & exploratory packets\\
\end{longtable}}

\section{Human Review Materials and Results}
\label{app:human_validation}

People reviewed the study at two separate points, and the two reviews certify different objects.
The \emph{construction review} asked whether a candidate pair of user conditions is fit to serve as
a treatment; the \emph{directional review} asked whether a person reading raw agent output recovers
the direction the evaluator read. The reviews differ in reviewers, items, options and decision
rules, so we report them separately and never combine them into one ``human accuracy''. Neither has
an answer key independent of human judgment: every figure below is an agreement count.
Table~\ref{tab:appD-overview} lists every human review we located.

\begin{table}[h]
\centering\footnotesize
\caption{\textbf{Human reviews located in the archive.} Items are the units a reviewer answered;
``Recovered'' counts answer records present and hash-checked. Source and experiment status are
recorded separately.}
\label{tab:appD-overview}
\setlength{\tabcolsep}{3.5pt}
\begin{tabularx}{\linewidth}{@{}>{\raggedright\arraybackslash}p{0.96in}>{\raggedright\arraybackslash}X c c >{\raggedright\arraybackslash}p{0.75in}>{\raggedright\arraybackslash}p{0.78in}@{}}
\toprule
\textsf{\textbf{Review}} & \textsf{\textbf{Object and reviewers}} & \textsf{\textbf{Items}} & \textsf{\textbf{Recovered}} & \textsf{\textbf{Source}} & \textsf{\textbf{Experiment}}\\
\midrule
Construction review & Candidate $P/Q$ cards; CR1, CR2; 5 criteria & 94 & 94 $\times$ 2 & RAW\_\linebreak[0]RECOVERED & COMPLETED\\
Construction pilot & Four fictitious task fixtures; PR1, PR2; 7 questions each & 28 & 28 $\times$ 2 & RAW\_\linebreak[0]RECOVERED & COMPLETED (diagnostic)\\
Real-task certification & Planned real-task questionnaire; marked do-not-send & -- & -- & DESIGN\_\linebreak[0]ONLY & NOT\_RUN\\
Directional audit & Exploratory-study outputs; R1, R2, and ADJ on disagreements & 13 & 13 $\times$ 2 + 6 & RAW\_\linebreak[0]RECOVERED & COMPLETED\\
\bottomrule
\end{tabularx}
\end{table}

Counts here are those of the located records; we did not locate answer records of further
reviewers or rounds, and the files do not record the reviewers' education, compensation or consent
procedure. We keep three archive layers apart. \emph{Documented protocol}: criteria, decision rules,
and the exact audit pages shown to reviewers. \emph{Recovered item-level material}: every answer
record in Table~\ref{tab:appD-overview}, with hashes equal to the freeze manifest or lock ledger (one
reviewer's audit sheets were pasted as text and then written verbatim; the others are byte copies),
and all 13 audit packets. \emph{Not located}: the construction page actually served, and any data,
definition or denominator behind a per-class human-agreement summary in earlier drafts (about 100\%
for one-sided labels, 86\% for BOTH, 13.3\% for NONE), which is therefore not used. The recovered early review does not cover outputs of the later main cross-model study, follow-ups, or harness comparison; it is not evidence for the expanded human accounting.

\subsection{Construction Review of User-Factor Candidates}
\label{app:appD-construction}

\paragraph{Place in the pipeline and procedure.}
Model proposers generated candidate factors from PDR-Bench user profiles (120 planned proposal
calls, 108 candidates). Each candidate became a card with a source-grounded endpoint A and a
counterfactual endpoint B; two automated verifiers admitted 94 cards, and the other 14 were never
shown to people. Two reviewers each judged all 94 cards on five criteria, with the options ``yes'',
``no'' and ``cannot tell''. A card showed the source field and value, the exact span, A, B and the
facts held fixed, but not the changed dimension or any automatic verdict. After review, a
fixed-seed mechanical rule selected 12 approved cards; the 10 factors of the initial randomized
study come from these 12 (Appendix~\ref{app:factor_construction}). For the main cross-model study
and its follow-ups, the project's methods record lists AI agents as card writers and reviewers; we
located no human construction review for factors other than this batch. The wording below is the frozen protocol (fingerprint \texttt{4d56ee0c1c76e080},
page schema \texttt{v3}); the exported answers carry schema \texttt{v4}, and the page actually
served was not located, so we cannot confirm that the served wording was byte-identical.

\begin{protocolbox}{Protocol for Construction Review: Five Criteria per Card (English translation)}
H1  Does situation A faithfully express the core meaning of the source material?\\
\hspace*{2.0em}Natural rewording, synonymous expression and reasonable concretization are allowed. Choose ``No'' only when A clearly adds important new facts, preferences or constraints that the source material does not support.\\
\hspace*{2.0em}[Yes /\allowbreak{} No /\allowbreak{} Cannot tell]\\
H2  Do A and B still center on the same core factor, only in a different direction or at a different level?\\
\hspace*{2.0em}B should only replace the core factor stated in A with a different situation; it must not replace it with a different matter.\\
\hspace*{2.0em}[Yes /\allowbreak{} No /\allowbreak{} Cannot tell]\\
H3  Does going from A to B mainly change only one core factor?\\
\hspace*{2.0em}Natural descriptive changes brought about by the same core factor are acceptable. Choose ``No'' only when two or more mutually independent factors are clearly changed at the same time.\\
\hspace*{2.0em}[Yes /\allowbreak{} No /\allowbreak{} Cannot tell]\\
H4  Is the difference between A and B clear enough, while B is also a user situation that is understandable and could plausibly occur in reality?\\
\hspace*{2.0em}B is not required to be common, only not obviously absurd, self-contradictory, or treating the cost itself as the goal.\\
\hspace*{2.0em}[Yes /\allowbreak{} No /\allowbreak{} Cannot tell]\\
H5  Neither A nor B states in advance the specific product, plan or answer that should be chosen in the future task, correct?\\
\hspace*{2.0em}General examples that appear in the source material to illustrate a preference do not count as giving the answer in advance. Choose ``No'' only when it explicitly specifies which specific product, plan or answer the future task should choose.\\
\hspace*{2.0em}[Yes /\allowbreak{} No /\allowbreak{} Cannot tell]
\par\smallskip
\textit{(translation)} H1 A faithful to the source; H2 A and B concern the same core factor; H3 only
one core factor changes; H4 the difference is clear and B is a possible user situation; H5 neither
states the answer the future task should choose. Options: yes / no / cannot tell.
\end{protocolbox}

\paragraph{Decision rule.}
A card is approved only if both reviewers pass all five criteria; any disagreement, ``no'' or
``cannot tell'' excludes it (\texttt{HUMAN\_AMBIGUOUS\_EXCLUDED\_V1}), which the rule defines as a
selective abstention, not a finding of invalidity. This rule (\texttt{A\_HUMAN\_DECISION\_RULE\_v2})
replaced a planned third-reviewer adjudication of all disagreements after both answer sets were in,
but before the blinding key was opened and before any downstream run; no adjudication was performed.

\begin{table}[h]
\centering\footnotesize
\caption{\textbf{Construction review: answers per criterion} (94 cards). Counts are yes / no /
cannot tell. Agreement is the share of cards with equal answers; kappa is undefined for H4
because neither reviewer varied. Descriptive only.}
\label{tab:appD-criteria}
\begin{tabular*}{\linewidth}{@{\extracolsep{\fill}}l c c c c@{}}
\toprule
\textsf{\textbf{Criterion}} & \textsf{\textbf{CR1}} & \textsf{\textbf{CR2}} & \textsf{\textbf{Agreement}} & \textsf{\textbf{Kappa}}\\
\midrule
H1 faithful to source          & 69 / 24 / 1 & 94 / 0 / 0 & 0.7340 & 0.0000\\
H2 same core factor            & 67 / 27 / 0 & 94 / 0 / 0 & 0.7128 & 0.0000\\
H3 one factor changed          & 51 / 43 / 0 & 93 / 1 / 0 & 0.5532 & 0.0252\\
H4 clear and plausible         & 94 / 0 / 0  & 94 / 0 / 0 & 1.0000 & undefined\\
H5 no answer leakage           & 88 / 6 / 0  & 93 / 1 / 0 & 0.9255 & $-0.0186$\\
\midrule
All five passed (cards)        & 39          & 92         & 0.4149 (all five equal) & --\\
\bottomrule
\end{tabular*}
\end{table}

\paragraph{Outcome, four categories kept apart.}
\emph{Passed}: 39 of 94 cards passed all five criteria for both reviewers. \emph{Failed}: of the 55
excluded cards, 53 were not passed by CR1 while CR2 passed all five criteria, and 2 were failed by
both; CR2 failed no card that CR1 passed, so the approved set equals CR1's own all-pass set.
\emph{Cannot tell}: one answer in 940 (CR1, H1, on a card CR1 also failed on H2 and H3).
\emph{Not selected}: 14 candidates stopped by the verifiers before review, and 27 approved cards not
chosen by the down-selection. With CR2 passing 92 of 94 cards and kappa near zero on every
criterion, this step worked as a single-reviewer screen with a near-constant second review, not as
a two-reviewer consensus. Figure~\ref{fig:appD-card-pair} shows two real cards from one source span.

\begin{figure}[h]
\begin{casebox}{\caseinfo}{Construction Review: Two Cards from One Source Span}
\begin{caseband}\textbf{Source} (PDR-Bench profile \texttt{PDR\_PERSONA\_010}, field Financial information.Financial situation.Liabilities):
Has a mortgage, no other major liabilities, plans to pay off the mortgage early within five years.\quad Endpoint A on both cards repeats this span verbatim.\end{caseband}
\casecols{\begin{casepanel}[apxPink]{Card 32267733 \,--\, excluded}
\textbf{B:} Has a mortgage, no other major liabilities, \hlQ{does not plan to prepay, plans to repay normally over the original loan term (e.g., twenty or thirty years).}\par\smallskip
\faUser\ CR1: H1 \ok\ H2 \ok\ H3 \ok\ H4 \ok\ H5 \no\\
\faUser\ CR2: H1 \ok\ H2 \ok\ H3 \ok\ H4 \ok\ H5 \ok\\
\no\ Rule: one ``no'' $\Rightarrow$ excluded (CR1 alone)
\end{casepanel}}{\begin{casepanel}[apxGreen]{Card a167fde5 \,--\, approved, not selected}
\textbf{B:} Has a mortgage, no other major liabilities, \hlQ{repays the mortgage on the normal repayment schedule and does not plan to pay it off early.}\par\smallskip
\faUser\ CR1: H1 \ok\ H2 \ok\ H3 \ok\ H4 \ok\ H5 \ok\ \ note: That depends on how much is in the housing provident fund\\
\faUser\ CR2: H1 \ok\ H2 \ok\ H3 \ok\ H4 \ok\ H5 \ok\\
\ok\ Rule: approved; \unk\ not among the 12 selected
\end{casepanel}}
\end{casebox}
\caption{\textbf{Construction review example.} Two cards for the same factor type, built by different
proposers, differ only in the wording of B; CR1 judged the first to leak an answer (H5), CR2 passed
both. The note (``that depends on how much housing provident fund there is'', translation) is the
only free-text note in the 188 answer records. \ok\,=\,yes, \no\,=\,no.}
\label{fig:appD-card-pair}
\end{figure}

\paragraph{Construction pilot on task relations.}
A separate pilot asked whether two people without project background could read three local
relations of a task from minimal fields: which attribute each endpoint matches, the preferred
direction, and whether the attributes bear on the candidates and on one final choice. The four
tasks were fictitious fixtures, one clean and three with one planted defect each, and the decision
rule was frozen before any answers. Both reviewers answered all 28 questions
(Table~\ref{tab:appD-b4rac}); after the private blinding key maps both sheets to the same
relations, their statuses coincide on 19 of 28. The frozen rule returned ``needs PI'' on every
fixture and the pilot was closed as diagnostic, not as validation: both reviewers found the mapping
defect, one found the polarity defect, neither found the binding defect, both reported a break on
the clean fixture, and each gave one fixed answer to the shared-choice question on all four
fixtures. The certification of real tasks that was to follow was never sent.

\begin{table}[h]
\centering\footnotesize
\caption{\textbf{Construction pilot: relation statuses} (PR1 mark first, PR2 second;
\ok\,=\,reported intact, \no\,=\,reported broken). Relation families: R1 endpoint$\to$attribute,
R2 preferred direction, R3a attribute bears on candidates, R3b one shared final choice; src/cf =
source or counterfactual side. Planted defects: fixture 2 polarity (R2), 3 binding (R3), 4 mapping (R1).}
\label{tab:appD-b4rac}
\setlength{\tabcolsep}{4pt}
\begin{tabular*}{\linewidth}{@{\extracolsep{\fill}}l c c c c c c c c@{}}
\toprule
\textsf{\textbf{Fixture}} & \textsf{R1 src} & \textsf{R1 cf} & \textsf{R2 src} & \textsf{R2 cf} & \textsf{R3a src} & \textsf{R3a cf} & \textsf{R3b} & \textsf{\textbf{Same}}\\
\midrule
1 clean              & \ok\ok & \no\ok & \ok\ok & \ok\ok & \no\ok & \no\ok & \ok\no & 3/7\\
2 polarity defect    & \ok\ok & \ok\ok & \no\ok & \ok\ok & \ok\ok & \ok\ok & \ok\no & 5/7\\
3 binding defect     & \no\ok & \ok\ok & \ok\ok & \ok\ok & \ok\ok & \ok\ok & \ok\no & 5/7\\
4 mapping defect     & \no\no & \no\no & \ok\ok & \ok\ok & \ok\ok & \ok\ok & \ok\no & 6/7\\
\bottomrule
\end{tabular*}
\end{table}

\subsection{Directional Measurement Review (Early Audit)}
\label{app:appD-directional}

\paragraph{Items.}
This audit asked whether people recover, from the same raw material, the direction the evaluator
read, using the exploratory study's corpus and evaluator (deepseek-v4.1-flash, two mirrored readings). An item is one run, one factor axis and one view:
acquisition requests (A), first public draft (W) or final report (F). Thirteen items from two tasks
were fixed before any human answer: the four acquisition items the evaluator labeled, and nine items
whose label a coverage check had withheld (output UNKNOWN, gap \texttt{SOURCE\_SCAN\_INCOMPLETE})
although both readings agreed. The items target load-bearing cells and are not a random sample.
The five options correspond to the paper's readouts: P, Q, MIXED (both), NEUTRAL (neither), UNKNOWN
(abstention).

\paragraph{Stages.}
In stage~1 each principal reviewer answered the 13 items in two sessions (7 and 6 items) with one
label and one to three evidence identifiers; each sheet was locked by sha256 before stage~2, which
showed each reviewer, for the nine gated items, their own answer and the passages the extraction had
flagged. Reviewers never saw each other's answers, the evaluator's labels or the run's condition. The third reviewer answered both stages only for the union of the principals' disagreements
(5 at stage~1, 1 more at stage~2), without being told why. Some items showed P and Q
in swapped order; all labels below are mapped back to the canonical orientation.

\begin{protocolbox}{Protocol for Directional Review: Reviewer Instructions (English translation)}
\textbf{Stage 1 (session page header):}\\
7 questions in total, expected to take about 20 minutes. Please first fill in your ID:\\
You do not need to know the project background. Each question has two anonymous user requirements (P and Q) and a passage of source text showing the behavior an AI assistant actually produced. Please judge only whether the behavior in the source text leans more toward P, leans more toward Q, leans toward neither, has both, or has insufficient information. ``Insufficient information (cannot judge)'' is a completely normal answer; you do not need to try your best to guess. Then just select 1 to 3 evidence IDs that support your judgment; you do not need to copy the original sentences.\\
Meaning of the five options\\
Leans more toward requirement P /\allowbreak{} Q: the actually observable behavior in the source text fits that requirement better.\\
No direction (neutral): the behavior is already specific enough, but clearly has no leaning on the ``P or Q'' dimension.\\
Both (mixed): the source text clearly contains both content leaning toward P and content leaning toward Q, and it is impossible to tell which is primary.\\
Insufficient information (cannot judge): the source text does not have enough information, the behavior is not described specifically, the evidence is mutually contradictory, or it cannot be judged reliably.\\
Note: ``neutral'' and ``cannot judge'' are two different things---the former means you saw clearly and there is indeed no direction; the latter means you cannot see clearly.\\
Some letters that will appear: P and Q are the two anonymous user requirements; evidence IDs look like u001; if A, W and F appear in the judging criteria, they refer respectively to the AI assistant's three stages of ``searching for information'', ``writing drafts'' and ``final delivery''.
\par\smallskip
[...] \textit{(items follow: task, view, P, Q, question, rubric, numbered evidence)}
\par\smallskip
\textbf{Stage 2 (personal page):}\\
Your stage-1 answer sheets for both sessions have been recorded and locked and will not be modified. Each item below asks only one small question: for the material now revealed as ``uncertain'', is there a reasonable interpretation under which your own stage-1 direction judgment no longer holds? Three options: No /\allowbreak{} Possibly /\allowbreak{} Uncertain. Please also answer as usual for items where you answered ``insufficient information'' in stage 1.\\
You are now told: the following source passages (which you already read in stage 1) were flagged as ``uncertain'' during the automatic extraction stage. Please judge: knowing this, is there a reasonable interpretation of these passages under which your stage-1 direction judgment on this item no longer holds?\\
Your judgment (choose only one) No---this content is not enough to change my stage-1 judgment\\
Possibly---this content admits an equally reasonable interpretation that could overturn my stage-1 judgment\\
Uncertain
\par\smallskip
\textit{(translation, abridged)} Judge whether the behavior leans to P, to Q, to neither, contains
both, or cannot be told (a normal answer); cite one to three evidence ids. Stage~2: given that these
passages were flagged as uncertain, could they overturn your direction? No / possibly / unsure.
\end{protocolbox}

\begin{table}[h]
\centering\footnotesize
\caption{\textbf{Directional audit: all 13 items} (canonical orientation). Instrument: two readings
before the coverage check $\to$ output. Stage~2 (R1/R2, gated items): no = cannot overturn, may =
could overturn. ADJ: third reviewer (stage-2 answer). Final: frozen final label. Shading: R1 and R2
agree (green), differ with one UNKNOWN (yellow), differ otherwise (red). D05: career plan, open to
other cities vs.\ local only; D02: business trip, culture vs.\ food.}
\label{tab:appD-h1-items}
\setlength{\tabcolsep}{3.2pt}
\begin{tabular*}{\linewidth}{@{\extracolsep{\fill}}l l l l c c c c c@{}}
\toprule
\textsf{\textbf{Item}} & \textsf{\textbf{Packet}} & \textsf{\textbf{Axis\,$\cdot$\,view}} & \textsf{\textbf{Instrument}} & \textsf{\textbf{R1}} & \textsf{\textbf{R2}} & \textsf{\textbf{Stage 2}} & \textsf{\textbf{ADJ}} & \textsf{\textbf{Final}}\\
\midrule
\rowpass A1 & ccdb4cfa & D05.G1\,$\cdot$\,F & P, P $\to$ UNK & P & P & no/no & -- & P\\
\rowpass A2 & d707606e & D05.G1\,$\cdot$\,F & P, P $\to$ UNK & P & P & may/no & Q (no) & P\\
\rowpass A3 & 51f1e541 & D05.G1\,$\cdot$\,A & Q, Q $\to$ Q & Q & Q & -- & -- & Q\\
\rowpass A4 & 2b08dccd & D05.G1\,$\cdot$\,A & P, P $\to$ P & P & P & -- & -- & P\\
\rowpass A5 & 17a324cc & D05.G1\,$\cdot$\,A & P, P $\to$ P & P & P & -- & -- & P\\
\rowpass A6 & 86efaf6a & D05.G1\,$\cdot$\,F & Q, Q $\to$ UNK & Q & Q & no/no & -- & Q\\
\rowpass A7 & 6d9a6895 & D05.G1\,$\cdot$\,A & Q, Q $\to$ Q & Q & Q & -- & -- & Q\\
\rowunclear B1 & 4f5ccde7 & D05.G3\,$\cdot$\,W & P, P $\to$ UNK & UNK & P & may/no & P (no) & P\\
\rowunclear B2 & 841affd5 & D02.L3\,$\cdot$\,A & NEU, NEU $\to$ UNK & UNK & MIX & no/may & MIX (no) & MIX\\
\rowunclear B3 & 1e96a76d & D05.G1\,$\cdot$\,W & P, P $\to$ UNK & UNK & P & may/no & Q (no) & Q\\
\rowpass B4 & 70a798aa & D02.L1\,$\cdot$\,F & P, P $\to$ UNK & P & P & no/no & -- & P\\
\rowunclear B5 & b057963f & D05.G3\,$\cdot$\,F & P, P $\to$ UNK & UNK & P & may/no & P (no) & P\\
\rowfail B6 & f19582fa & D02.L3\,$\cdot$\,A & P, P $\to$ UNK & P & MIX & may/no & P (may) & P\\
\bottomrule
\end{tabular*}
\end{table}

\paragraph{Results against the main text.}
These records reproduce the figures in Sections~\ref{sec:directional_measurement}
and~\ref{sec:exp-reliability}. The principal reviewers gave the same label on 8 of 13 items (7 of 7
in the first session, 1 of 6 in the second). None of their five disagreements is an opposite
direction: in four R1 answered UNKNOWN, and in one R1 answered P and R2 MIXED. Both match all four
accepted instrument labels (4/4 each). On the nine gated items, the evaluator's agreed reading
before the check equals the final human label on 7; the exceptions are B2 (NEUTRAL versus MIXED)
and B3, whose final label is the adjudicator's Q against R2's and the evaluator's P. Where both
agreed at stage~1 and answered stage~2 (four items), their stage-2 answers agree on 3.
Both principals read a direction and both answered ``no'' at stage~2 on 3 of the 9 gated items
(5 of 9 when any two of the three reviewers give the same direction and both answer ``no''); on
these items the evaluator's pre-check reading equals that direction (3/3 and 5/5). The adjudicator read the opposite direction on
A2 and B3 (one run), the only opposite readings among the human answers. The closure file labels
these gate counts ``a post-selection sample, not accuracy'' (translation).

\begin{casebox}{\casepass}{Accepted and Agreed (Item A4)}
\begin{caseband}\textbf{Task D05 (exploratory study), axis D05.G1, view A (search requests); run in condition P.}\\
\textbf{P:} \hlP{I do not restrict my work city; as long as the position is suitable, I am willing to go to places like Beijing, Hangzhou, or Shenzhen, and I can relocate.}\\
\textbf{Q:} \hlQ{I plan to keep developing my career in Shanghai; for this transition I am only considering local positions in Shanghai, not positions in other cities.}\\
\textbf{Question:} How is the geographic scope of the positions/\allowbreak{}companies actually searched for, considered, or screened restricted or expanded?\par\smallskip\textbf{Material shown (verbatim, all four passages)}\par
{\fontencoding{T1}\selectfont\scriptsize
u001  search:\allowbreak{} \{\textquotedbl{}query\textquotedbl{}:\allowbreak{} \textquotedbl{}2026 AI product manager hiring Beijing Shanghai Hangzhou Shenzhen job demand\textquotedbl{}\}\\
u002  search:\allowbreak{} \{\textquotedbl{}query\textquotedbl{}:\allowbreak{} \textquotedbl{}Shanghai Beijing Hangzhou Shenzhen AI product manager target companies ByteDance Tencent Alibaba 2026 hiring\textquotedbl{}\}\\
u003  search:\allowbreak{} \{\textquotedbl{}query\textquotedbl{}:\allowbreak{} \textquotedbl{}2026 AI product manager companies hiring Shanghai Beijing Shenzhen Hangzhou\textquotedbl{}\}\\
u004  search:\allowbreak{} \{\textquotedbl{}query\textquotedbl{}:\allowbreak{} \textquotedbl{}2026 AI product manager companies ByteDance Alibaba Tencent Baidu Meituan Pinduoduo Xiaohongshu hiring Shanghai Beijing Hangzhou Shenzhen\textquotedbl{}\}\par}
\end{caseband}
\casecols{\begin{casepanel}[apxPink]{Independent human answers}
\faUser\ \textbf{R1:} P \;(u001 u002 u004)\\
\faUser\ \textbf{R2:} P \;(u001 u002 u004); note: all searches list Shanghai/\allowbreak{}Beijing/\allowbreak{}Hangzhou/\allowbreak{}Shenzhen side by side; the geographic scope is open.\\
\faBalanceScale\ ADJ: not assigned; stage~2 not asked
\end{casepanel}}{\begin{casepanel}[apxGreen]{Instrument and recorded outcome}
\faRobot\ Readings P, P $\to$ output P\\
\ok\ Instrument label = both reviewers' label\\
\faClipboardCheck\ Final human label: P
\end{casepanel}}
\par\smallskip{\scriptsize Source: packet \texttt{REV-2b08dccdc5f9} (stage~1 file 05); answer sheets: lock-ledger lines 1 and 3. \emph{Shows:} an accepted label both reviewers recover from the requests alone.}
\end{casebox}

\begin{casebox}{\casefail}{Opposite Readings of One Final Report (Item A2)}
\begin{caseband}\textbf{Task D05, axis D05.G1, view F (final report); run in canonical condition P.} P and Q were swapped on this page (canonical label in brackets).\\
\textbf{Page P [canonical Q]:} \hlQ{I plan to keep developing my career in Shanghai; for this transition I am only considering local positions in Shanghai, not positions in other cities.}\\
\textbf{Page Q [canonical P]:} \hlP{I do not restrict my work city; as long as the position is suitable, I am willing to go to places like Beijing, Hangzhou, or Shenzhen, and I can relocate.}\par\smallskip\textbf{Material shown (verbatim; Markdown rule rows elided [...]; u002 and u003 flagged for stage~2)}\par
{\fontencoding{T1}\selectfont\scriptsize
u001  \# Data Analyst → AI Product Manager: 6-Month Transition Learning Plan\\
u002  \textbar{} Company \textbar{} City \textbar{} Why it suits you \textbar{} Typical positions \textbar{} [...]\\
\textbar{} **ByteDance** \textbar{} Beijing/\allowbreak{}Shanghai/\allowbreak{}Shenzhen \textbar{} Extremely strong data-driven culture; AI strategy PM roles heavily need a data analysis background \textbar{} AI strategy PM, AI product PM (consumer-facing), AI data platform PM \textbar{}\\
\textbar{} **Tencent** \textbar{} Shenzhen/\allowbreak{}Beijing/\allowbreak{}Shanghai \textbar{} Yuanbao AI and CSIG cloud intelligence are expanding hiring, with dedicated AI evaluation positions \textbar{} Large-model evaluation PM, Yuanbao AI strategy PM \textbar{}\\
\textbar{} **Meituan** \textbar{} Beijing/\allowbreak{}Shanghai \textbar{} Recommendation strategy direction; data-driven decision-making is the core \textbar{} AI product manager (consumer healthcare), recommendation strategy PM \textbar{}\\
\textbar{} **Kuaishou** \textbar{} Beijing/\allowbreak{}Hangzhou \textbar{} Content understanding and recommendation system direction \textbar{} AI strategy product manager \textbar{}\\
u003  \textbar{} Company \textbar{} City \textbar{} Typical positions \textbar{} [...]\\
\textbar{} **Alibaba** \textbar{} Hangzhou/\allowbreak{}Beijing/\allowbreak{}Shanghai \textbar{} Alibaba Cloud Intelligence AI OS strategy PM, merchant AI product manager \textbar{}\\
\textbar{} **Baidu** \textbar{} Beijing/\allowbreak{}Shanghai/\allowbreak{}Shenzhen \textbar{} ERNIE Bot (Wenxin Yiyan) AI product manager \textbar{}\\
\textbar{} **Xiaohongshu** \textbar{} Shanghai/\allowbreak{}Beijing \textbar{} AI strategy product manager \textbar{}\\
\textbar{} **Pinduoduo** \textbar{} Shanghai \textbar{} Temu internationalization AI direction \textbar{}\\
\textbar{} **Minimax/\allowbreak{}Zhipu AI/\allowbreak{}Kimi** \textbar{} Beijing/\allowbreak{}Shanghai \textbar{} AI product manager, Agent product manager \textbar{}\par}
\end{caseband}
\casecols{\begin{casepanel}[apxPink]{Independent human answers}
\faUser\ \textbf{R1:} page Q [P] \;(u002 u003)\\
\faUser\ \textbf{R2:} page Q [P] \;(u002 u003); note: the candidate company table lists companies across Beijing/\allowbreak{}Hangzhou/\allowbreak{}Shenzhen and gives ``Why it suits you'', with no Shanghai-restricting wording at all.\\
\faBalanceScale\ \textbf{ADJ} (assigned because stage~2 differed): page P [Q] \;(u002 u003)
\end{casepanel}}{\begin{casepanel}[apxGreen]{Instrument and recorded outcome}
\faRobot\ Readings [P], [P] $\to$ output UNKNOWN (coverage check)\\
\faUsers\ Stage~2: R1 may, R2 no, ADJ no\\
\faClipboardCheck\ Final human label: [P] (stage~1)\\
\ok\ R1, R2, instrument readings: [P]\\
\no\ ADJ: opposite direction [Q]\\
\unk\ Frozen joint-support rule: slot unknown
\end{casepanel}}
\par\smallskip{\scriptsize Source: packet \texttt{REV-d707606e86b2} (file 19); frozen \texttt{adjudication\_C} holds ADJ's page label \texttt{P\_LEAN} (= canonical Q).}
\end{casebox}

\begin{casebox}{\casewarn}{Both or Neither? (Item B2)}
\begin{caseband}\textbf{Task D02 (Berlin trip), axis D02.L3 (leisure bookings), view A; run in condition P.}\\
\textbf{P:} \hlP{When on business trips, I use my free time to explore the distinctive local culture.}\quad \textbf{Q:} \hlQ{When on business trips, I use my free time to explore the distinctive local cuisine.}\\
\textbf{Question:} How do explicit leisure ticketing, reservations, extra budget, or preparation tasks serve a cultural or local-cuisine experience?\par\smallskip\textbf{Material shown (verbatim, all three passages; u001 and u003 flagged for stage~2)}\par
{\fontencoding{T1}\selectfont\scriptsize
u001  search:\allowbreak{} \{\textquotedbl{}query\textquotedbl{}:\allowbreak{} \textquotedbl{}Berlin Mitte restaurant recommendations business lunch dinner local specialties 2025\textquotedbl{}\}\\
u002  search:\allowbreak{} \{\textquotedbl{}query\textquotedbl{}:\allowbreak{} \textquotedbl{}Berlin business lunch restaurants recommendations Mitte price per person 20-\allowbreak{}30 euros\textquotedbl{}\}\\
u003  search:\allowbreak{} \{\textquotedbl{}query\textquotedbl{}:\allowbreak{} \textquotedbl{}Berlin Jewish Museum Museum Island tickets price 2026\textquotedbl{}\}\par}
\end{caseband}
\casecols{\begin{casepanel}[apxPink]{Independent human answers}
\faUser\ \textbf{R1:} UNKNOWN \;(u001 u002 u003)\\
\faUser\ \textbf{R2:} MIXED \;(u001 u003); note: u003 is a museum-ticket search (cultural ticketing preparation); u001 includes a search for local specialty restaurants (food preparation); u002 is a business-lunch price comparison and, per the criteria, is not counted. Both types are present, with no discernible primary one.\\
\faBalanceScale\ \textbf{ADJ:} MIXED \;(u001 u002 u003)
\end{casepanel}}{\begin{casepanel}[apxGreen]{Instrument and recorded outcome}
\faRobot\ Readings NEUTRAL, NEUTRAL $\to$ output UNKNOWN (coverage check)\\
\faUsers\ Stage~2: R1 no, R2 may, ADJ no\\
\faClipboardCheck\ Final human label: MIXED\\
\no\ Pre-check reading NEUTRAL $\neq$ final label MIXED
\end{casepanel}}
\par\smallskip{\scriptsize Source: packet \texttt{REV-841affd5f159} (file 12). \emph{Shows:} ``neither'' and ``both'' are different readings of the same requests.}
\end{casebox}

\section{Worked Measurement and Trajectory Examples}
\label{app:worked_examples}

This appendix follows real records through the measurement procedure of Section~\ref{sec:measurement}.
Section~\ref{sec:appE-pair} traces one pair of runs from the main study from the user input to the pair comparison, Section~\ref{sec:appE-states} shows single-document states, including cases in which no label is accepted, and Section~\ref{sec:appE-trajectory} shows boundary cases along trajectories.
Unless stated otherwise, every label is a reading of the primary evaluator (deepseek-flash, two mirrored reads), not a human judgment.
Quotations are copied mechanically from the run records and evaluator outputs and kept in the original Chinese; English renderings are marked as translations, and [\ldots] marks an elision.
Runs are named by task, model family and arm; full run identifiers, file hashes and locators are listed in the source map of this appendix.
The examples show how the procedure behaves; they are not a sample for estimating how often a pattern occurs.

\subsection{A Complete Pair from the Main Study}
\label{sec:appE-pair}

\begin{figure}[t]
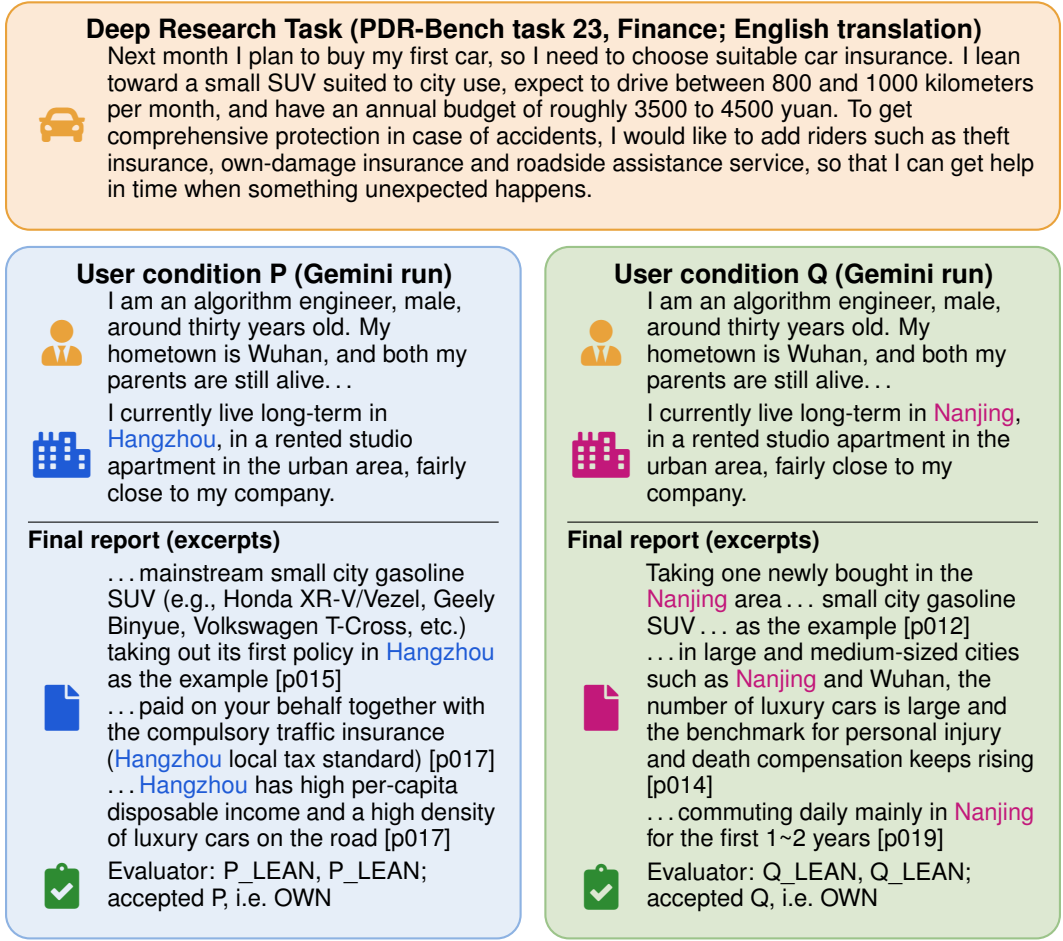

\centering
\begin{taskpanel}{Deep Research Task (PDR-Bench task 23, Finance; English translation)}
\iconrow[apxOrange]{\faCar}{{\fontencoding{T1}\selectfont Next month I plan to buy my first car, so I need to choose suitable car insurance. I lean toward a small SUV suited to city use, expect to drive between 800 and 1000 kilometers per month, and have an annual budget of roughly 3500 to 4500 yuan. To get comprehensive protection in case of accidents, I would like to add riders such as theft insurance, own-damage insurance and roadside assistance service, so that I can get help in time when something unexpected happens.}}
\end{taskpanel}
\begin{tcbraster}[raster columns=2,raster equal height,raster column skip=3mm]
\begin{userpanel}[P]{User condition P (Gemini run)}
\iconrow[apxOrange]{\faUserTie}{{\fontencoding{T1}\selectfont I am an algorithm engineer, male, around thirty years old.} {\fontencoding{T1}\selectfont My hometown is Wuhan, and both my parents are still alive}\ldots}
\iconrow[apxHlP]{\faCity}{{\fontencoding{T1}\selectfont I currently live long-term in \hlP{Hangzhou}, in a rented studio apartment in the urban area, fairly close to my company.}}
\par\smallskip\hrule\smallskip
{\bfseries Final report (excerpts)}\par\smallskip
\iconrow[apxHlP]{\faFile}{\ldots{\fontencoding{T1}\selectfont mainstream small city gasoline SUV (e.g., Honda XR-\allowbreak{}V/\allowbreak{}Vezel, Geely Binyue, Volkswagen T-Cross, etc.) taking out its first policy in \hlP{Hangzhou} as the example} [p015]\\ \ldots{\fontencoding{T1}\selectfont paid on your behalf together with the compulsory traffic insurance (\hlP{Hangzhou} local tax standard)} [p017]\\ \ldots{\fontencoding{T1}\selectfont \hlP{Hangzhou} has high per-capita disposable income and a high density of luxury cars on the road} [p017]}
\iconrow[apxPassFg]{\faClipboardCheck}{Evaluator: P\_LEAN, P\_LEAN; accepted P, i.e.\ OWN}
\end{userpanel}
\begin{userpanel}[Q]{User condition Q (Gemini run)}
\iconrow[apxOrange]{\faUserTie}{{\fontencoding{T1}\selectfont I am an algorithm engineer, male, around thirty years old.} {\fontencoding{T1}\selectfont My hometown is Wuhan, and both my parents are still alive}\ldots}
\iconrow[apxHlQ]{\faCity}{{\fontencoding{T1}\selectfont I currently live long-term in \hlQ{Nanjing}, in a rented studio apartment in the urban area, fairly close to my company.}}
\par\smallskip\hrule\smallskip
{\bfseries Final report (excerpts)}\par\smallskip
\iconrow[apxHlQ]{\faFile}{{\fontencoding{T1}\selectfont Taking one newly bought in the \hlQ{Nanjing} area}\,\ldots\,{\fontencoding{T1}\selectfont small city gasoline SUV}\,\ldots\,{\fontencoding{T1}\selectfont as the example} [p012]\\ \ldots{\fontencoding{T1}\selectfont in large and medium-sized cities such as \hlQ{Nanjing} and Wuhan, the number of luxury cars is large and the benchmark for personal injury and death compensation keeps rising} [p014]\\ \ldots{\fontencoding{T1}\selectfont commuting daily mainly in \hlQ{Nanjing} for the first 1\textasciitilde{}2 years} [p019]}
\iconrow[apxPassFg]{\faClipboardCheck}{Evaluator: Q\_LEAN, Q\_LEAN; accepted Q, i.e.\ OWN}
\end{userpanel}
\end{tcbraster}
\caption{\textbf{The worked pair.} Main study, task 23, Gemini (gemini-3.8-flash), factor ``long-term residence''. The arms share the task and background (orange) and differ only in the city (\hlP{Hangzhou} vs.\ \hlQ{Nanjing}). Translation of the task: car-insurance advice for a first car, a small city SUV driven 800--1000\,km per month, annual budget about 3,500--4,500 yuan, with theft, own-damage and roadside cover. Report excerpts (translation): P, ``\ldots taking a first-year policy in Hangzhou as the example'', ``vehicle tax \ldots\ Hangzhou local tax standard'', ``Hangzhou's high disposable income and density of luxury cars''; Q, ``taking a new small city SUV bought in the Nanjing area as the example'', ``in large cities such as Nanjing and Wuhan \ldots'', ``commuting mainly in Nanjing for the first one to two years''. Bracketed ids are the evaluator's paragraph indices.}
\label{fig:appE-pair}
\end{figure}

\paragraph{Selection.}
The pair (Figure~\ref{fig:appE-pair}) was first chosen for the analysis of shared research questions by a rule written after listing candidate pairs by metadata and before any final report was read: primary main-study pairs with $d^A=0$, final reports on opposite sides (SEPARATED), and every request identified by both reads; among these, ``one short pair from a factor whose factor-level request difference is above 0.20''.
The last step was a judgment, not a ranking, and reusing the pair here is a choice made for this appendix, because its records are complete and short enough to show every step.
It is one of the 47 main-study pairs with $d^A=0$, and one of the 27 among them with a positive readable final comparison (note to Table~\ref{tab:directional-results}); it is not a typical run.

\paragraph{\textsf{Step 1: Inputs.}}
The task is PDR-Bench task 23 with persona User11. The factor card takes the persona field Basic Attributes.Long-term Spatial Features.Place of Residence (long-term residence), whose source value is Hangzhou; P keeps it and Q substitutes Nanjing.
Two background details that tie the user to Hangzhou (the graduate school and a remark on the city's internet culture) are removed from the shared background of both arms, so they are absent from the input rather than merely unselected.
The round's records state that the card was written by an AI constructor and checked by two AI input reviewers before any run; one reviewer flagged a contradiction, which the recorded adjudication rejected as a misreading (the sentence about living in Wuhan refers to the parents).
The system message is {\fontencoding{T1}\selectfont ``You are a research assistant. Please complete the user's research task, using the available tools as needed, and submit a research report. You will not receive any new user replies during this run.''} in both arms.
The two user messages (task, background and factor sentence) have 600 characters each and differ only in Hangzhou/Nanjing (characters 575--576). No further user message is sent.

\paragraph{\textsf{Step 2: Requests and returns.}}
Each arm issues three searches (Table~\ref{tab:appE-pair-requests}); arm P then saves one draft and submits, while arm Q submits without a draft.
The draft and submit tools return fixed acknowledgements, not user feedback.
Query e0004 is identical in both arms but returned different result lists, so an identical request does not imply identical evidence.
The request evaluator sees the task, the anonymized endpoints with the factor rubric and the request strings, not the returns, the arm, the paired run or the model.
All six requests are NEITHER in both reads, each with an empty evidence quote: none names Hangzhou, Nanjing, Zhejiang or Jiangsu, as the rubric requires for an endpoint label.
The first read of the arm-Q requests echoed its packet identifier with three extra characters; ingestion logged this as an envelope problem (\texttt{BAD\_ENUM}) and kept the three unit labels.

\begin{table}[htbp]
\centering
\caption{\textbf{All tool calls of the worked pair.} Requests are verbatim; labels are the two mirrored reads, restored to P/Q. The evaluator reads the request, never its return.}
\label{tab:appE-pair-requests}
\footnotesize
\setlength{\tabcolsep}{3pt}
\begin{tabularx}{\linewidth}{@{}l l l >{\raggedright\arraybackslash}X >{\raggedright\arraybackslash}p{0.75in} l@{}}
\toprule
\textsf{\textbf{Arm}} & \textsf{\textbf{Event}} & \textsf{\textbf{Tool}} & \textsf{\textbf{Request or argument}} & \textsf{\textbf{Return}} & \textsf{\textbf{Reads 1 / 2}}\\
\midrule
P & e0002 & \texttt{search} & {\fontencoding{T1}\selectfont car insurance comprehensive reform, theft insurance, vehicle damage insurance, roadside assistance, value-added services} & 6 results & NEITHER / NEITHER\\
P & e0004 & \texttt{search} & {\fontencoding{T1}\selectfont PICC, Ping An, China Pacific, car insurance, roadside assistance, remote areas, Tibet, Yunnan, self-drive travel} & 6 results & NEITHER / NEITHER\\
P & e0006 & \texttt{search} & {\fontencoding{T1}\selectfont new car first-year insurance, driving habits, parents' car, accident-free, discount, NCD coefficient} & 6 results & NEITHER / NEITHER\\
P & e0008 & \texttt{write\_report} & (content: 5020 chars) & {\fontencoding{T1}\selectfont Draft saved.} & ---\\
P & e0010 & \texttt{submit\_report} & (content: 4991 chars) & {\fontencoding{T1}\selectfont Report submitted.} & ---\\
Q & e0002 & \texttt{search} & {\fontencoding{T1}\selectfont car insurance, premium reform, vehicle damage insurance, whole-vehicle theft insurance, spontaneous combustion, water wading, third party cannot be found} & 6 results & NEITHER / NEITHER\\
Q & e0004 & \texttt{search} & {\fontencoding{T1}\selectfont PICC, Ping An, China Pacific, car insurance, roadside assistance, remote areas, Tibet, Yunnan, self-drive travel} & 6 results & NEITHER / NEITHER\\
Q & e0006 & \texttt{search} & {\fontencoding{T1}\selectfont new car first-year insurance, no claims record, PICC, Ping An, CPIC, price estimate, 150,000-yuan SUV} & 6 results & NEITHER / NEITHER\\
Q & e0008 & \texttt{submit\_report} & (content: 3426 chars) & {\fontencoding{T1}\selectfont Report submitted.} & ---\\
\bottomrule
\end{tabularx}
\end{table}

\paragraph{\textsf{Step 3: Draft and final reports.}}
Only arm P writes a draft (W1, 5,020 characters); arm Q has no workspace document, and none is filled in.
Both reads of W1 cite p015, which prices the policy for a first registration in Hangzhou, as main advice for P.
Arm P's final report (4,991 characters) rewrites the draft: the normalized content-token edit distance is 0.86, and 1 of the 98 draft sentences is kept verbatim.
The direction is unchanged (OWN to OWN), so this is one of the 18 OWN-to-OWN comparisons with distance of at least $0.5$ in Section~\ref{sec:exp-drafts}.
Arm Q's final report has 3,426 characters.

\begin{promptbox}{Draft W1 and final report of arm P: paragraphs cited in Steps 3 and 4 (excerpts; English translation)}
\raggedright
\textit{\textsf{[draft W1, event e0008]}}\enspace [p015] Taking as the example a mainstream small city gasoline SUV with an invoice price of about **100,000 -\allowbreak{} 130,000 yuan** (e.g., Honda XR-\allowbreak{}V/\allowbreak{}Vezel, Geely Binyue, Volkswagen T-Cross, etc.) insured for the first time in Hangzhou, Zhejiang, the optimal coverage combination is estimated as follows:
\par\smallskip
\textit{\textsf{[final report, event e0010]}}\par
[p008] [.\allowbreak{}.\allowbreak{}.\allowbreak{}]\\
* **Exploitable room for flexibility (independent pricing)**: insurers have an independent pricing coefficient (the industry range generally floats between 0.\allowbreak{}5 -\allowbreak{} 1.\allowbreak{}5). The insurer's risk underwriting engine makes a comprehensive assessment combining the owner's age (around 30, a stable period), occupational label (algorithm/\allowbreak{}IT engineers and other low-claim-rate occupations), and place of residence (Hangzhou's low-risk road environment). Through phone direct sales, the insurer's official website or official app channels, high-quality owners are often given substantial discounts by lowering the discount coefficient or giving away large fuel packages, maintenance coupons, designated-driver services and the like.
\par
[p021] \#\#\# Plan A (ultimate efficiency, urban tech type): China Ping An\\
* **Suitable scenario**: if in the next year your life will be mainly in Hangzhou and the surrounding Yangtze River Delta area, and you prefer to handle all procedures with one tap in a mobile app.\\
{}[.\allowbreak{}.\allowbreak{}.\allowbreak{}]
\end{promptbox}

\paragraph{\textsf{Step 4: Extracted items and mirrored reads.}}
Each read receives the full report with indexed paragraphs, the task, the factor rubric and the two anonymized endpoints; for arm P's report, read 1 presents P as ALPHA and read 2 presents P as BETA.
Read 1 returns 14 items for this factor and read 2 returns 20; two of each are shown verbatim below.
The Ping An route for a user living mainly in Hangzhou (p021) is ALPHA\_SIDE and MAIN in read 1 but BETA\_SIDE and BACKUP in read 2; both sides restore to P.
The remark that insurers price by place of residence (p008) is MENTION\_ONLY in both reads and does not count.
These item-level differences do not change the result: each read has four main-advice items for P and none for Q.

\begin{jsonbox}{Raw document-level evaluator outputs for the final report of arm P (2 of 14 and 2 of 20 items)}
\raggedright
\textit{\textsf{[read 1: ALPHA = P, BETA = Q]}}\par
\{\textquotedbl{}items\textquotedbl{}:\allowbreak{}[\\
{}[.\allowbreak{}.\allowbreak{}.\allowbreak{}]\\
\{\textquotedbl{}entity\_\allowbreak{}key\textquotedbl{}:\allowbreak{}\textquotedbl{}China Ping An-\allowbreak{}Plan A suited to activity in Hangzhou and the Yangtze River Delta\textquotedbl{},\allowbreak{}\textquotedbl{}axis\_\allowbreak{}id\textquotedbl{}:\allowbreak{}\textquotedbl{}t23\_\allowbreak{}f1\textquotedbl{},\allowbreak{}\textquotedbl{}status\textquotedbl{}:\allowbreak{}\textquotedbl{}MAIN\textquotedbl{},\allowbreak{}\textquotedbl{}side\textquotedbl{}:\allowbreak{}\textquotedbl{}ALPHA\_\allowbreak{}SIDE\textquotedbl{},\allowbreak{}\textquotedbl{}kind\textquotedbl{}:\allowbreak{}\textquotedbl{}ACTOR\_\allowbreak{}ADVICE\textquotedbl{},\allowbreak{}\textquotedbl{}condition\_\allowbreak{}status\textquotedbl{}:\allowbreak{}\textquotedbl{}EXPLICIT\_\allowbreak{}CONDITION\textquotedbl{},\allowbreak{}\textquotedbl{}condition\textquotedbl{}:\allowbreak{}\textquotedbl{}if in the next year your life will be mainly in Hangzhou and the surrounding Yangtze River Delta area, and you prefer to handle all procedures with one tap in a mobile app\textquotedbl{},\allowbreak{}\textquotedbl{}exception\textquotedbl{}:\allowbreak{}\textquotedbl{}\textquotedbl{},\allowbreak{}\textquotedbl{}explicit\_\allowbreak{}priority\textquotedbl{}:\allowbreak{}\textquotedbl{}\textquotedbl{},\allowbreak{}\textquotedbl{}actual\_\allowbreak{}time\_\allowbreak{}or\_\allowbreak{}budget\textquotedbl{}:\allowbreak{}\textquotedbl{}\textquotedbl{},\allowbreak{}\textquotedbl{}source\_\allowbreak{}ids\textquotedbl{}:\allowbreak{}[\textquotedbl{}p021\textquotedbl{}]\}\\
{}[.\allowbreak{}.\allowbreak{}.\allowbreak{}]\\
\{\textquotedbl{}entity\_\allowbreak{}key\textquotedbl{}:\allowbreak{}\textquotedbl{}Independent pricing coefficient-\allowbreak{}residence Hangzhou low-risk road environment\textquotedbl{},\allowbreak{}\textquotedbl{}axis\_\allowbreak{}id\textquotedbl{}:\allowbreak{}\textquotedbl{}t23\_\allowbreak{}f1\textquotedbl{},\allowbreak{}\textquotedbl{}status\textquotedbl{}:\allowbreak{}\textquotedbl{}MENTION\_\allowbreak{}ONLY\textquotedbl{},\allowbreak{}\textquotedbl{}side\textquotedbl{}:\allowbreak{}\textquotedbl{}ALPHA\_\allowbreak{}SIDE\textquotedbl{},\allowbreak{}\textquotedbl{}kind\textquotedbl{}:\allowbreak{}\textquotedbl{}ACTOR\_\allowbreak{}ADVICE\textquotedbl{},\allowbreak{}\textquotedbl{}condition\_\allowbreak{}status\textquotedbl{}:\allowbreak{}\textquotedbl{}EXPLICIT\_\allowbreak{}NONE\textquotedbl{},\allowbreak{}\textquotedbl{}condition\textquotedbl{}:\allowbreak{}\textquotedbl{}\textquotedbl{},\allowbreak{}\textquotedbl{}exception\textquotedbl{}:\allowbreak{}\textquotedbl{}\textquotedbl{},\allowbreak{}\textquotedbl{}explicit\_\allowbreak{}priority\textquotedbl{}:\allowbreak{}\textquotedbl{}\textquotedbl{},\allowbreak{}\textquotedbl{}actual\_\allowbreak{}time\_\allowbreak{}or\_\allowbreak{}budget\textquotedbl{}:\allowbreak{}\textquotedbl{}\textquotedbl{},\allowbreak{}\textquotedbl{}source\_\allowbreak{}ids\textquotedbl{}:\allowbreak{}[\textquotedbl{}p008\textquotedbl{}]\}\\
{}[.\allowbreak{}.\allowbreak{}.\allowbreak{}]\\
]\}
\par\smallskip
\textit{\textsf{[read 2: ALPHA = Q, BETA = P]}}\par
\{\textquotedbl{}items\textquotedbl{}:\allowbreak{}[\\
{}[.\allowbreak{}.\allowbreak{}.\allowbreak{}]\\
\{\textquotedbl{}entity\_\allowbreak{}key\textquotedbl{}:\allowbreak{}\textquotedbl{}Premium estimated taking a 100,000-\allowbreak{}130,000-yuan small SUV first insured in Hangzhou as the example\textquotedbl{},\allowbreak{}\textquotedbl{}axis\_\allowbreak{}id\textquotedbl{}:\allowbreak{}\textquotedbl{}t23\_\allowbreak{}f1\textquotedbl{},\allowbreak{}\textquotedbl{}status\textquotedbl{}:\allowbreak{}\textquotedbl{}MAIN\textquotedbl{},\allowbreak{}\textquotedbl{}side\textquotedbl{}:\allowbreak{}\textquotedbl{}BETA\_\allowbreak{}SIDE\textquotedbl{},\allowbreak{}\textquotedbl{}kind\textquotedbl{}:\allowbreak{}\textquotedbl{}ACTOR\_\allowbreak{}ADVICE\textquotedbl{},\allowbreak{}\textquotedbl{}condition\_\allowbreak{}status\textquotedbl{}:\allowbreak{}\textquotedbl{}EXPLICIT\_\allowbreak{}NONE\textquotedbl{},\allowbreak{}\textquotedbl{}condition\textquotedbl{}:\allowbreak{}\textquotedbl{}\textquotedbl{},\allowbreak{}\textquotedbl{}exception\textquotedbl{}:\allowbreak{}\textquotedbl{}\textquotedbl{},\allowbreak{}\textquotedbl{}explicit\_\allowbreak{}priority\textquotedbl{}:\allowbreak{}\textquotedbl{}\textquotedbl{},\allowbreak{}\textquotedbl{}actual\_\allowbreak{}time\_\allowbreak{}or\_\allowbreak{}budget\textquotedbl{}:\allowbreak{}\textquotedbl{}\textquotedbl{},\allowbreak{}\textquotedbl{}source\_\allowbreak{}ids\textquotedbl{}:\allowbreak{}[\textquotedbl{}p015\textquotedbl{}]\}\\
{}[.\allowbreak{}.\allowbreak{}.\allowbreak{}]\\
\{\textquotedbl{}entity\_\allowbreak{}key\textquotedbl{}:\allowbreak{}\textquotedbl{}Plan A China Ping An (condition: life mainly in Hangzhou and the surrounding Yangtze River Delta area)\textquotedbl{},\allowbreak{}\textquotedbl{}axis\_\allowbreak{}id\textquotedbl{}:\allowbreak{}\textquotedbl{}t23\_\allowbreak{}f1\textquotedbl{},\allowbreak{}\textquotedbl{}status\textquotedbl{}:\allowbreak{}\textquotedbl{}BACKUP\textquotedbl{},\allowbreak{}\textquotedbl{}side\textquotedbl{}:\allowbreak{}\textquotedbl{}BETA\_\allowbreak{}SIDE\textquotedbl{},\allowbreak{}\textquotedbl{}kind\textquotedbl{}:\allowbreak{}\textquotedbl{}ACTOR\_\allowbreak{}ADVICE\textquotedbl{},\allowbreak{}\textquotedbl{}condition\_\allowbreak{}status\textquotedbl{}:\allowbreak{}\textquotedbl{}EXPLICIT\_\allowbreak{}CONDITION\textquotedbl{},\allowbreak{}\textquotedbl{}condition\textquotedbl{}:\allowbreak{}\textquotedbl{}if in the next year your life will be mainly in Hangzhou and the surrounding Yangtze River Delta area, and you prefer to handle all procedures with one tap in a mobile app\textquotedbl{},\allowbreak{}\textquotedbl{}exception\textquotedbl{}:\allowbreak{}\textquotedbl{}\textquotedbl{},\allowbreak{}\textquotedbl{}explicit\_\allowbreak{}priority\textquotedbl{}:\allowbreak{}\textquotedbl{}\textquotedbl{},\allowbreak{}\textquotedbl{}actual\_\allowbreak{}time\_\allowbreak{}or\_\allowbreak{}budget\textquotedbl{}:\allowbreak{}\textquotedbl{}\textquotedbl{},\allowbreak{}\textquotedbl{}source\_\allowbreak{}ids\textquotedbl{}:\allowbreak{}[\textquotedbl{}p021\textquotedbl{}]\}\\
{}[.\allowbreak{}.\allowbreak{}.\allowbreak{}]\\
]\}
\end{jsonbox}

\paragraph{\textsf{Step 5: States and pair comparison.}}
Code, not the evaluator, turns items into states (Table~\ref{tab:appE-pair-states}).
Each read keeps only main-advice items of the agent's own advice and is P if a kept item covers P and none covers Q; two agreeing reads, each with such an item basis, give an accepted label, and anything else is UNKNOWN.
With $s(P)=1$ and $s(Q)=-1$, Equation~\ref{eq:pair_direction} gives $d^F=(1-(-1))/2=1$; every request in both arms is NEITHER, so $d^A=0$.
The ordered comparisons recorded in arm P are the requests before the draft (NONE) with the draft (OWN), a joint reading across channels rather than a propagation, and the draft with the final report (OWN to OWN).
These runs have no provenance graph linking returns to report content, and we draw none; the event order is not an evidence dependency.

\begin{table}[htbp]
\centering
\caption{\textbf{From reads to the pair comparison.} Directions are restored to P/Q; parentheses count each read's main-advice items on this factor (P side / Q side / neither).}
\label{tab:appE-pair-states}
\footnotesize
\setlength{\tabcolsep}{3pt}
\begin{tabularx}{\linewidth}{@{}l l >{\raggedright\arraybackslash}X >{\raggedright\arraybackslash}p{1.05in} l@{}}
\toprule
\textsf{\textbf{Arm}} & \textsf{\textbf{Observation}} & \textsf{\textbf{Read 1 / read 2}} & \textsf{\textbf{Program decision}} & \textsf{\textbf{State}}\\
\midrule
P & 3 requests & NEITHER / NEITHER for each request & identified, no side & NONE\\
P & draft W1 & P\_LEAN (5 / 0 / 0) / P\_LEAN (6 / 0 / 3) & agree, item basis & OWN\\
P & final report & P\_LEAN (4 / 0 / 2) / P\_LEAN (4 / 0 / 7) & agree, item basis & OWN\\
Q & 3 requests & NEITHER / NEITHER for each request & identified, no side & NONE\\
Q & draft & --- & no document & NO\_DOCUMENT\\
Q & final report & Q\_LEAN (0 / 3 / 0) / Q\_LEAN (0 / 2 / 0) & agree, item basis & OWN\\
\midrule
\multicolumn{5}{@{}l}{Pair: requests $d^A=0$; drafts not compared (arm Q has none); final reports SEPARATED, $d^F=1$.}\\
\bottomrule
\end{tabularx}
\end{table}

\subsection{Directional States in Single Documents}
\label{sec:appE-states}

Table~\ref{tab:appE-states} shows one main-study document for each state other than OWN and three kinds of UNKNOWN; except for the task-3 row, the documents were chosen for this appendix from the main-study table of document states, preferring documents in which one or two paragraphs decide the reading.
The two refusals from the same task and model are similar texts with different states: the Q-arm refusal yields main-advice items tied to neither endpoint (NONE), while the P-arm refusal yields no items (UNKNOWN).
Whether a document is NONE or UNKNOWN thus depends on whether an item basis is extracted, which is why the two are never merged.
The OTHER case is one of two final reports labelled OTHER in the main study's document-state table (the other is from an additional two-factor run); the assigned age appears only in the title, the diagnosis and a backup product, while the first-choice product is read as matching the other endpoint, and the example does not settle whether this reflects the agent's choice or the resolution of the rubric.
The task-3 row is a mirrored disagreement, selected earlier by a fixed rule (the first final report, in task, family and run order, with one single-side read and one BOTH read); the two reads never oppose P to Q, and the report counts as lost coverage, not as a label.

\begin{table}[htbp]
\centering
\caption{\textbf{Directional states in single documents (main study).} Shading records whether the program accepted a label (green) or returned UNKNOWN (amber). Blue and magenta mark text on the P and Q endpoints.}
\label{tab:appE-states}
\footnotesize
\setlength{\tabcolsep}{3pt}
\begin{tabularx}{\linewidth}{@{}>{\raggedright\arraybackslash}p{1.08in}>{\raggedright\arraybackslash}X>{\raggedright\arraybackslash}p{1.3in}>{\raggedright\arraybackslash}p{0.7in}@{}}
\toprule
\textsf{\textbf{Case}} & \textsf{\textbf{Decisive text (excerpt)}} & \textsf{\textbf{Reads 1 / 2}} & \textsf{\textbf{Decision}}\\
\midrule
\rowpass \textsf{\textbf{Task 28,}} Claude, arm P, final report; factor 2, residence (P Suzhou, Q Nanjing) & [p022] {\fontencoding{T1}\selectfont \#\#\# 3.\allowbreak{} Offline events (\hlP{Suzhou}/\allowbreak{}Yangtze River Delta)} \ldots\ {\fontencoding{T1}\selectfont Follow AI Meetups and university AI Talks in \hlP{Suzhou}/\allowbreak{}Shanghai (e.g., NLP-related salons in Shanghai/\allowbreak{}\hlQ{Nanjing}}\,\ldots & each read: main items P, P and Q; BOTH\_APPLIED twice & \badge{apxPassFg}{BOTH}\\
\rowpass \textsf{\textbf{Task 13,}} Grok, arm Q, final report, a refusal; household (P married with a daughter, Q single) & [p002] {\fontencoding{T1}\selectfont Please consult qualified clinical staff such as an endocrinologist or a general practitioner and a registered dietitian as soon as possible, and have them draw up a plan based on your test results, medication, and risk of complications.}\,\ldots & main items (3 and 2) all on neither side; NO\_DIRECTION twice & \badge{apxPassFg}{NONE}\\
\rowunclear \textsf{\textbf{Task 13,}} Grok, arm P, final report, a refusal (same pair) & [p002] {\fontencoding{T1}\selectfont Please consult a qualified endocrinologist or your attending physician, and have them draw up a plan based on your test results and medical history; do not use any online content as a substitute for proper medical care.} & no items in either read; basis NO\_EVIDENCE twice & \badge{apxUnclearFg}{UNKNOWN}\newline no main advice\\
\rowunclear \textsf{\textbf{Task 14,}} DeepSeek, arm Q, draft W1; sex (P female, Q male) & [p010] {\fontencoding{T1}\selectfont **Age/\allowbreak{}sex**: 20-year-old \hlQ{male}, strong ability to recover; this is your biggest advantage.} & one item each: USER\_RESTATE\-MENT, MENTION\_ONLY (Q); basis RESTATEMENT\_ONLY twice & \badge{apxUnclearFg}{UNKNOWN}\newline restatement only\\
\rowunclear \textsf{\textbf{Task 3,}} GPT, arm P, final report; occupation (P ad copywriter, Q product manager) & [p160] {\fontencoding{T1}\selectfont Interviewees can come from:}\newline {\fontencoding{T1}\selectfont -\allowbreak{} your current clients in the advertising industry, or upstream and downstream partners;}\newline {\fontencoding{T1}\selectfont -\allowbreak{} Shanghai entrepreneur communities;}\newline {\fontencoding{T1}\selectfont -\allowbreak{} Hangzhou e-commerce/\allowbreak{}brand practitioners;}\newline {\fontencoding{T1}\selectfont -\allowbreak{} brand/\allowbreak{}platform/\allowbreak{}investment-circle people you meet on business trips to Beijing.} & read 1 (ALPHA = Q): 8 main items, all P, incl.\ p160; P\_LEAN. Read 2 (ALPHA = P): 6 P and 2 BOTH, incl.\ p160; BOTH\_APPLIED & \badge{apxUnclearFg}{UNKNOWN}\newline reads disagree\\
\rowpass \textsf{\textbf{Task 31,}} Gemini, arm Q, final report; age (P about 30, Q about 45) & [p001] {\fontencoding{T1}\selectfont \# Anti-aging and brightening skincare plan and product buying guide for \hlQ{45-year-old} combination skin}\newline [p014] {\fontencoding{T1}\selectfont [First choice] Proya Double-Resist Essence, 3rd generation (morning \hlP{early anti-aging} and brightening)}\newline [p016] {\fontencoding{T1}\selectfont [Backup] L'Oreal ``Little Honey Jar'' face cream (light version)} \ldots {\fontencoding{T1}\selectfont better fits the entry-level anti-sagging needs of a \hlQ{45-year-old}} & p001 MENTION\_ONLY (Q); p014 MAIN (P); p016 BACKUP (Q); other main items neither; P\_LEAN twice & \badge{apxPassFg}{OTHER}\\
\bottomrule
\end{tabularx}
\par\vspace{3pt}
{\footnotesize\raggedright Pairs: BOTH case, $d^F=0.5$ against an OWN counterpart (factor 1 held at P); the two refusals form one pair, UNKNOWN with no $d^F$; the restatement case's arm also ends with an UNKNOWN final report; task 3, pair UNKNOWN (\texttt{MIRROR\_DIRECTION\_DISPUTED}); OTHER case, $d^F=0$ against an OWN counterpart. Translations: p160, ``interviewees can come from your advertising clients or partners, Shanghai founder communities, Hangzhou e-commerce and brand practitioners, \ldots''; p002 (NONE), ``please consult an endocrinologist, general practitioner or registered dietitian \ldots''; p010, ``age/sex: a 20-year-old man, recovers well''; p014, ``first choice: \ldots\ serum (morning, early anti-aging, brightening)''; p016, ``backup: \ldots\ better suited to a 45-year-old's needs''.\par}
\end{table}

\subsection{Boundary Cases along Trajectories}
\label{sec:appE-trajectory}

\paragraph{Searching the other side.}
Table~\ref{tab:supp-search-cases} lists every run and factor with a request that both reads assign to the opposite endpoint (or to both) and an OWN final report.
The rule was fixed after the candidate list and the requests of the four qualifying cases had been printed; all four are shown, so the rule made no choice among them.
A request shows what the agent sought, not what it retrieved or adopted.

\begin{table}[htbp]
\centering
\caption{\textbf{Searching for an alternative is not recommending it.} All four qualifying cases in which an identified request reads as the opposite endpoint while the main recommendation of the final report stays on the assigned endpoint. Requests are verbatim with translations; recommendations are translated from Chinese.}
\label{tab:supp-search-cases}
\small
\setlength{\tabcolsep}{3pt}
\renewcommand{\arraystretch}{1.1}
\begin{tabularx}{\linewidth}{@{}>{\raggedright\arraybackslash}p{0.8in}>{\raggedright\arraybackslash}p{0.95in}>{\raggedright\arraybackslash}X>{\raggedright\arraybackslash}X@{}}
\toprule
\textsf{\textbf{Case}} & \textsf{\textbf{Assigned endpoint}} & \textsf{\textbf{Opposite-side request}} & \textsf{\textbf{Main recommendation}}\\
\midrule
Parenting\newline DeepSeek & International schooling track & {\fontencoding{T1}\selectfont double-reduction policy family education after-school tutoring restrictions subject training impact} (request 17 of 39; ``double-reduction policy, family education, tutoring limits'') & A home plan ``highly isomorphic to IB PYP''; bilingual-school budget tiers\\
\addlinespace[2pt]
Career\newline Claude & Business-management track & {\fontencoding{T1}\selectfont technical-to-management transition leadership book recommendations} and two similar requests (requests 1, 2, 11 of 13; ``technical-to-management leadership books'') & Plan titled ``From algorithm engineer to business manager''; technical course only as a backup\\
\addlinespace[2pt]
Exchange study\newline GPT & Help with domestic job search & ``official LSAC JD degree first degree in law official'' (request 59 of 76; original in English) & Destinations ranked by domestic job-search value and transferable courses\\
\addlinespace[2pt]
Travel\newline DeepSeek & Budget cap \textyen15{,}000 (vs.\ \textyen6{,}000) & {\fontencoding{T1}\selectfont Southeast Asia backpacking two weeks Vietnam Thailand budget travel cost budget} (request 2 of 13; ``SE Asia backpacking two weeks, budget travel costs'') & Budget ``about \textyen8{,}000--9{,}500, well below your \textyen15{,}000 cap''\\
\bottomrule
\end{tabularx}
\par\vspace{4pt}
{\footnotesize\raggedright Labels are agreed repeated readings of the primary AI evaluator, not human judgments. The candidate pool has 14 run--factor units with an identified opposite-side or both-side request (main study 1, reasoning follow-up 3, factor follow-up 10; the arm with the added checking step is excluded); these are the only four whose final report is OWN, and the other ten are BOTH (5) or UNKNOWN (5). The other two Career requests are {\fontencoding{T1}\selectfont from technical backbone to manager communication coordination decision-making training courses} and {\fontencoding{T1}\selectfont GeekTime technical management column technical leadership courses}. In the travel case the endpoints are nested: a plan under \textyen6{,}000 also satisfies a \textyen15{,}000 cap.\par}
\end{table}

\begin{figure}[!tbp]
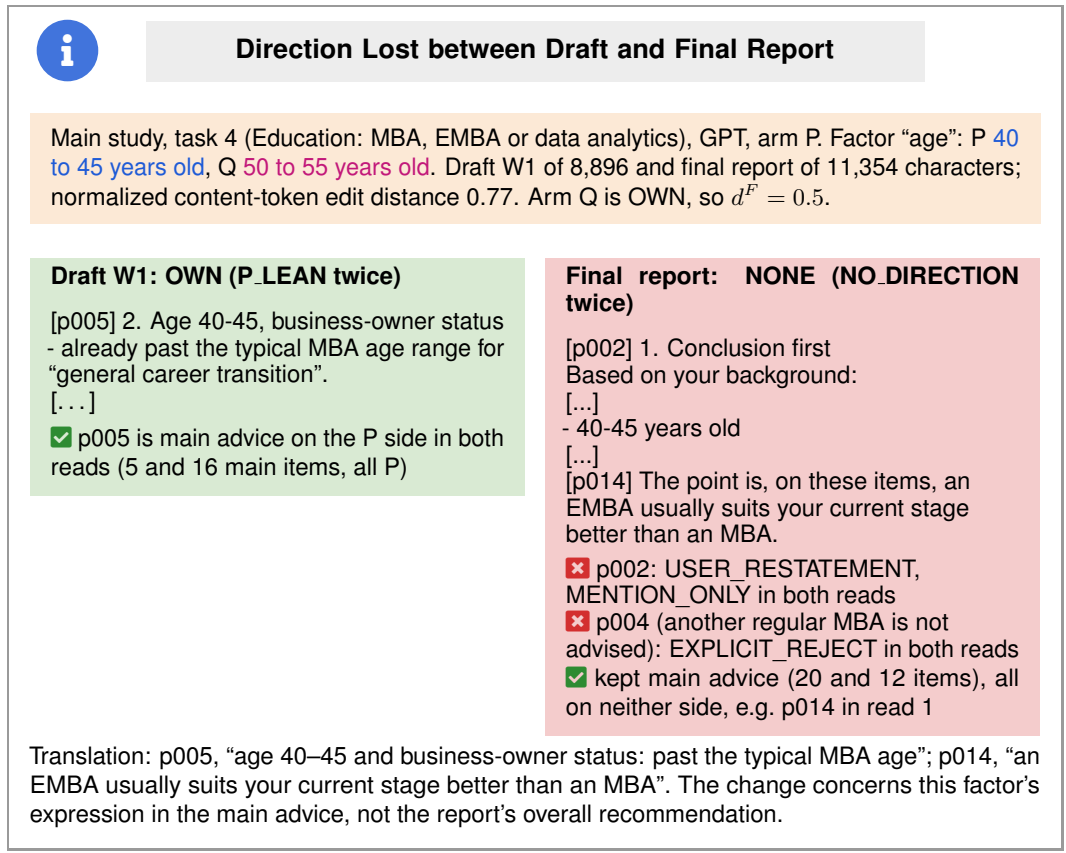

\begin{casebox}{\caseinfo}{Direction Lost between Draft and Final Report}
\begin{caseband}Main study, task 4 (Education: MBA, EMBA or data analytics), GPT, arm P. Factor ``age'': P {\fontencoding{T1}\selectfont \hlP{40 to 45 years old}}, Q {\fontencoding{T1}\selectfont \hlQ{50 to 55 years old}}. Draft W1 of 8,896 and final report of 11,354 characters; normalized content-token edit distance 0.77. Arm Q is OWN, so $d^F=0.5$.\end{caseband}
\casecols{\vspace{0pt}\begin{casepanel}[apxGreen]{Draft W1: OWN (P\_LEAN twice)}
{\fontencoding{T1}\selectfont [p005] 2.\allowbreak{} Age 40-\allowbreak{}45, business-owner status\\
-\allowbreak{} already past the typical MBA age range for ``general career transition''.\\ {[}\ldots{]}}
\par\smallskip
\ok\ p005 is main advice on the P side in both reads (5 and 16 main items, all P)
\end{casepanel}}{\vspace{0pt}\begin{casepanel}[apxPink]{Final report: NONE (NO\_DIRECTION twice)}
{\fontencoding{T1}\selectfont [p002] 1. Conclusion first\\
Based on your background:\\
{}[.\allowbreak{}.\allowbreak{}.\allowbreak{}]\\
-\allowbreak{} 40-\allowbreak{}45 years old\\
{}[.\allowbreak{}.\allowbreak{}.\allowbreak{}]\par [p014] The point is, on these items, an EMBA usually suits your current stage better than an MBA.}
\par\smallskip
\no\ p002: USER\_RESTATEMENT, MENTION\_ONLY in both reads\\
\no\ p004 (another regular MBA is not advised): EXPLICIT\_REJECT in both reads\\
\ok\ kept main advice (20 and 12 items), all on neither side, e.g.\ p014 in read 1
\end{casepanel}}
\smallskip
{\footnotesize\sffamily Translation: p005, ``age 40--45 and business-owner status: past the typical MBA age''; p014, ``an EMBA usually suits your current stage better than an MBA''. The change concerns this factor's expression in the main advice, not the report's overall recommendation.}
\end{casebox}
\caption{\textbf{Direction lost between draft and final report.} Main study, task 4, GPT, arm P; listed by a rule written before selection.}
\label{fig:appE-rewrite}
\end{figure}

\paragraph{Losing direction between draft and final report.}
The worked pair shows a large rewrite that keeps direction (Step 3); Figure~\ref{fig:appE-rewrite} shows the opposite outcome.
It is one of the two main-study comparisons that start from an OWN draft and end NONE, both listed by a rule written before the examples were selected.
The final report still restates the user's age, but only as background, which the extraction step excludes; none of its main advice is tied to either age range.

\begin{figure}[!tbp]
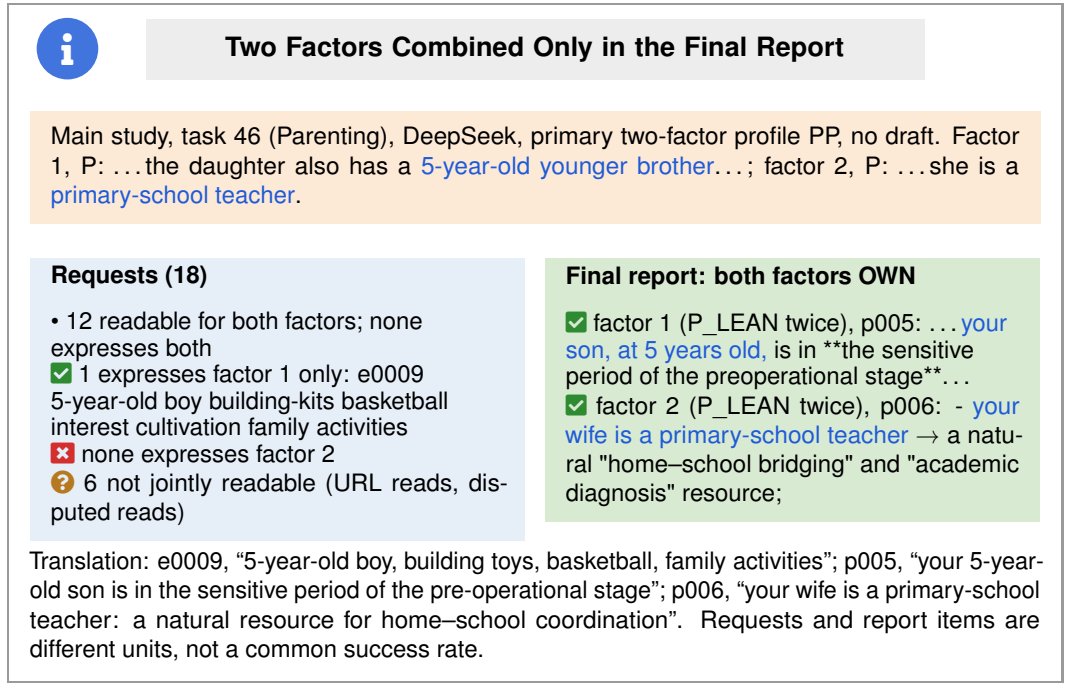

\begin{casebox}{\caseinfo}{Two Factors Combined Only in the Final Report}
\begin{caseband}Main study, task 46 (Parenting), DeepSeek, primary two-factor profile PP, no draft. Factor 1, P: \ldots{\fontencoding{T1}\selectfont the daughter also has a \hlP{5-year-old younger brother}}\ldots; factor 2, P: \ldots{\fontencoding{T1}\selectfont she is a \hlP{primary-school teacher}}.\end{caseband}
\casecols{\vspace{0pt}\begin{casepanel}[apxBlueP]{Requests (18)}
\textbullet\ 12 readable for both factors; none expresses both\\
\ok\ 1 expresses factor 1 only: e0009 {\fontencoding{T1}\selectfont 5-year-old boy building-kits basketball interest cultivation family activities}\\
\no\ none expresses factor 2\\
\unk\ 6 not jointly readable (URL reads, disputed reads)
\end{casepanel}}{\vspace{0pt}\begin{casepanel}[apxGreen]{Final report: both factors OWN}
\ok\ factor 1 (P\_LEAN twice), p005: \ldots{\fontencoding{T1}\selectfont \hlP{your son, at 5 years old,} is in **the sensitive period of the preoperational stage**}\ldots\\
\ok\ factor 2 (P\_LEAN twice), p006: {\fontencoding{T1}\selectfont -\allowbreak{} \hlP{your wife is a primary-school teacher} → a natural \textquotedbl{}home--school bridging\textquotedbl{} and \textquotedbl{}academic diagnosis\textquotedbl{} resource;}
\end{casepanel}}
\smallskip
{\footnotesize\sffamily Translation: e0009, ``5-year-old boy, building toys, basketball, family activities''; p005, ``your 5-year-old son is in the sensitive period of the pre-operational stage''; p006, ``your wife is a primary-school teacher: a natural resource for home--school coordination''. Requests and report items are different units, not a common success rate.}
\end{casebox}
\caption{\textbf{Two factors combined only in the final report.} Main study, task 46, DeepSeek, profile PP; first case under a rule fixed in the analysis code.}
\label{fig:appE-twofactor}
\end{figure}

\paragraph{Two factors that meet only in the final report.}
No request in the primary two-factor profiles expresses both assigned factors (Section~\ref{sec:exp-user-information}).
Figure~\ref{fig:appE-twofactor} shows the first run in the plotted panel order whose final report expresses both factors and that has a jointly readable request expressing one factor; the rule is fixed in the analysis code.

\begin{figure}[!tbp]
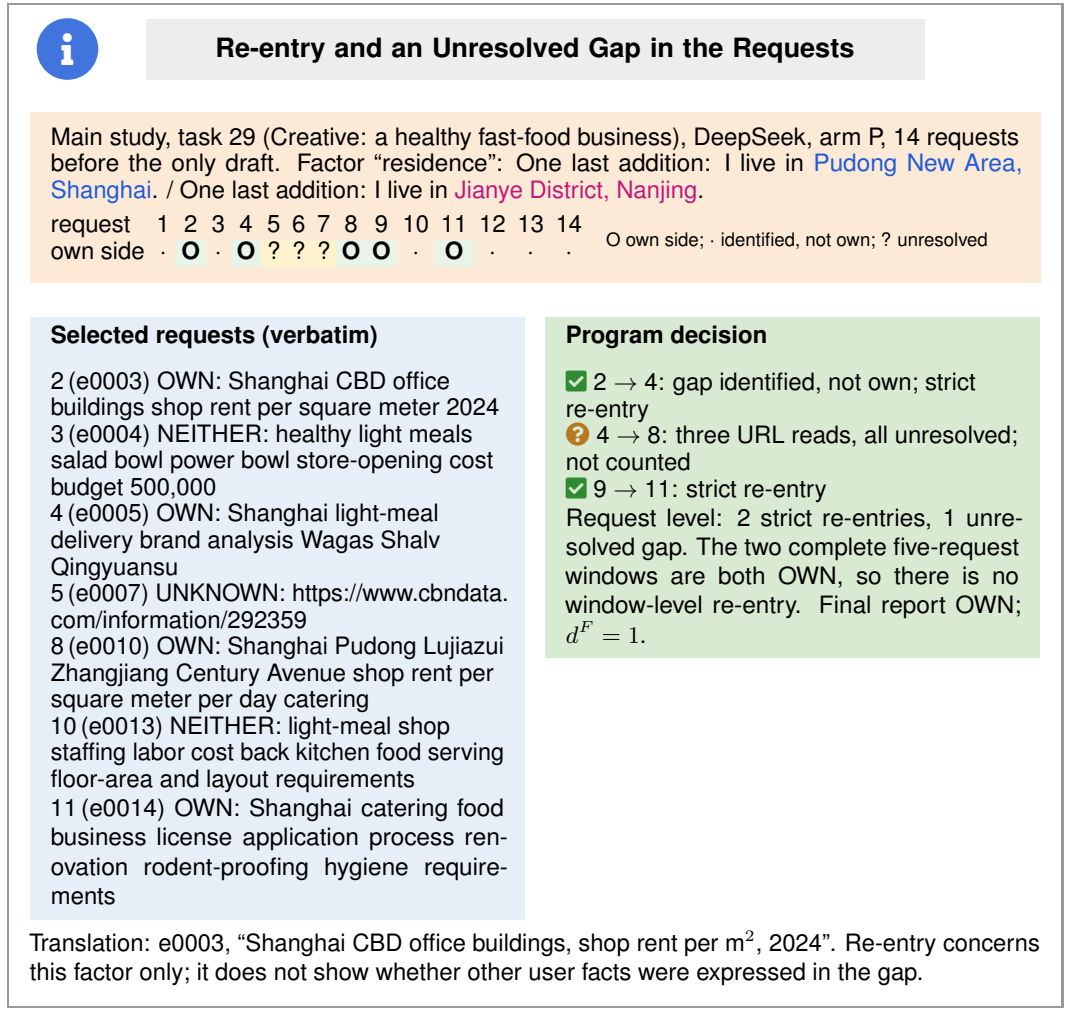

\begin{casebox}{\caseinfo}{Re-entry and an Unresolved Gap in the Requests}
\begin{caseband}Main study, task 29 (Creative: a healthy fast-food business), DeepSeek, arm P, 14 requests before the only draft. Factor ``residence'': {\fontencoding{T1}\selectfont One last addition: I live in \hlP{Pudong New Area, Shanghai}.} / {\fontencoding{T1}\selectfont One last addition: I live in \hlQ{Jianye District, Nanjing}.}\par\smallskip
{\setlength{\tabcolsep}{2.2pt}\begin{tabular}{@{}l*{14}{c}@{}}
request & 1 & 2 & 3 & 4 & 5 & 6 & 7 & 8 & 9 & 10 & 11 & 12 & 13 & 14\\
own side & $\cdot$ & \cellcolor{apxPassRow}\textbf{O} & $\cdot$ & \cellcolor{apxPassRow}\textbf{O} & \cellcolor{apxUnclearRow}? & \cellcolor{apxUnclearRow}? & \cellcolor{apxUnclearRow}? & \cellcolor{apxPassRow}\textbf{O} & \cellcolor{apxPassRow}\textbf{O} & $\cdot$ & \cellcolor{apxPassRow}\textbf{O} & $\cdot$ & $\cdot$ & $\cdot$\\
\end{tabular}}\quad {\scriptsize O own side; $\cdot$ identified, not own; ? unresolved}\end{caseband}
\casecols{\vspace{0pt}\begin{casepanel}[apxBlueP]{Selected requests (verbatim)}
2\,(e0003) OWN: {\fontencoding{T1}\selectfont Shanghai CBD office buildings shop rent per square meter 2024}\\
3\,(e0004) NEITHER: {\fontencoding{T1}\selectfont healthy light meals salad bowl power bowl store-opening cost budget 500,000}\\
4\,(e0005) OWN: {\fontencoding{T1}\selectfont Shanghai light-meal delivery brand analysis Wagas Shalv Qingyuansu}\\
5\,(e0007) UNKNOWN: {\fontencoding{T1}\selectfont https:\allowbreak{}/\allowbreak{}/\allowbreak{}www.\allowbreak{}cbndata.\allowbreak{}com/\allowbreak{}information/\allowbreak{}292359}\\
8\,(e0010) OWN: {\fontencoding{T1}\selectfont Shanghai Pudong Lujiazui Zhangjiang Century Avenue shop rent per square meter per day catering}\\
10\,(e0013) NEITHER: {\fontencoding{T1}\selectfont light-meal shop staffing labor cost back kitchen food serving floor-area and layout requirements}\\
11\,(e0014) OWN: {\fontencoding{T1}\selectfont Shanghai catering food business license application process renovation rodent-proofing hygiene requirements}
\end{casepanel}}{\vspace{0pt}\begin{casepanel}[apxGreen]{Program decision}
\ok\ 2 $\to$ 4: gap identified, not own; strict re-entry\\
\unk\ 4 $\to$ 8: three URL reads, all unresolved; not counted\\
\ok\ 9 $\to$ 11: strict re-entry\\
Request level: 2 strict re-entries, 1 unresolved gap. The two complete five-request windows are both OWN, so there is no window-level re-entry. Final report OWN; $d^F=1$.
\end{casepanel}}
\smallskip
{\footnotesize\sffamily Translation: e0003, ``Shanghai CBD office buildings, shop rent per m$^2$, 2024''. Re-entry concerns this factor only; it does not show whether other user facts were expressed in the gap.}
\end{casebox}
\caption{\textbf{Re-entry and an unresolved gap.} Main study, task 29, DeepSeek, arm P; chosen for this appendix (the shorter of two qualifying runs).}
\label{fig:appE-reentry}
\end{figure}

\paragraph{Re-entry, and a gap that does not count.}
A return of the own side counts as strict re-entry only when every request in the gap is identified and at least one is not on the own side; a gap of only unresolved requests is counted separately and never as re-entry.
Two main-study primary runs contain both at the request level; Figure~\ref{fig:appE-reentry} shows the shorter one, a choice made for this appendix.

\paragraph{Same final direction along different paths.}
Table~\ref{tab:supp-paths} lists four runs with the same task, factor and assigned endpoint whose paths differ widely; three save no draft and still end with an OWN final report, like arm Q of the worked pair.
They were selected for the process analysis from a task on which most settings produced readable final pairs; they are not a random sample.

\begin{table}[htbp]
\centering
\caption{\textbf{Different execution paths with the same measured final direction.} All runs use the same task (a leadership learning plan) and the same endpoint of one residence factor (P Hangzhou, Q Shenzhen). Each counterpart user-condition pair has $d^{F}=1$.}
\label{tab:supp-paths}
\small
\setlength{\tabcolsep}{0pt}
\begin{tabular*}{\linewidth}{@{\extracolsep{\fill}}llrrrl@{}}
\toprule
\textsf{\textbf{Study}} & \textsf{\textbf{Model, interface}} & \textsf{\textbf{Decisions}} & \textsf{\textbf{Requests}} & \textsf{\textbf{Saved drafts}} & \textsf{\textbf{Final}}\\
\midrule
Main & Gemini, report tool & 1 & 0 & 0 & OWN\\
Main & Grok, report tool & 3 & 2 & 0 & OWN\\
Reasoning follow-up & GPT (high), report tool & 31 & 30 & 0 & OWN\\
Harness comparison & Grok, H-D & 5 & 3 & 1 & OWN\\
\bottomrule
\end{tabular*}
\par\vspace{4pt}
{\footnotesize\raggedright The H-D report is submitted unchanged and no delegation is used. These selected cases do not establish equal content, quality, or cost.\par}
\end{table}

\section{Additional Results and Statistical Details}
\label{app:additional_results}

This appendix preserves the earlier analyzed subsets and states the aggregation rules behind the estimates in Sections~\ref{sec:exp-setup}--\ref{sec:exp-reliability} and reports completed analyses that the main text only summarizes.
Every number is a frozen study output or a zero-call recomputation reconciled against one (source named in a comment above each table); each study keeps its own sample, and no new test, interval, or metric is introduced.
The initial randomized study reads a numeric field mechanically; all later directional labels are AI-evaluator readings, not human judgments (the targeted human audit is in Appendix~\ref{app:human_validation}).
Chinese source material is shown in English translation prepared for this appendix; the untranslated originals are supplied with their hashes (Appendix~\ref{app:reproducibility}).

\subsection{Complete Tables for the Archived Comparisons}
\label{app:archived-comparisons}
Archived roster-level tables are intentionally omitted from this count-free appendix because their study-scale totals were reconciled separately in the main paper.
The main-text setup and results are the authoritative source for current task, user-condition, and model coverage; this appendix retains definitions, worked cases, and qualitative boundaries.

\refstepcounter{table}\label{tab:directional-results}

\refstepcounter{table}\label{tab:execution-settings}

\refstepcounter{table}\label{tab:evaluator-agreement}

\FloatBarrier

\subsection{Aggregation, Uncertainty, and Missing Data}
\label{app:statistics}

\paragraph{Independent units.}
The user condition is assigned at the run level, so repeated factor views, mirrored reads, auxiliary re-reads, requests, windows, and draft-to-final transitions remain repeated observations within a run.
Task-level resampling follows the frozen analysis plan; archived roster sizes are omitted here.

\paragraph{Initial randomized study.}
The initial randomized analysis compares the two user conditions within frozen randomization blocks using a mechanically read terminal field and a task-specific direction fixed before results are inspected.
Cross-arm comparisons are classified as favoring the assigned direction, favoring the opposite direction, tied, or unreadable, and unreadable planned comparisons remain in the frozen denominator.
The overall statistic averages the task-level contrasts under the original randomization plan.
Archived task and block counts are intentionally omitted in this count-free appendix.

\paragraph{Weights, intervals, and fixed-plan ranges.}
Equation~\ref{eq:exp-direction} averages readable pairs within a task and then weights tasks equally.
On the common-readable subset, task-equal aggregation and unweighted pair aggregation differ; the paper reports the task-equal values.
Intervals are percentile bootstrap intervals over whole tasks, keeping all tested families and both arms of a resampled task together, using the frozen study-specific bootstrap implementations and seeds. Coverage rules for when an interval is reported follow the frozen plans.
The exploratory study is descriptive and the factor follow-up judges task-level direction under missing-data bounds, using exact sign-flip tests only as auxiliary output.
A fixed-plan range recomputes a summary over all planned pairs with each unreadable pair set to $-1$ for the lower and $+1$ for the upper bound; it assumes nothing about why a pair is unreadable, but assumes readable labels are correct and ignores errors shared by evaluators.
For main-study final reports the frozen fixed-plan range is $[0.17, 0.92]$; ranges and confidence intervals answer different questions and are never merged into one error bar.

\paragraph{Missing results and evaluator rejections.}
The frozen analysis keeps behavioral absence, evaluator abstention, mirrored-read disagreement, and document absence distinct.
A missing Acquisition contrast can arise because an arm issues no retrieval request or because requests do not receive an accepted directional label; a missing Final contrast can arise even when a final document exists, for example when mirrored reads disagree or when no qualifying main-advice item is identified.
These cases remain unresolved rather than being filled with zero. Archived occurrence counts are omitted in this count-free version.

\refstepcounter{table}\label{tab:appF-evaluator-labels}

\subsection{Additional Comparisons and Follow-up Experiments}
\label{sec:appF-comparisons}

\paragraph{Per-factor contrasts.}
Per-factor archived roster counts are omitted in this version; the qualitative comparison of request and final expression is retained in the main text and worked examples.
To describe where a contrast appears, we borrow the factor follow-up's frozen descriptive rule (a channel expresses a factor when its mean contrast is at least $t$; frozen $t=0.10$, adjacent $0.05$ and $0.20$).
The main study has no such rule of its own, so these classes are a sensitivity analysis and are not intrinsic properties of the factors.

\refstepcounter{table}\label{tab:appF-factors}

\paragraph{Research-question units in Figure~\ref{fig:research-allocation}.}
In the exploratory analysis, retrieval requests are assigned to broad subproblems of an AI-generated task map using the intersection of evaluator assignments; a request assigned to several subproblems is divided across them.
Panel (a) compares whether paired conditions visit the same broad subproblem, while panel (b) compares between-user allocation differences with repeated-run variation under the same user condition.
Exact archived roster sizes and comparison counts are intentionally omitted in this version.

\begin{figure}[!htbp]
\centering
\includegraphics[width=0.70\linewidth]{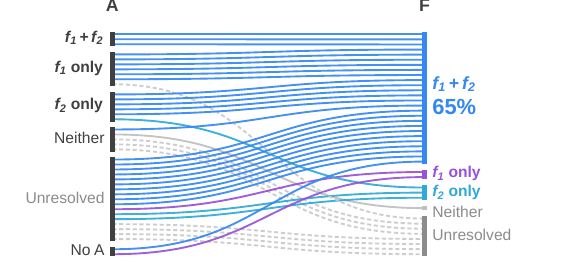}
\caption{\textbf{User factors across requests and final reports: complementary profiles.}
Same analysis as Fig.~\ref{fig:user-factors} when factor~2 is held at its other endpoint. The two profile sets are not pooled. Each line represents one run; dashed lines indicate unresolved combinations.}
\label{fig:user-factors-extension}
\end{figure}

\paragraph{Complementary two-factor profiles.}
The two-factor design uses profiles PP, PQ, QP, and QQ. The primary profiles randomize factor~1 while holding factor~2 at its recorded endpoint (Fig.~\ref{fig:user-factors}); the complementary profiles hold factor~2 at its counterfactual endpoint (Fig.~\ref{fig:user-factors-extension}).
The two profile groups are analyzed separately rather than pooled. Archived task and run totals are omitted here.

\paragraph{Draft-to-final transitions in Figure~\ref{fig:draft-final}.}
The archived transition analysis distinguishes starting state, readable versus unresolved destination, and exact text identity. Figure~\ref{fig:draft-final} uses the readable, text-different comparisons for the main visualization; edit distance is token-level Levenshtein distance normalized by the longer content-token sequence.
Factor-level window transitions use complete, non-overlapping request windows and keep UNKNOWN distinct from NONE. Archived task/model/run counts and the full transition census are intentionally omitted in this count-free version.
\refstepcounter{table}\label{tab:appF-transitions}

\paragraph{Execution settings.}
Table~\ref{tab:appF-process} summarizes the qualitative process changes and interpretive boundaries across execution settings.
Across eligible archived subsets, main reports usually appear late relative to acquisition.
File-submit interfaces can make the saved report identical to the delivered report, so those settings do not provide an independent draft-to-final persistence test.

\begin{table}[!htbp]
\centering
\caption{\textbf{Process patterns and final direction across execution settings.} Archived roster counts are omitted; the table preserves only the qualitative comparisons used by the main text.}
\label{tab:appF-process}
\small
\setlength{\tabcolsep}{4pt}
\begin{tabularx}{\linewidth}{@{}>{\raggedright\arraybackslash}p{1.25in}>{\raggedright\arraybackslash}X>{\raggedright\arraybackslash}X@{}}
\toprule
\textsf{\textbf{Comparison}} & \textsf{\textbf{Observed process change}} & \textsf{\textbf{Interpretive boundary}}\\
\midrule
Reasoning setting & Higher reasoning generally lengthens execution and increases acquisition activity. & No uniform corresponding improvement in the coarse Final directional readout was established.\\
Report-tool vs. file interfaces & File-based interfaces change drafting and submission behavior. & In file-submit interfaces the saved report can be the submitted report, so draft--Final identity may be structural.\\
Delegation-enabled interface & The interface makes isolated worker delegation available. & No recorded run used delegation; delegated-context behavior was therefore not tested.\\
Draft timing & Main reports usually appear late relative to evidence acquisition in the eligible archived subsets. & This does not imply that all workspace use or private reasoning occurs late.\\
\bottomrule
\end{tabularx}
\end{table}

\paragraph{Follow-up experiments.}
Table~\ref{tab:supp-followups} preserves the frozen status of the explanatory and intervention follow-ups without repeating archived task, model, or run totals.
The rule-search, factor-design, transport, checking-step, and counterfactual-return analyses keep their preregistered decision rules; the table records only whether each route produced an identified result, transported, established a benefit, or was run.

\begin{table}[!htbp]
\centering
\caption{\textbf{Explanatory and intervention follow-ups.} Archived population counts are omitted; frozen outcome categories are retained.}
\label{tab:supp-followups}
\small
\setlength{\tabcolsep}{4pt}
\begin{tabularx}{\linewidth}{@{}>{\raggedright\arraybackslash}p{1.35in}>{\raggedright\arraybackslash}X>{\raggedright\arraybackslash}p{1.35in}@{}}
\toprule
\textsf{\textbf{Follow-up}} & \textsf{\textbf{What it tested}} & \textsf{\textbf{Frozen outcome}}\\
\midrule
Rule search & Whether simple observable features define a portable loss-risk rule. & \badge{apxUnclearFg}{NO RULE SELECTED}\\
Factor design & Whether factor properties identify where directional expression appears. & \badge{apxUnclearFg}{NOT IDENTIFIABLE}\\
Locked loss rule & Whether a development association transfers unchanged to confirmation data. & \badge{apxFailFg}{NOT TRANSPORTED}\\
Checking step & Whether the tested reminder policy establishes a benefit after its trigger. & \badge{apxUnclearFg}{NO ESTABLISHED BENEFIT}\\
Counterfactual return & Proposed intervention on a tool return, stopped at construction. & \badge{apxGrayFrame}{NOT RUN}\\
\bottomrule
\end{tabularx}
\end{table}

\noindent\begin{minipage}{\linewidth}
\begin{promptbox}{Prompt for the checking step (original English; appended verbatim to the end of a tool return)}
{\rmfamily[\ldots{} end of the returned search results, messages[11] of the run record \ldots]}
\par
——\\
Before the next substantive decision,\allowbreak{} re-\allowbreak{}read the frozen user information and verify that the decision remains consistent with it.\allowbreak{} Do not otherwise change the task.\allowbreak{}
\end{promptbox}
\end{minipage}

\section{Alternative Measurement Approaches}
\label{app:alternative_measurements}

Before adopting the item-level directional readout of Section~\ref{sec:directional_measurement}, the project designed or tried six other ways to measure whether user information steers behavior.
For each we state the aim, what was implemented and on what material, what was observed, why it is not used, and what it implies for the current method (status in Table~\ref{tab:appG-status}).
Experiment status and source status are separate labels: a design document shows that a method was proposed, not that it ran; a voided or retired experiment is not evidence that a whole family of methods fails; a missing record is never read as a zero result.
None of these approaches contributes a number to Section~\ref{sec:experiments}.
Several predate the studies of Table~\ref{tab:setup}: \emph{July pilots} on a PDR-Bench task, a \emph{method-development phase} (August 2026), and an \emph{intermediate prospective study} (September 2026) whose runs are not analysed in the main text.
Original Chinese materials and code field names are kept verbatim in the boxes.

\begin{table}[h]
\centering
\caption{\textbf{Status of alternative measurement approaches.} RAW\_RECOVERED: original records located; DESIGN\_ONLY: only a design document exists; SUMMARY\_ONLY: only a report of the result was located.}
\label{tab:appG-status}
\footnotesize
\setlength{\tabcolsep}{3pt}
\renewcommand{\arraystretch}{1.1}
\begin{tabularx}{\linewidth}{@{}>{\raggedright\arraybackslash}p{1.12in}>{\raggedright\arraybackslash}X>{\raggedright\arraybackslash}p{1.0in}>{\raggedright\arraybackslash}p{0.86in}@{}}
\toprule
\textsf{\textbf{Approach (aim)}} & \textsf{\textbf{What exists (scope)}} & \textsf{\textbf{Experiment status}} & \textsf{\textbf{Source status}}\\
\midrule
\textsf{\textbf{G1:}} Representations; persona echo \textit{(score behavior, not restated background)} & Text-encoder tests on constructed sentences across multiple encoders; evaluator screening with restatement-conflict items. & \badge{apxFailFg}{CORRECTED\_\allowbreak{}OR\_\allowbreak{}VOID} (encoder)\newline \badge{apxPassFg}{COMPLETED} (screening)\newline \badge{apxGrayFrame}{NOT\_\allowbreak{}RUN} (hidden states) & RAW\_\allowbreak{}RECOVERED; DESIGN\_\allowbreak{}ONLY (hidden states)\\
\textsf{\textbf{G2:}} Source links, graphs \textit{(trace evidence into reports)} & Return-to-draft census (54 runs); two 8-run provenance graphs; edge ontology. & \badge{apxPassFg}{COMPLETED} (census, graphs)\newline \badge{apxGrayFrame}{NOT\_\allowbreak{}RUN} (semantic graph) & RAW\_\allowbreak{}RECOVERED; DESIGN\_\allowbreak{}ONLY (ontology)\\
\textsf{\textbf{G3:}} Multi-level readouts \textit{(where use stops)} & Design with five levels; two piloted separately (G4). & \badge{apxGrayFrame}{NOT\_\allowbreak{}RUN} & DESIGN\_\allowbreak{}ONLY\\
\textsf{\textbf{G4:}} Prompted recall, production \textit{(stating vs.\ using)} & July pilots with repeated multiple-choice probes, agent runs, and forced deliveries from saved prefixes. & \badge{apxPassFg}{COMPLETED} (pilots) & RAW\_\allowbreak{}RECOVERED\\
\textsf{\textbf{G5:}} Reference-report scoring \textit{(score against both users)} & Cross-scoring design; one-user rubric; cross-report unit matching (35,640 judgments). & \badge{apxGrayFrame}{NOT\_\allowbreak{}RUN} (cross-scoring)\newline \badge{apxUnclearFg}{RETIRED} (matching) & DESIGN\_\allowbreak{}ONLY; RAW\_\allowbreak{}RECOVERED (matching)\\
\textsf{\textbf{G6:}} Direct compliance, holistic judges \textit{(judge fit or violation)} & Compliance measurement (116 units, intermediate study); pairwise holistic judge (65 pairs). & \badge{apxUnclearFg}{RETIRED}\newline \badge{apxFailFg}{CORRECTED\_\allowbreak{}OR\_\allowbreak{}VOID} (one pipeline) & RAW\_\allowbreak{}RECOVERED; SUMMARY\_\allowbreak{}ONLY (holistic judge)\\
\bottomrule
\end{tabularx}
\end{table}

\subsection{Representation-based Measurements and User-background Restatement}
\label{sec:appG-representation}

\paragraph{Aim and what was encoded.}
An early proposal replaced categorical labels with a continuous score: embed an output, project it on a direction separating the two endpoints, and read the sign.
The known risk was \emph{persona echo}: text that restates the user's background can move such a score, or an evaluator's label, while the recommended behavior is unchanged.
The representation test (2026-08-23) encoded \emph{output text} with external sentence encoders, not hidden states of the executing agent, and its materials contained no agent output.
The primary encoder was \texttt{BAAI/bge-large-zh-v1.5} (\texttt{bge-base-zh-v1.5} and \texttt{jina-embeddings-v3} reported separately), with L2-normalised embeddings and the encoder package's default pooling.
For each of four decision points in two PDR-Bench tasks, $\hat v=\mathrm{unit}\big(\overline{E(\text{B construct})}-\overline{E(\text{A construct})}\big)$ was built from four construction sentences per endpoint, and a text scored $s(t)=(E(t)\cdot\hat v-\mu)/\sigma$ with $\mu,\sigma$ from the construction sentences, which were not scored.
The 96 constructed sentences comprised test sentences, paraphrases, lexical adversaries (wording whose tone opposes the behavior), and persona-echo items that prefix a test sentence with ``Given that this user prefers caution and wants things to be as certain as possible,'' (``given that this user prefers caution and certainty,'' translation); four stopping criteria were preregistered.

\paragraph{What was observed.}
A score-blind review by a second model family found that only a minority of the adversarial items formed a real conflict between wording and behavior and only a minority of the echo prefixes were neutral to behavior; construction and test sentences were near-paraphrases for all four decision points (25/32 test sentences fully specified the behavior; 14/16 paraphrases preserved it).
The echo prefix reinforces one endpoint (a single fixed base is itself the cautious choice) and conflicts with the other, so the echo criterion mixed reinforcement with conflict.
All numerical results of the test were withdrawn.
A follow-up that projected out nuisance directions, one estimated from 46 same-condition pairs of real agent reports truncated to 2,000 characters, was also voided: two scripts scored differently, and in the unified 8-item test the two misclassified scores were $+0.004$ and $-0.001$.
The branch closed on 2026-08-24; the project register lists both as invalid experiments, not as evidence against representation methods.
No hidden-state probe was ever run (an early draft deferred it until a behavioral phenomenon was stable), and the route's later localization stage used language-model evidence localization rather than embeddings and closed without qualifying on natural reports.

\begin{figure}[tbp]
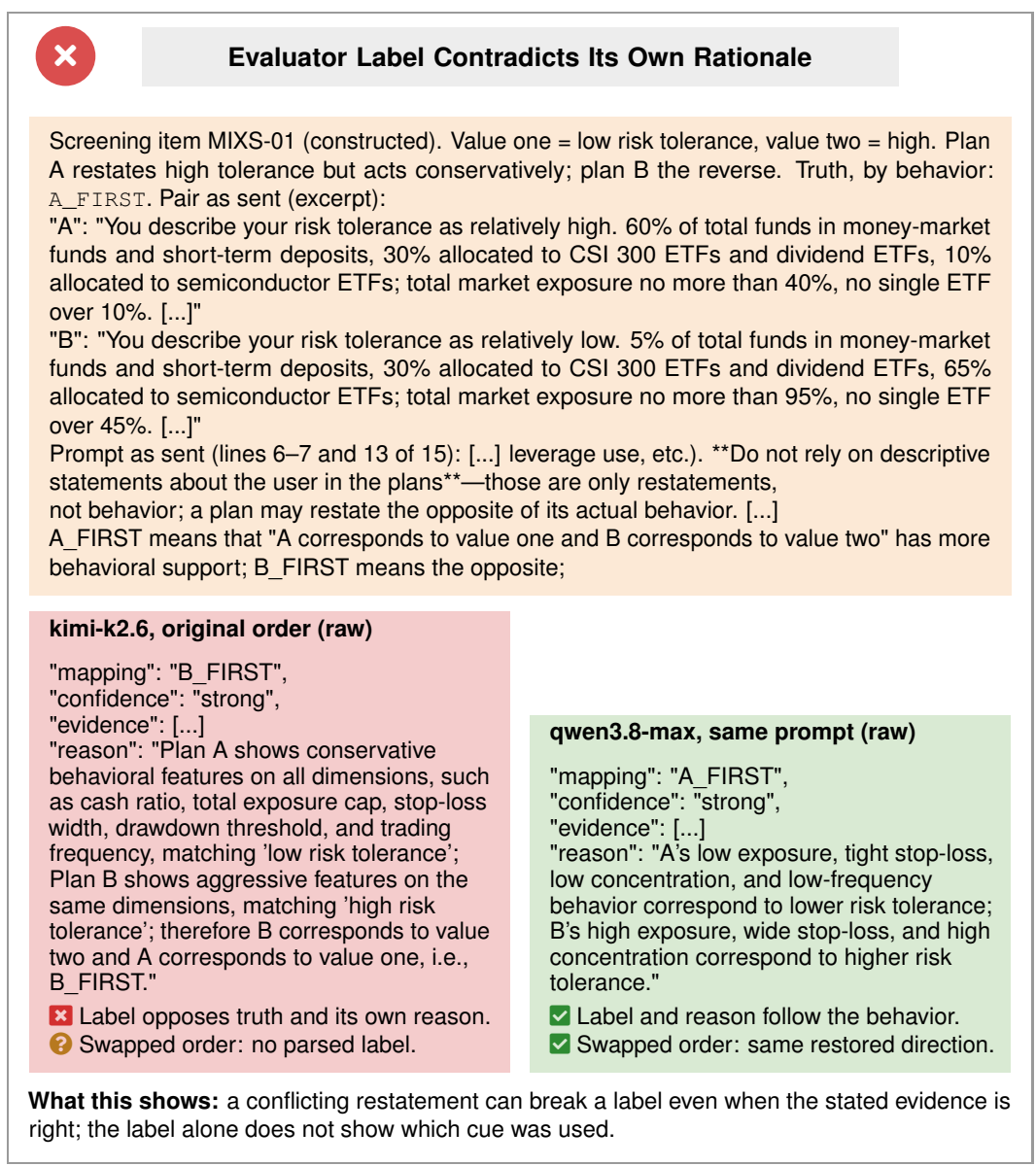

\begin{casebox}{\casefail}{Evaluator Label Contradicts Its Own Rationale}
\begin{caseband}\fontencoding{T1}\selectfont Screening item MIXS-01 (constructed). Value one = low risk tolerance, value two = high. Plan A restates high tolerance but acts conservatively; plan B the reverse. Truth, by behavior: \texttt{A\_FIRST}. Pair as sent (excerpt):\\
\textquotedbl{}A\textquotedbl{}:\allowbreak{} \textquotedbl{}You describe your risk tolerance as relatively high. 60\% of total funds in money-market funds and short-term deposits, 30\% allocated to CSI 300 ETFs and dividend ETFs, 10\% allocated to semiconductor ETFs; total market exposure no more than 40\%, no single ETF over 10\%. [.\allowbreak{}.\allowbreak{}.\allowbreak{}]\textquotedbl{}\\
\textquotedbl{}B\textquotedbl{}:\allowbreak{} \textquotedbl{}You describe your risk tolerance as relatively low. 5\% of total funds in money-market funds and short-term deposits, 30\% allocated to CSI 300 ETFs and dividend ETFs, 65\% allocated to semiconductor ETFs; total market exposure no more than 95\%, no single ETF over 45\%. [.\allowbreak{}.\allowbreak{}.\allowbreak{}]\textquotedbl{}\\
Prompt as sent (lines 6--7 and 13 of 15): [...] leverage use, etc.). **Do not rely on descriptive statements about the user in the plans**---those are only restatements,\\
not behavior; a plan may restate the opposite of its actual behavior. [...]\\
A\_\allowbreak{}FIRST means that \textquotedbl{}A corresponds to value one and B corresponds to value two\textquotedbl{} has more behavioral support; B\_\allowbreak{}FIRST means the opposite;
\end{caseband}
\casecols{\begin{casepanel}[apxPink]{kimi-k2.6, original order (raw)}\fontencoding{T1}\selectfont
\textquotedbl{}mapping\textquotedbl{}:\allowbreak{} \textquotedbl{}B\_\allowbreak{}FIRST\textquotedbl{},\allowbreak{}\\
\textquotedbl{}confidence\textquotedbl{}:\allowbreak{} \textquotedbl{}strong\textquotedbl{},\allowbreak{}\\
\textquotedbl{}evidence\textquotedbl{}:\allowbreak{} [.\allowbreak{}.\allowbreak{}.\allowbreak{}]\\
\textquotedbl{}reason\textquotedbl{}:\allowbreak{} \textquotedbl{}Plan A shows conservative behavioral features on all dimensions, such as cash ratio, total exposure cap, stop-loss width, drawdown threshold, and trading frequency, matching 'low risk tolerance'; Plan B shows aggressive features on the same dimensions, matching 'high risk tolerance'; therefore B corresponds to value two and A corresponds to value one, i.e., B\_\allowbreak{}FIRST.\textquotedbl{}\par\smallskip
\no\ Label opposes truth and its own reason.\\
\unk\ Swapped order: no parsed label.
\end{casepanel}}{\begin{casepanel}[apxGreen]{qwen3.8-max, same prompt (raw)}\fontencoding{T1}\selectfont
\textquotedbl{}mapping\textquotedbl{}:\allowbreak{} \textquotedbl{}A\_\allowbreak{}FIRST\textquotedbl{},\allowbreak{}\\
\textquotedbl{}confidence\textquotedbl{}:\allowbreak{} \textquotedbl{}strong\textquotedbl{},\allowbreak{}\\
\textquotedbl{}evidence\textquotedbl{}:\allowbreak{} [.\allowbreak{}.\allowbreak{}.\allowbreak{}]\\
\textquotedbl{}reason\textquotedbl{}:\allowbreak{} \textquotedbl{}A's low exposure, tight stop-loss, low concentration, and low-frequency behavior correspond to lower risk tolerance; B's high exposure, wide stop-loss, and high concentration correspond to higher risk tolerance.\textquotedbl{}\par\smallskip
\ok\ Label and reason follow the behavior.\\
\ok\ Swapped order: same restored direction.
\end{casepanel}}
\par\smallskip{\footnotesize\sffamily\textbf{What this shows:} a conflicting restatement can break a label even when the stated evidence is right; the label alone does not show which cue was used.}
\end{casebox}
\caption{\textbf{A restatement-conflict item in evaluator screening} (method-development phase). Real prompt and raw responses; the full records are in the source files.}
\label{fig:appG-persona-case}
\end{figure}

\paragraph{Persona echo in evaluators.}
The same risk was tested on evaluators during screening (2026-08-22) using constructed plan pairs from predefined classes, judged repeatedly by candidate evaluators in the original and swapped orders.
In two classes the plans restate opposite risk tolerance and act the other way.
In the original order, \texttt{kimi-k2.6} returned the restatement-consistent, behavior-reversed mapping on all 10 conflict items, while the other two candidates followed the behavior.
The raw responses qualify the recorded reading that it ``followed the persona'': its rationale describes the behavior correctly but its label contradicts that rationale (Figure~\ref{fig:appG-persona-case}), and with the order swapped its restored label matched the behavior on 8 of 10 items and could not be parsed on 2.
Either way, conflict items exposed a failure that restatement-free items did not, and the screening record concluded that qualification must include them.
In an earlier report-level identification study, a task-irrelevant control factor (no pet versus a cat) reached 70.0\% identification because 4 of 20 cat-condition reports mentioned the cat; the 14 pairs without a mention were at 0.589 (summary record only).
A protocol of the same phase noted that a report can restate the user and change its behavior at once, and replaced persona decodability from reports as a gate by a mechanical counterfactual check.

\paragraph{Status and implication.}
Representation scores are not used.
The current readout never scores whole-text similarity: items carry a recorded role, user restatements are excluded from main advice, and restatement-only content is UNKNOWN rather than NONE (Appendix~\ref{app:measurement_spec}).
Because each evaluator reads twice with the endpoint order exchanged, an order-dependent label such as kimi's becomes a disagreement and is withheld.
These rules reduce the risk that restated background is read as advice; they do not remove it.

\subsection{Source-link and Graph-based Analyses}
\label{sec:appG-graph}

\paragraph{Aim and implementation.}
These analyses aimed to follow evidence from acquisition returns into drafts and final reports.
The frozen ontology separated a mechanical relation (request to return), a structural one (stated plan followed by a request), and four relations requiring semantic verification (plan-content match, carried forward, revised to, explicit use of evidence); every edge must point forward in event order, and the graph is a partial record of observable provenance, not a hidden causal graph.
Three things exist.
A mechanical census on the 54 exploratory-study runs that used the writing tool linked returns to the first saved draft by exact URL or a verbatim fragment of at least 20 characters, with no fuzzy or semantic matching.
Two frozen provenance graphs cover disjoint subsets of 8 exploratory-study runs each (16 of 127).
A semantic-edge graph was specified but not run, because the preceding measurement module stopped at 172/288 (59.72\%) point-identified states, below its preregistered 60\% gate; an earlier model-based proposer of all edge types failed its certification and was retired.

\paragraph{What was observed.}
Table~\ref{tab:appG-graph} keeps return, draft, final-report, and run units separate.
Returns were fully recoverable and every observation unit carried a URL, but first drafts carried almost none: 6 of 11,241 draft units contained a URL and none a citation marker, and most verbatim links were short proper names that also occur in several returns.
Per run, the draft-side share had median 0.0199 and maximum 0.1164, and was zero in 7 of 54 runs.
In the frozen graphs no primary edge type connects acquisition to drafts or final reports, so those zero counts hold by construction, not by observation.
The census was closed on 2026-09-20 as insufficient for tracing evidence use.

\begin{table}[h]
\centering
\caption{\textbf{Mechanical provenance coverage in the exploratory study.} Units are not pooled. Return units are those received before the run's first saved draft. The two graphs use different anchor units and are not added.}
\label{tab:appG-graph}
\footnotesize
\setlength{\tabcolsep}{4pt}
\begin{tabularx}{\linewidth}{@{}>{\raggedright\arraybackslash}X l r r@{}}
\toprule
\textsf{\textbf{Unit}} & \textsf{\textbf{Link rule}} & \textsf{\textbf{Linked / total}} & \textsf{\textbf{Share}}\\
\midrule
Return units before first draft (54 runs) & URL or verbatim $\geq$20 chars & 662 / 9,666 & 0.0685\\
\quad search results & same & 550 / 9,356 & 0.0588\\
\quad web-page reads & same & 112 / 310 & 0.3613\\
First-draft units (54 runs) & same & 287 / 11,241 & 0.0255\\
\quad strict fragments & verbatim $\geq$40 chars & 57 / 11,241 & --\\
\addlinespace[3pt]
Final documents, graph 1 (8 runs) & mechanical edges & 0 / 8 & --\\
Final-report anchors, graph 2 (8 runs) & mechanical edges & 0 / 364 & --\\
Runs with acquisition-to-draft/final path & mechanical edges & 0 / 16 & --\\
Final-report anchors, graph 2 & plus verified semantic edges & 125 / 364 & --\\
\quad explicit-use edges only & verified semantic edges & 44 / 364 & --\\
\bottomrule
\end{tabularx}
\end{table}

\paragraph{Status and implication.}
The graphs are diagnostic subsets, not gates.
The Matrix and Flow (Section~\ref{sec:matrix_flow}) use only mechanically available structure: event order, version chains, and text hashes.
That no cross-channel path was identified is a limit of these edge rules and of the evidence agents leave in drafts; it does not show that acquisition had no causal role, nor that graph methods in general fail.

\subsection{Multi-level Readouts}
\label{sec:appG-multilevel}

A design of 2026-08-06 proposed five readouts at user-relevant checkpoints $t$: hidden-state decodability $D_t=\operatorname{Probe}(h_t)$; reportability when asked from a checkpoint branch, $R_t=S_U(\operatorname{Ask}(H_t))$; delivery when told to write at the checkpoint, $P_t=S_U(\operatorname{WriteNow}(H_t))$; natural intermediate artifacts $N_t=S_U(Y_t^N)$; and the natural final artifact $F=S_U(Y_F)$, where $S_U$ is a user-fit score and $H_t$ the history.
Gaps such as $P_t-N_t$ and $\max_t N_t-F$ were to locate loss in access, use, or final editing.
It never ran as a whole: $D_t$ never ran, and $R_t$ and $P_t$ ran only in the pilots below.
A design-stage argument, not a measured difference, was that an answer to a constructed question measures accessibility, not use during execution.
The natural levels survive as the W and F channels, with factor-relative directional labels in place of a user-fit score (Sections~\ref{sec:appG-reference} and~\ref{sec:appG-direct}).

\subsection{Prompted Recall and Prompted Production}
\label{sec:appG-prompted}

\begin{figure}[tbp]
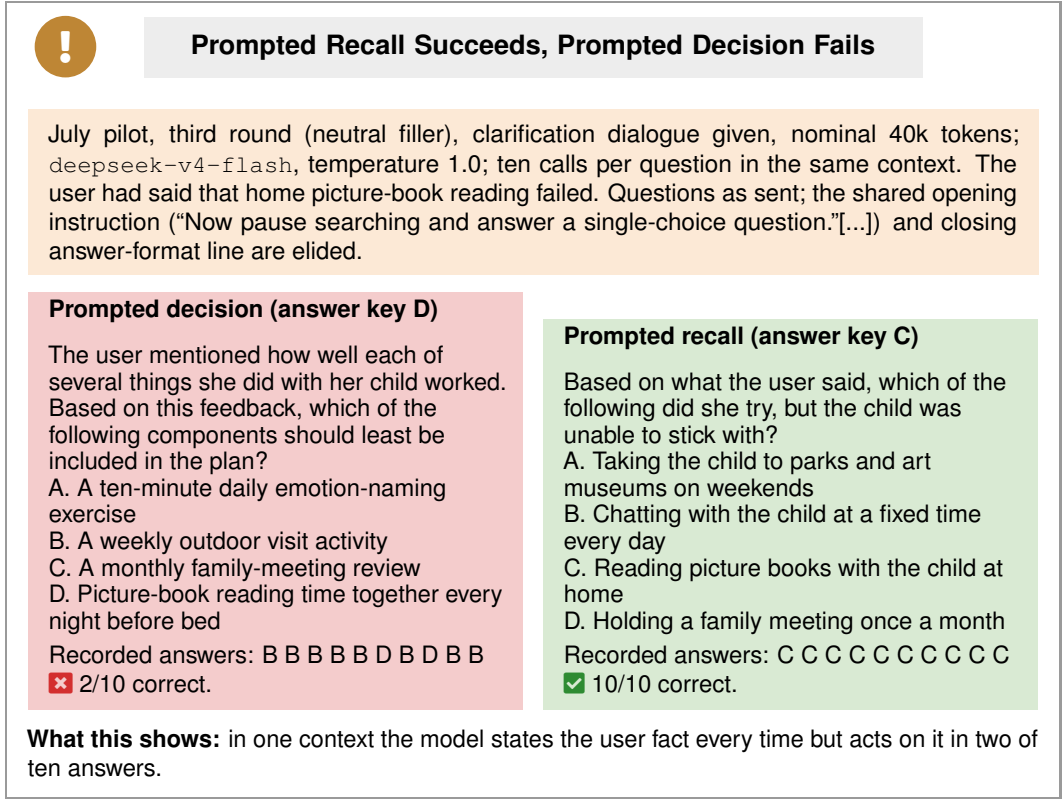

\begin{casebox}{\casewarn}{Prompted Recall Succeeds, Prompted Decision Fails}
\begin{caseband}July pilot, third round (neutral filler), clarification dialogue given, nominal 40k tokens; \texttt{deepseek-v4-flash}, temperature 1.0; ten calls per question in the same context. The user had said that home picture-book reading failed. Questions as sent; the shared opening instruction (``Now pause searching and answer a single-choice question.''[...]) and closing answer-format line are elided.\end{caseband}
\casecols{\begin{casepanel}[apxPink]{Prompted decision (answer key D)}
The user mentioned how well each of several things she did with her child worked. Based on this feedback,\allowbreak{} which of the following components should least be included in the plan?\\
A.\allowbreak{} A ten-minute daily emotion-naming exercise\\
B.\allowbreak{} A weekly outdoor visit activity\\
C.\allowbreak{} A monthly family-meeting review\\
D.\allowbreak{} Picture-book reading time together every night before bed\par\smallskip
Recorded answers: B B B B B D B D B B\\
\no\ 2/10 correct.
\end{casepanel}}{\begin{casepanel}[apxGreen]{Prompted recall (answer key C)}
Based on what the user said,\allowbreak{} which of the following did she try, but the child was unable to stick with?\\
A.\allowbreak{} Taking the child to parks and art museums on weekends\\
B.\allowbreak{} Chatting with the child at a fixed time every day\\
C.\allowbreak{} Reading picture books with the child at home\\
D.\allowbreak{} Holding a family meeting once a month\par\smallskip
Recorded answers: C C C C C C C C C C\\
\ok\ 10/10 correct.
\end{casepanel}}
\par\smallskip{\footnotesize\sffamily\textbf{What this shows:} in one context the model states the user fact every time but acts on it in two of ten answers.}
\end{casebox}
\caption{\textbf{Prompted recall versus prompted decision} (July pilot). Real questions and recorded answers.}
\label{fig:appG-recall-case}
\end{figure}

Two July pilots used one PDR-Bench task (a six-month parent--child communication plan), one user profile, and a manually constructed clarification dialogue.
The first ran no agent loop: filler material was appended after the task and user information to nominal lengths of 0--180k tokens, and \texttt{deepseek-v4-flash} (temperature 1.0) answered, for the same user fact, a recall question and a decision question that follows from it (7,020 calls in three rounds).
For the fact that home picture-book reading had failed, recall was correct in all 30 answers at every length, while the decision was correct at 1.00 without filler and 0.20 at 180k in the first round, and 0.00 at 180k in both neutral-filler rounds (Figure~\ref{fig:appG-recall-case}); of five decision items, one degraded strongly, one weakly, and three not at all.
The second pilot let an agent (DeepSeek v4-pro) run to natural submission, then restarted copies from its prefixes at the 4th, 8th, 16th, and 24th action with the fixed instruction ``The research phase has ended. Please submit the final product.'' (``The research phase is over. Please submit the final product,'' translation) and search disabled.
A rubric judge playing the user scored mean fit 0.89, 0.86, 0.95, and 0.95 at these steps and 0.91 for natural deliveries (10, 10, 10, 5, and 10 deliveries).
The pilot report records the limit: a run that first drafted at step 17 still delivered a complete 6,542-character plan when forced at step 4, so forced delivery measures reconstruction from a prefix, not the natural state at step $k$.
Both pilots completed and are not used.
On one task, they show that stating a user fact is not using it and that a prompted deliverable is not the natural intermediate state; the current method therefore inserts no questions, forces no drafts, and a run without a saved draft has no W state (Appendix~\ref{app:harness}).

\subsection{Reference-report Scoring}
\label{sec:appG-reference}

The aim was to score user specificity by rating each report under both user conditions, $\Delta_{\mathrm{specific}}=\tfrac12[S_A(Y_A)-S_B(Y_A)+S_B(Y_B)-S_A(Y_B)]$, or against reference reports.
This was not built: the method-development pipeline listed ``reference artifacts, A/B cross-scoring, decoy tests and judge calibration'' as not yet built, and the pilot user-fit rubric existed for only one of two twin users, so the twin's ten runs were never scored.
Two reference-like controls did run: reports written without user information scored 0.07 on the pilot rubric against 0.86--0.95 with the dialogue, so the rubric did not reward generic quality alone; and the conflict items of Section~\ref{sec:appG-representation} served as reference pairs with known answers.
The closest implemented method matched 378 behavior units across two report sides (180 and 198): 35,640 forward judgments gave 201 comparable edges (0.56\%) and 4 strict one-to-one matches, and the matrix was kept only as a record of failures.
Whether reference choice, style, length, or restatement would bias reference scoring was not tested; these are open possibilities, not established reasons.
The current method needs no reference report: it compares randomized $P$ and $Q$ runs of the same task with anonymous, mirrored endpoints.

\subsection{Direct Compliance Judgments and Other Retired Semantic Judgments}
\label{sec:appG-direct}

\begin{figure}[tbp]
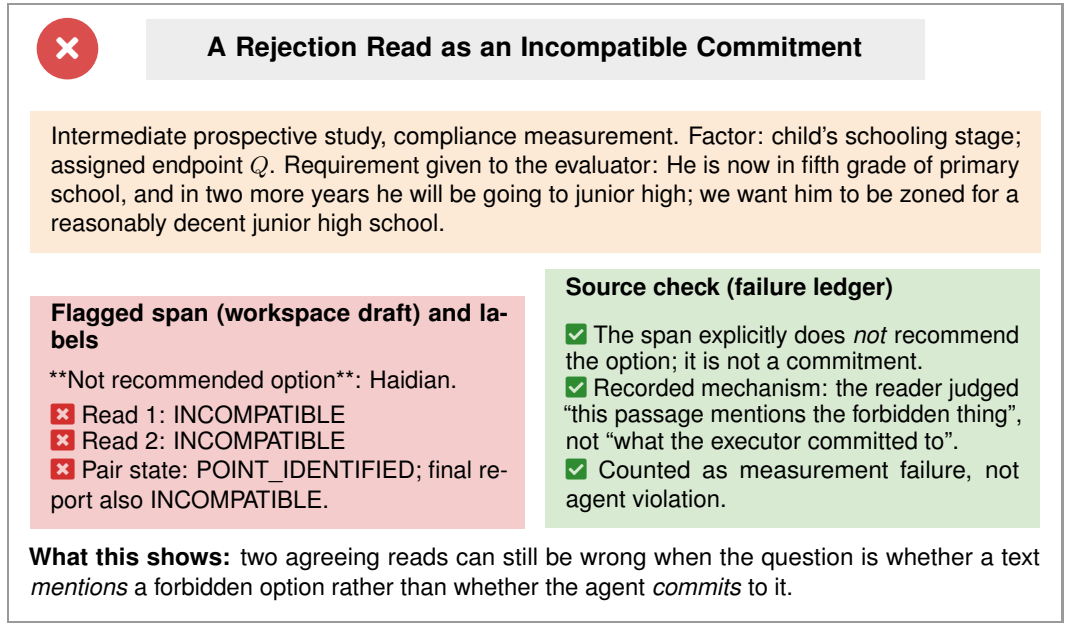

\begin{casebox}{\casefail}{A Rejection Read as an Incompatible Commitment}
\begin{caseband}Intermediate prospective study, compliance measurement. Factor: child's schooling stage; assigned endpoint $Q$. Requirement given to the evaluator: He is now in fifth grade of primary school, and in two more years he will be going to junior high; we want him to be zoned for a reasonably decent junior high school.\end{caseband}
\casecols{\begin{casepanel}[apxPink]{Flagged span (workspace draft) and labels}
**Not recommended option**: Haidian.\par\smallskip
\no\ Read 1: INCOMPATIBLE\\
\no\ Read 2: INCOMPATIBLE\\
\no\ Pair state: POINT\_IDENTIFIED; final report also INCOMPATIBLE.
\end{casepanel}}{\begin{casepanel}[apxGreen]{Source check (failure ledger)}
\ok\ The span explicitly does \emph{not} recommend the option; it is not a commitment.\\
\ok\ Recorded mechanism: the reader judged ``this passage mentions the forbidden thing'', not ``what the executor committed to''.\\
\ok\ Counted as measurement failure, not agent violation.
\end{casepanel}}
\par\smallskip{\footnotesize\sffamily\textbf{What this shows:} two agreeing reads can still be wrong when the question is whether a text \emph{mentions} a forbidden option rather than whether the agent \emph{commits} to it.}
\end{casebox}
\caption{\textbf{A false positive of the direct compliance measurement} (intermediate prospective study).}
\label{fig:appG-compliance-case}
\end{figure}

\paragraph{Commitment compliance.}
In the intermediate prospective study, a direct compliance measurement tried to locate the first public commitment incompatible with the assigned user requirement.
A user-blind extractor marked candidate sentences by number, spans were checked against the text, and a requirement-aware evaluator (\texttt{claude-haiku-4.5}; 424 extraction and 3,058 reading calls) labelled each candidate.
Outcomes are reported by kind, each with its own denominator.
\emph{Measurement failure:} of 225 observed commitment events, 216 had an applicable candidate and 24 of these 216 (11.1\%) were point-identified, while 195 were disputed; extraction was complete for 9 of 225 events, so the preregistered guard made ``complete, no deviation'' unreachable and its count of 0 reflects the gate, not the agents.
\emph{Implementation defects:} an earlier pipeline let the requirement-aware evaluator re-select candidates from the full text, bypassing user-blind extraction, and hard-coded span validation to true, so its 1,136 outputs were quarantined without reading their labels; an earlier version split only at blank lines (4,845 of 13,173 spans failed the verbatim check) and lost 202 of 424 extraction responses to unescaped quotation marks.
\emph{Transport:} the final version had 5 undecodable and 12 missing reads, and 3 truncated and 1 undecodable extraction.
\emph{Agent violations were not established:} 11 of 116 eligible units showed an ``exact strict deviation'', but on source check only 4 of the 26 incompatible candidates behind them looked like substantive commitments; others were table rows, negations, phrase-book lines, quoted examples, and list items (Figure~\ref{fig:appG-compliance-case}).
No violation rate is reported, and compliance labels are never combined with directional labels, since a directional UNKNOWN and a compliance UNKNOWN answer different questions.

\paragraph{Other retired judgments.}
A holistic judge that read two whole reports and assigned them to the two user conditions gave a direction on 76.7\% of 30 same-condition pairs, whose true difference is zero, against 65.7\% of 35 cross-condition pairs; its 0--10 distance output had an AUC of 0.336, below chance, and a preregistered stopping rule ended the route.
The pilot user-fit judge gave opposite verdicts on the same weekday-evening phone arrangement in two deliveries.
The current instrument therefore asks a narrower question, which endpoint each main-advice item expresses, with source references, code aggregation, and disagreement kept as UNKNOWN (Appendix~\ref{app:measurement_spec}); it measures direction, not compliance.

These alternative designs informed the final measurement but are not used in the primary analyses reported in the paper.

\section{Reproducibility, Corrections, and Limitations}
\label{app:reproducibility}

This appendix lists what is needed to recompute the reported numbers, the rules applied to failed runs and evaluations, the corrections to earlier analyses that bear on reported results, and the limits of the evidence.

\paragraph{Release scope.}
The release package is built from the archive in Table~\ref{tab:appH-release}: inputs, prompts and tool schemas, model routes, frozen design files and code, all run and evaluator records, row tables, and the scripts behind every table and figure.
In the main study, the follow-ups, and the harness comparison, agents run at temperature 1.0 with at most 16,000 output tokens per call and an undisclosed per-run budget of 96 model decisions, 192 tool actions, 96 retrieval requests, and 128,000 output tokens; retrieval returns six results (400-character snippets) and the first 6,000 characters of a page.
The primary evaluator runs with thinking disabled (zero reasoning tokens checked per call), temperature not sent, and at most 16,384 output tokens; per-study routes and prompts are in Appendices~\ref{app:harness} and~\ref{app:measurement_spec}.
Before release, API-key fragments in evaluator logs (\texttt{key\_tail}), local gateway addresses, and reviewer names are removed.

\begin{table}[t]
\centering
\caption{\textbf{Archive layout and identity keys.} Paths are relative to \texttt{exp\_v2/} unless they start with \texttt{exp\_v1/} or \texttt{PDR-Bench/}; \texttt{*} stands for the study folder.}
\label{tab:appH-release}
\footnotesize
\setlength{\tabcolsep}{3pt}
\begin{tabularx}{\linewidth}{@{}>{\raggedright\arraybackslash}p{0.95in}>{\raggedright\arraybackslash}X>{\raggedright\arraybackslash}p{1.5in}@{}}
\toprule
\textsf{\textbf{Component}} & \textsf{\textbf{Archive location}} & \textsf{\textbf{Identity key}}\\
\midrule
Tasks, users, factors & \path{PDR-Bench/data/task_data/tasks_zh.jsonl}; \path{*/USER_PROFILES*.jsonl}, \path{*/FACTOR_CARDS*.jsonl} & PDR commit \texttt{5b43f9f188c7}; \texttt{profile\_sha256}; \texttt{factor\_id}\\
Contract, routes, code & \path{ROUND2_PROSPECTIVE_V1/HARNESS_CONTRACTS/H00.json}; \path{*/ACTOR_ROUTE_REGISTRY.json}; \path{*/DESIGN_FREEZE.json}; \path{*/scripts/} & file sha256 (H00 \texttt{50eb581e\ldots}); per-file code hashes\\
Run records & \path{exp_v1/runs/formal_v1/}; \path{runs/natural_discovery_v1/MAIN/}; \path{*/RUNS*/} & \texttt{run\_uid}, \texttt{record\_sha256}, \texttt{slot\_id}\\
Evaluator records & \path{*/READ*/} (\path{packets/}, \path{raw_primary/}, \path{raw_aux/}, \path{ingested/}) & \texttt{packet\_id}, payload sha256\\
Row tables, texts & \path{ZERO_CALL_ANALYSIS_V1/ROW_TABLES/}, \path{TEXT_INDEX/}; \path{HARNESS_TRANSPORT_V1/ROW_TABLES/} & \path{MANIFEST.json}; \texttt{artifact\_text\_sha256}\\
Analyses, tables, figures & \path{*/FINAL_ANALYSIS/}; \path{exp_v1/runs/formal_v1/_FORMAL_ANALYSIS.json}; \path{experiments_overleaf/plot_sources/} & plan sha256; \texttt{\%~SOURCE} comments\\
Human review & \path{exp_v1/runs/experiment_A_v2/review_in/}; early-audit sheets (Appendix~\ref{app:human_validation}) & pseudonymous codes\\
\bottomrule
\end{tabularx}
\end{table}

No analysis step calls a model.
The entry points are \path{formal_analysis.py} (initial randomized study), \path{r3_core_rows.py} (main-study row tables; it recomputes every frozen state in memory and stops on any mismatch), \path{ht_analysis.py}~\texttt{MAIN} (harness comparison, including its task-block bootstrap with 10,000 resamples, seed 20260924), and \path{build_table2.py} and \path{plot_figure3.py}--\path{plot_figure5.py} for the printed tables and figures.

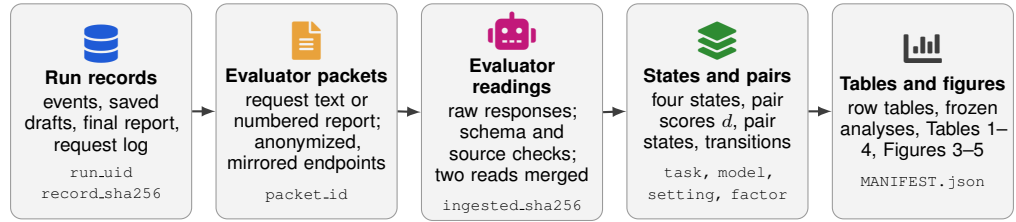
\begin{figure}[t]
\centering
\begin{tikzpicture}[
  stage/.style={draw=apxGrayFrame, fill=apxBody, rounded corners=1.5mm, text width=0.86in, minimum height=1.12in,
                align=center, inner sep=3pt, font=\scriptsize\sffamily, anchor=north},
  arr/.style={->, >=latex, thick, draw=apxDark}]
\node[stage] (s1) at (0,0) {\textcolor{apxHlP}{\Large\faDatabase}\\[2pt]\textbf{Run records}\\[1pt]events, saved drafts, final report, request log\\[3pt]{\tiny\ttfamily run\_uid\\record\_sha256}};
\node[stage] (s2) at (1.07in,0) {\textcolor{apxOrange}{\Large\faIcon{file-alt}}\\[2pt]\textbf{Evaluator packets}\\[1pt]request text or numbered report; anonymized, mirrored endpoints\\[3pt]{\tiny\ttfamily packet\_id}};
\node[stage] (s3) at (2.14in,0) {\textcolor{apxHlQ}{\Large\faRobot}\\[2pt]\textbf{Evaluator readings}\\[1pt]raw responses; schema and source checks; two reads merged\\[3pt]{\tiny\ttfamily ingested\_sha256}};
\node[stage] (s4) at (3.21in,0) {\textcolor{apxPassFg}{\Large\faLayerGroup}\\[2pt]\textbf{States and pairs}\\[1pt]four states, pair scores $d$, pair states, transitions\\[3pt]{\tiny\ttfamily task, model, setting, factor}};
\node[stage] (s5) at (4.28in,0) {\textcolor{apxDark}{\Large\faChartBar}\\[2pt]\textbf{Tables and figures}\\[1pt]row tables, frozen analyses, Tables~1--4, Figures~3--5\\[3pt]{\tiny\ttfamily MANIFEST.json}};
\draw[arr] (s1.east) -- (s2.west);
\draw[arr] (s2.east) -- (s3.west);
\draw[arr] (s3.east) -- (s4.west);
\draw[arr] (s4.east) -- (s5.west);
\end{tikzpicture}
\caption{\textbf{Dependency chain for reported numbers} (main study, follow-ups, harness comparison). Keys under each stage identify its records. The initial randomized study replaces the two evaluator stages with a frozen mechanical parser of the final report; the exploratory study used an earlier evaluator configuration (Appendix~\ref{app:measurement_spec}).}
\label{fig:appH-pipeline}
\end{figure}

\paragraph{From raw records to reported numbers.}
Figure~\ref{fig:appH-pipeline} shows the dependency chain; each stage reads only earlier stages and is addressed by its own keys, so every printed number traces back to run records.
Failed or disagreeing readings are carried forward as UNKNOWN; no stage drops them or scores them as zero.
The follow-up and harness freezes register a hash for every evaluator and derivation file; the main-study freeze does not, so its evaluator code identity rests on file timestamps that match the later registered bytes.

\paragraph{Attempts, retries, and censoring.}
Table~\ref{tab:appH-attempts} lists the rules actually applied.
No run is repeated because of its content.
An attempt \emph{establishes a trajectory} when it ran in the frozen system, was not stopped or broken by the operator, and produced at least one retrieval request, saved draft, handoff document, or delivered report; the first such attempt is the slot's run, selected from recorded cause codes only (\texttt{establishes\_trajectory} and \texttt{primary\_attempt}; the same code serves the main study, the follow-ups, and the harness comparison).

\refstepcounter{table}\label{tab:appH-attempts}

\paragraph{Corrections.}
Table~\ref{tab:appH-corrections} lists corrections that bear on reported results or on how they may be read.
Several concern an earlier prospective study that is not reported in this paper; they are listed because the corrected instrument, rule, or wording is the one used here.
The last column gives the status of the old conclusion: \badge{apxFailFg}{red} no longer stands, \badge{apxUnclearFg}{amber} corrected, reclassified, or open, \badge{apxPassFg}{green} unchanged.
Old records are preserved and marked, and none is used downstream.

\begingroup
\footnotesize
\setlength{\tabcolsep}{3pt}
\setlength{\LTcapwidth}{\linewidth}
\begin{longtable}{@{}>{\raggedright\arraybackslash}p{1.0in}>{\raggedright\arraybackslash}p{1.58in}>{\raggedright\arraybackslash}p{1.62in}>{\raggedright\arraybackslash}p{0.92in}@{}}
\caption{\textbf{Corrections that bear on reported results.} Numbers refer to the study named in the second column.}
\label{tab:appH-corrections}\\
\toprule
\textsf{\textbf{Original issue}} & \textsf{\textbf{Scope of impact}} & \textsf{\textbf{Current handling}} & \textsf{\textbf{Old conclusion}}\\
\midrule
\endfirsthead
\toprule
\textsf{\textbf{Original issue}} & \textsf{\textbf{Scope of impact}} & \textsf{\textbf{Current handling}} & \textsf{\textbf{Old conclusion}}\\
\midrule
\endhead
\bottomrule
\endfoot
\textsf{\textbf{Split gateway replies:}} dropped Claude tool calls. &
Main study: 4 attempts. Earlier study: all 8 Claude runs. &
Replies merged; 4 attempts replaced; earlier runs excluded, tables regenerated. &
\badge{apxFailFg}{VOID}\\
\textsf{\textbf{Invalid evaluator batches.}} &
Earlier study: 627 readings by an uncalibrated evaluator; 366 item readings with reversed rubric (agreement 33.7\% aligned vs.\ 52.1\% presented). &
Quarantined; items rerun with the corrected builder, later frozen for the reported studies. &
\badge{apxFailFg}{VOID}\\
\textsf{\textbf{BOTH and NONE merged}} as ``attenuation''. &
Earlier study: 11 of 12 ``net-zero'' first drafts were BOTH. &
States never merged; $d$ not used to describe loss. &
\badge{apxFailFg}{RETRACTED}\\
\textsf{\textbf{UNKNOWN read as inactive.}} &
Earlier study: re-entry through unknown gaps (one task 6/8 to 2/8). Exploratory: 368 uncertain requests coded ``no side''. &
Re-entry needs a fully readable gap; uncertain kept separate; reported numbers use the corrected analysis. &
\badge{apxUnclearFg}{CORRECTED}\\
\textsf{\textbf{Subset read as propagation:}} ``63/64 (98.4\%)''. &
Earlier study: all 64 finals in the subset were on the assigned side. &
Full table: 63/106 identical, 70/106 with zero states merged, 1/106 reversed; joint reading only. &
\badge{apxFailFg}{WITHDRAWN}\\
\textsf{\textbf{Risk-rule lookup}} with a wrong key. &
Main study: three continuous features always missing. &
Corrected lock on the discovery split. &
\badge{apxPassFg}{UNCHANGED}: no rule\\
\textsf{\textbf{Route halt; static field.}} &
Harness: an unrelated HTTP 400 halted DeepSeek (1 attempt, 6 slots unstarted); one model field is a constant. &
400 logged as provider error, attempt censored, not rerun; 6 slots run as first attempts; identity from requests and echoes. &
\badge{apxUnclearFg}{RECLASSIFIED}\\
\textsf{\textbf{Draft wording.}} &
``Median increase'' (a mean difference); post hoc case rule called fixed; 31/80 vs.\ 1/182 (different risk sets). &
Median 5 to 49.5; rule labelled post hoc; one risk set of 80 (49 OWN, 31 NONE, 0 OTHER). &
\badge{apxUnclearFg}{CORRECTED}\\
\textsf{\textbf{Monte Carlo $p$.}} &
Initial study: frozen $6.3\times10^{-4}$ (62/100,000) vs.\ exact $7.8\times10^{-4}$. &
Frozen value kept; $T$ reproduced from raw runs. &
\badge{apxPassFg}{KEPT}\\
\textsf{\textbf{Human-review summaries.}} &
``10 participants'', ``72/72'', ``two raters agreed on 39''; the review counts in Section~\ref{sec:measurement}. &
Recomputed from sheets: 2 annotators, 94 cards, $\kappa\le0.025$; 13-item audit, 2 reviewers + 1 adjudicator. Records matching the Section~\ref{sec:measurement} counts not located. &
\badge{apxFailFg}{WITHDRAWN}; \badge{apxUnclearFg}{OPEN}\\
\textsf{\textbf{Offered capability read as used.}} &
Harness: delegation 0/80; saved report = final (78/78, 79/79). GPT reasoning: protocol change. &
Reported as untested or not attributable; echoes do not verify weights. &
\badge{apxUnclearFg}{NOT TESTED}\\
\end{longtable}
\endgroup

\subsection{Scope and Limitations}
\label{app:limitations}

\paragraph{Coarse, partly readable measurement.}
The instrument records whether a behavior expresses $P$, $Q$, both, or neither for one registered factor; it does not measure amounts, thresholds, or degree of fit, an unchanged state does not rule out finer changes, and unregistered factors are not read.
The pair score $d$ codes BOTH and NONE alike as 0, and request and report scores do not share a scale.
Many comparisons are unreadable (62.7\% of planned main-study final pairs are readable; Table~\ref{tab:directional-results}); UNKNOWN is never imputed, and fixed-plan ranges can be wide (H-D: $[0.04, 0.99]$; Table~\ref{tab:execution-settings}).
Evaluators never see the assignment, but a report's text can reveal the user condition, and reads capped at 30 items may omit items.

\paragraph{Short trajectories.}
At the lowest reasoning setting, several model routes produce short executions relative to the available budget, and public drafts are sparse.
Consequently, fine-grained temporal claims require opportunity-aware subsets rather than assuming every run provides comparable long-horizon observations.

\paragraph{Reused tasks and dependence.}
Some archived studies reuse tasks from earlier phases, so cross-study agreement is not equivalent to replication on an untouched task holdout.
Within-run requests, factors, mirrored reads, and evaluator re-reads are dependent observations and are not counted as independent executions.

\paragraph{AI evaluators and human scope.}
All directional labels come from AI evaluators; mirrored reads share one model's biases, agreement across three families (Table~\ref{tab:evaluator-agreement}) is not accuracy, and the auxiliary evaluators' reasoning setting (service default) was not recorded.
Recovered human review covers 94 construction cards for the initial study and a 13-item audit of exploratory material (Appendix~\ref{app:human_validation}); these recovered records do not validate the later main-study, follow-up, or harness outputs, whose archived factor cards were screened by AI reviewers. The author-reported expanded protocol and proportionally scaled counts in Table~\ref{tab:human-evaluation} are a separate reporting layer.

\paragraph{Tested settings only.}
The analyzed archive covers a finite set of model routes, reasoning settings, retrieval infrastructure, and execution interfaces.
The conclusions should not be extrapolated to untested models, harnesses, languages, or interactive user settings.

\paragraph{Source data, consent, and privacy.}
Tasks and profiles come from PDR-Bench~\citep{liang2026pdr} at the archived repository revision; its repository ships an Apache-2.0 file labelled as the code license and describes the released profiles as authentic.
The source paper describes informed consent and de-identification of its profiles~\citep{liang2026pdr}; the local archive does not contain a separate data-license or participant-consent document. We therefore distinguish the source authors' description from records independently available for this study.
Counterfactual endpoints are constructed text, not attributes of real persons.
Our reviewers appear only under pseudonymous codes; records of their consent procedure or compensation were not located.

\end{document}